\documentclass[10pt,twocolumn,letterpaper]{article}

\usepackage{cvpr}              %

\usepackage[dvipsnames]{xcolor}

\newcommand{\todo}[1]{{\color{red}#1}}

\usepackage{multirow}
\usepackage{enumerate}
\usepackage{color}
\usepackage{xspace}
\usepackage{booktabs}
\usepackage{makecell}
\usepackage{dirtree}
\usepackage{adjustbox}
\usepackage{colortbl}
\usepackage{hhline}
\usepackage{textcomp}  %
\usepackage{scalerel}  %
\usepackage{float}
\usepackage{subcaption}

\providecommand{\ie}{\textit{ie}\xspace}
\providecommand{\mypara}[1]{{\noindent{\bf #1}}}
\usepackage{enumitem}

\usepackage[accsupp]{axessibility}

\providecommand{\ObjectNav}{\mbox{\sc{ObjNav}}\xspace}
\providecommand{\PickUp}{\mbox{\sc{PickUp}}\xspace}
\providecommand{\Fetch}{\mbox{\sc{Fetch}}\xspace}
\providecommand{\SimpleExploreHouse}{\mbox{\sc{RoomVisit}}\xspace}

\providecommand{\ObjectNavAffordance}{\mbox{\sc{ObjNavAfford}}\xspace}
\providecommand{\ObjectNavLocalRef}{\mbox{\sc{ObjNavLocalRef}}\xspace}
\providecommand{\ObjectNavRelAttr}{\mbox{\sc{ObjNavRelAttr}}\xspace}
\providecommand{\ObjectNavRoom}{\mbox{\sc{ObjNavRoom}}\xspace}
\providecommand{\ObjectNavOpenVocab}{\mbox{\sc{ObjNavDesc}}\xspace}
\providecommand{\RoomNav}{\mbox{\sc{RoomNav}}\xspace}

\providecommand{\bench}{\mbox{\sc{Chores}}\xspace}

\providecommand{\benchnav}{\mbox{\sc{ChoresNav}}\xspace}
\providecommand{\fifteennodash}{\mbox{$\mathbb{S}$}\xspace}
\providecommand{\alltypenodash}{\mbox{$\mathbb{L}$}\xspace}
\providecommand{\fifteen}{\mbox{-\fifteennodash}\xspace}
\providecommand{\alltype}{\mbox{-\alltypenodash}\xspace}

\def\spockemoji{\scalerel*{\includegraphics{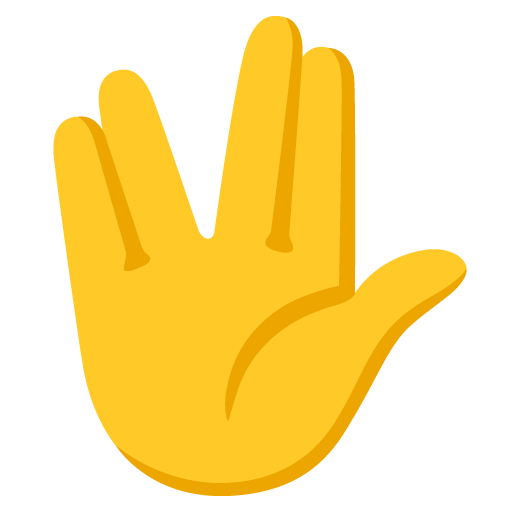}}{\textrm{\textbigcircle}}\xspace}

\providecommand{\model}{\mbox{\sc{Spoc}}\xspace} %
\providecommand{\modelb}{\mbox{\sc{\textbf{Spoc}}}\xspace} %

\providecommand{\clip}{\mbox{\sc{Clip}}\xspace}
\providecommand{\detic}{\mbox{\sc{Detic}}\xspace}
\providecommand{\siglip}{\mbox{\sc{SigLIP}}\xspace}
\providecommand{\dino}{\mbox{\sc{DinoV2}}\xspace}
\providecommand{\tfive}{\mbox{\sc{T5}}\xspace}
\providecommand{\numvidatasks}{10\xspace}

\providecommand{\numsynsets}{863\xspace}
\providecommand{\numhypernyms}{1,371\xspace}

\providecommand{\objaversehome}{\emph{ObjaverseHome}\xspace}

\definecolor{SuccessColor}{rgb}{0.818,0.9,0.983}
\definecolor{SELColor}{rgb}{0.95,0.95,0.95}
\definecolor{\%RoomsColor}{gray}{1}
\definecolor{RoomsColor}{gray}{1}

\providecommand{\FetchType}{\Fetch}

\providecommand{\ObjectNavRelAttribute}{\ObjectNavRelAttr}

\providecommand{\ObjectNavType}{\ObjectNav}

\providecommand{\PickupType}{\PickUp}
\providecommand{\RoomNav}{\mbox{\sc{NavigateToRoom}}\xspace}

\definecolor{cvprblue}{rgb}{0.21,0.49,0.74}
\usepackage[pagebackref,breaklinks,colorlinks,citecolor=cvprblue]{hyperref}

\def\paperID{694} %
\def\confName{CVPR}
\def\confYear{2024}

\title{SPOC: Imitating \emph{Shortest Paths} in Simulation \\Enables Effective Navigation and Manipulation in the Real World}

\author{
\makebox[\linewidth][s]{\makebox{Kiana Ehsani}$^{\dagger*}$ \makebox{Tanmay Gupta}$^{\dagger*}$ \makebox{Rose Hendrix}$^{\dagger*}$ \makebox{Jordi Salvador}$^{\dagger*}$ \makebox{Luca Weihs}$^{\dagger*}$ \makebox{Kuo-Hao Zeng}$^{\dagger*}$}\\
Kunal Pratap Singh$^{\ddagger}$
\qquad Yejin Kim$^{\dagger}$
\qquad Winson Han$^{\dagger}$
\qquad Alvaro Herrasti$^{\dagger}$\\
Ranjay Krishna$^{\dagger\psi}$
\qquad Dustin Schwenk$^{\dagger}$
\qquad Eli VanderBilt$^{\dagger}$
\qquad Aniruddha Kembhavi$^{\dagger\psi}$\\
\vspace*{-0.3cm}\\
$^{\dagger}$Allen Institute for AI \qquad $^{\psi}$University of Washington \qquad $^{\ddagger}$EPFL
}

\let\svthefootnote\thefootnote
\newcommand\freefootnote[1]{%
  \let\thefootnote\relax%
  \footnotetext{#1}%
  \let\thefootnote\svthefootnote%
}

\begin{document}

\twocolumn[{
\renewcommand\twocolumn[1][]{#1}
\maketitle
\vspace*{-0.25in}
\centering
\captionsetup{type=figure}\includegraphics[width=0.98\linewidth]{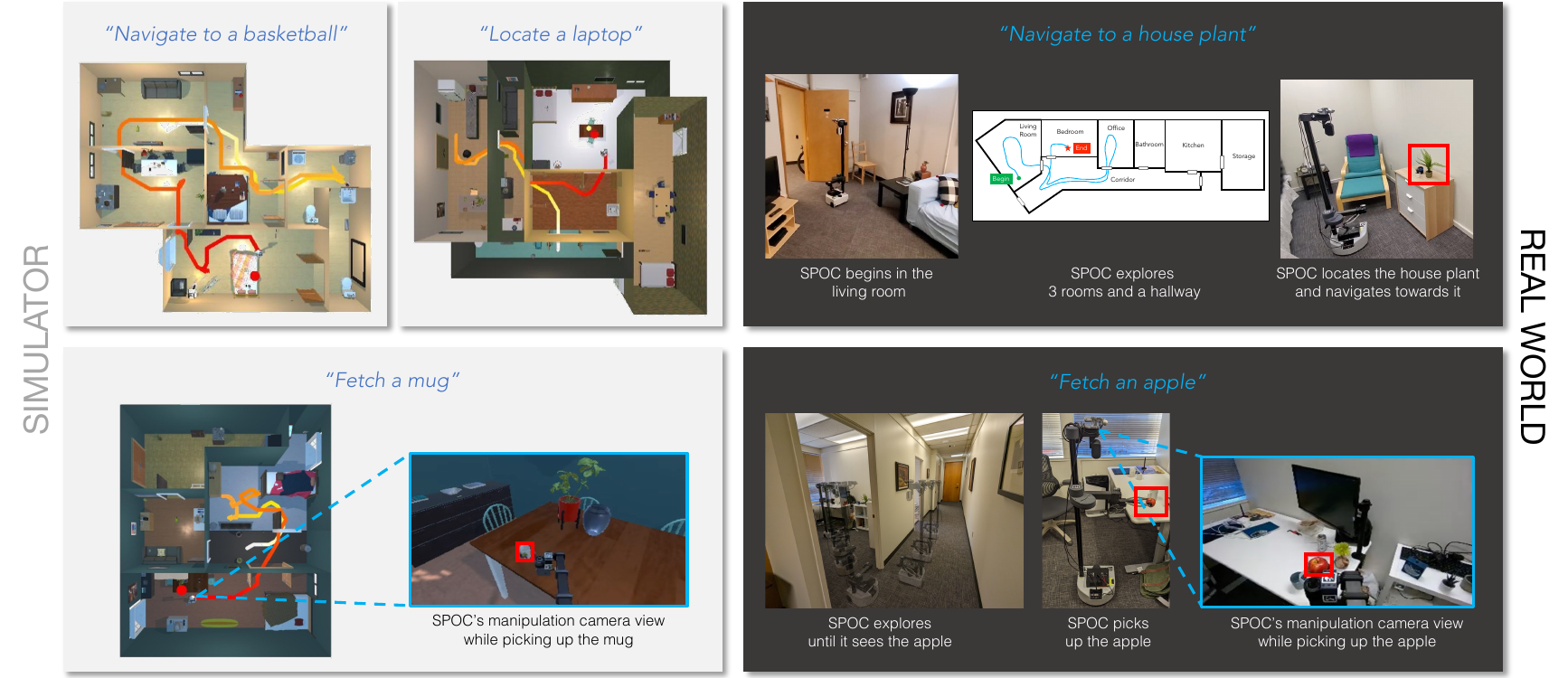}

\captionof{figure}{We present \model, an embodied navigation and manipulation agent trained by imitating shortest-path experts in simulation. Visualized paths in simulation are white in the beginning and red at the end; red circles and squares highlight target object locations only for illustration. \emph{Top-Left:} A variant of \model, \textbf{trained only on shortest path episodes for object goal navigation} demonstrates complex behavior like exploration, obstacle avoidance, and back-tracking in novel environments; \emph{Top-Right}: The same agent transfers to the real world with no further adaptation and navigates to the house plant; \emph{Bottom-Left:} \model exploring the house to navigate to the mug and then picking it up; \emph{Bottom-Right:} \model navigating and picking up objects in the real world, again with no change in weights. \\[0.01in]}
\label{fig:teaser}
}]
\maketitle
\freefootnote{$^{*}$Equal contribution, ordered alphabetically by last name}

\begin{abstract}
\vspace{-0.2cm}

        Reinforcement learning (RL) with dense rewards and imitation learning (IL) with human-generated trajectories are the most widely used approaches for training modern embodied agents. RL requires extensive reward shaping and auxiliary losses and is often too slow and ineffective for long-horizon tasks. While IL with human supervision is effective, collecting human trajectories at scale is extremely expensive. In this work, we show that imitating shortest-path planners in simulation produces agents that, given a language instruction, can proficiently navigate, explore, and manipulate objects in both simulation and in the real world using only RGB sensors (no depth map or GPS coordinates). This surprising result is enabled by our end-to-end, transformer-based, \model architecture, powerful visual encoders paired with extensive image augmentation, and the dramatic scale and diversity of our training data: millions of frames of shortest-path-expert trajectories collected inside approximately 200{,}000 procedurally generated houses containing 40{,}000 unique 3D assets. Our models, data, training code, and newly proposed 10-task benchmarking suite \bench{} are available in \href{https://spoc-robot.github.io}{\texttt{spoc-robot.github.io}}. %

\end{abstract}

\vspace{-0.2cm}

\vspace{-1em}
\section{Introduction}
\label{sec:intro}
The prevalent method to build embodied agents employs a rich set of sensors such as RGB, depth maps, and GPS coordinates that feed into modular architectures with mapping components~\cite{chaplot2020neural,chaplot2020learning,gervet2023navigating,chang2023goat}, off-the-shelf computer vision models~\cite{Khandelwal2022EmbCLIP,yadav2023offline,majumdar2023we}, and even large language models (LLM)  that can be used for planning~\cite{Huang2022LanguageMA,song2023llm,liu2023llm+,wang2023describe,vemprala2023chatgpt,Ichter2022SayCan}. These agents are either trained with on-policy reinforcement learning (RL) using careful reward shaping and auxiliary losses~\cite{Singh2023SGC,zeng2021pushing,guo2020bootstrap,guo2018neural}, which tends to be slow and ineffective, or trained with imitation learning (IL) on large corpora of human demonstrations, which is incredibly expensive. The transfer gap from simulators to real is often mitigated by optimizing the photorealism of the simulator or via sim-to-real image translation models.
Finally, at test time, it is common practice if navigation is required to provide the locations of objects or assume a map of the environment is known~\cite{Ichter2022SayCan, Brohan2023RT1, Yokoyama2023AdaptiveSC}. 

In contrast to the above, in this paper, we surprisingly find that \emph{imitating shortest path experts in simulation can produce embodied agents effective at navigation, exploration, and manipulation, in both simulation and the real world}. 
Our model, \model (\textbf{S}hortest \textbf{P}ath \textbf{O}racle \textbf{C}lone), uses (a) only RGB observations with no depth and no GPS sensors, (b) a transformer-based architecture with no mapping module and no LLM, and (c) imitation learning on heuristic \emph{shortest path} planners with no human demonstrations and no RL. \model is trained in the AI2-THOR~\cite{Kolve2017AI2THORAI} simulator transfers well into the real world with no adaptation or fine-tuning. This all without making any assumptions about  scene layout or object appearance at test time.

Notably and unexpectedly \model, when trained to imitate a \emph{shortest path} expert for the singular task of object-goal navigation, demonstrates the capability to explore its environment comprehensively, peek into rooms, and backtrack along its path
as it searches for its target, despite never having seen this behavior in its training data. Fig.~\ref{fig:teaser} visualizes paths that showcase the exploration capabilities of \model.
We hypothesize that \model's ability to be an effective explorer is less hampered by the use of \emph{shortest path} training data and instead seems gated by its object perception.
Indeed, perception errors, not exploration failures, appear to be the primary cause of failures: employing ground truth target object detection alongside raw RGB results in a very high success rate of 85\% for \ObjectNav.

\model is a very effective multi-tasking agent. When trained jointly on four tasks -- Object-Goal Navigation (\ObjectNav), Room Visitation (\SimpleExploreHouse), PickUp Object (\PickUp), and Fetch Object (\Fetch) -- \model achieves an impressive average success rate of 49.9\% in unseen simulation environments at test time. These numbers easily outperform carefully tuned agents trained with reinforcement learning by a large margin of close to 30 points across the task suite. The high success rates achieved by \model translate to the real world where it achieves 56\% success across these tasks. We further train \model to follow open vocabulary instructions via a suite of seven navigation tasks where it achieves a high success rate of 51\%. This benchmark is designed to evaluate a wide range of capabilities such as recognizing objects, discerning affordances, identifying scene elements, recognizing relative attributes of objects (e.g. \emph{bigger}, \emph{lower}), and understanding local references (e.g. \emph{near}, \emph{on}).

We identify four key factors that enable effective imitation learning from heuristic experts. First, diversity of simulated worlds plays an important role -- we used almost 3 orders of magnitude more unique houses to train with IL than past work~\cite{Weihs2021Advisor,ramrakhya2023pirlnav,Ramrakhya2022HabitatWeb,yadav2023offline}. Second, using powerful visual encoders is critical -- we moved from the defacto ResNet-50 \clip encoder~\cite{Radford2021CLIP} employed in the literature~\cite{Khandelwal2022EmbCLIP} to \dino~\cite{oquab2023dinov2} and \siglip~\cite{Zhai2023SigLIP} and found huge gains. Third, moving to transformer architectures with long context windows of up to 100 frames outperforms previously employed recurrent architectures~\cite{AllenAct}. Finally, scaling up the size of the training data matters.%

This work shows the promise of imitating heuristic experts in simulators as a means to develop capable robots for the real world. Our experiments show that scale and diversity play an important role in enabling this behavior, and we posit that further scaling up this paradigm has huge merits and can lead to large improvements on challenging tasks.

\section{Related Work}
\label{sec:related}

\noindent\textbf{Simulators, tasks, and benchmarks.} Rapid progress in Embodied AI has led to an explosion of simulators, tasks, and benchmarks. 
Early simulators were built for navigation, often using 3D scans of the real world and supported only basic, if any, object interaction~\cite{Xia2018Gibson, Anderson2018R2R,Puig2018VirtualHome,Kolve2017AI2THORAI,Deitke2020RoboTHOR,Savva2019Habitat}.
Recent simulators model realistic robotic agents but often trade off physical fidelity to increase simulation speed~\cite{Xiang2020SAPIEN,Shen2021iGibson1,Li2021iGibson2,Li2022Behavior, Gan2021ThreeDWorld,Ehsani2021ManipulaTHOR,Ehsani2022ObjDis,Szot2021Habitat2,Puig2023Habitat3}. We use the AI2-THOR environment~\cite{Kolve2017AI2THORAI} and ProcTHOR~\cite{Deitke2022ProcTHORLE} to produce unbounded numbers of procedurally generated households which support object manipulation with a Stretch RE-1~\cite{Kemp2022StretchRobot}. 

Many embodied AI benchmarks focus on navigation~\cite{Zhu2017VisualNav,Wijmans2020DDPPO,Batra2020ObjectNav,Deitke2020RoboTHOR,Chen2020SoundSpaces, Chen2022SoundSpaces2,Anderson2018R2R,Ku2020RxR,Qi2020Reverie}.
Beyond navigation, many tasks (\eg, \textsc{ALFRED}~\cite{Shridhar2020ALFRED}, Visual Room Rearrangement~\cite{Weihs2021Rearrangement}, \textsc{ArmPointNav} / \textsc{ObjDis}~\cite{Ehsani2021ManipulaTHOR,Ehsani2022ObjDis}, BEHAVIOR~\cite{Srivastava2021Behavior100, Li2022Behavior}, and others~\cite{Gan2021TransportChallenge,Gao2022DialFRED}) require the agent to interact directly with objects in the environment at various levels of abstraction. Most similar to the tasks used in this work is \emph{Open-Vocabulary Mobile Manipulation} (\textsc{OVMM})~\cite{Yenamandra2023HomeRobotOVMM} task in which an RE-2 Stretch robotic agent must transport an object of a given type from an initial receptacle of a given type to a goal receptacle of a different type. While \textsc{OVMM} focuses on the core task of mobile manipulation, we benchmark our agent across a wide variety of tasks related to navigation, manipulation, and language understanding.

\noindent\textbf{Embodied architectures and training methods.}
A common architecture for Embodied AI is based on an observation encoder implemented by means of a CNN, and an RNN 
providing episode memory and producing the hidden state upon which to condition the policy at each step. For the visual backbone, common choices include \emph{ResNets}~\cite{He2016ResNet} or CLIP~\cite{Radford2021CLIP, Khandelwal2022EmbCLIP}, or trained from scratch~\cite{Wijmans2020DDPPO}. \cite{Majumdar2023VisualCortex} have comprehensively studied the impact of visual backbones on embodied performance.

Embodied agents are frequently trained with on-policy actor-critic RL methods, \eg A3C~\cite{Mnih2016A3C}, A2C~\cite{Schulman2017A2CPPO}, or DD-PPO~\cite{Wijmans2020DDPPO}. Auxiliary losses co-trained with these RL have also been proposed to improve sample efficiency \cite{Ye2021AuxTasksObjectNav, Singh2023SGC}. Transformer-based architectures have been proposed in combination with IL and BC bypassing the need to use RL for training. Decision transformer \cite{Chen2021DecisionTransformers} and trajectory transformer \cite{Janner2021TrajTransf} cast the sequential decision problem as a sequence modeling problem, thus enabling BC to replace RL. \emph{Gato} \cite{Reed2022GATO} trains an autoregressive transformer across tasks and embodiments, towards extended generalization. 

Imitation learning has recently gained popularity in robotics, significantly impacting areas like autonomous driving~\cite{pulver2021pilot,Pan2019ImitationLF,LeMero2022ASO,Codevilla2017EndtoEndDV,Li2018OILOI,Kebria2020DeepIL}.  With the effectiveness of in-context learning, using LLMs as robotic planners has become increasingly prominent \cite{huang2022inner, Dasgupta2023CollaboratingWL, Ichter2022SayCan}. RT-1 \cite{Brohan2023RT1} scales up model capacity and multi-task data focusing on manipulation. RT-2 \cite{Brohan2023RT2} encodes actions as text tokens to enable large joint VL and decision-taking pretraining for further generalization improvements in manipulation. \emph{DualMind} \cite{Wei2023imitation} is trained with self-supervised learning on state-action interactions and imitation learning with prompts. The main bottleneck of these approaches is the tremendously costly human generated trajectories, frequently in the real-world. Importantly, these works usually do not train navigation and manipulation jointly, frequently assuming ground-truth navigation knowledge or that the navigation can be solved by SLAM-based systems. We train an end-to-end system using cheap, shortest-path-planner generated data, and show its effectiveness in the real-world.

\noindent\textbf{Sim-to-Real.} One way to reduce the sim-to-real gap is to train in high-fidelity simulators. However, accurately modeling the real world, including camera miscalibration, or out-of-distribution lighting changes, is hard. An alternative is domain randomization \cite{Tobin2017DomainRand,Chen2022DomainRand}, where camera poses, lighting, textures, or visual degradations can be randomly sampled during training time, thus making the trained agent resilient to such fluctuations. Another type of augmentation is \emph{Phone2Proc} \cite{Deitke2023Phone2Proc}, where a scanned layout of the real-world house is used to generate many simulated variations for agent fine-tuning. Finally, domain transfer as in CycleGAN \cite{Zhu2017CycleGAN} allows adapting visual appearances during training (sim-to-real) or inference (real-to-sim). RetinaGAN \cite{Ho2021RetinaGan} additionally enforces object-detection consistency and is employed in \cite{Ichter2022SayCan}.

\begin{figure}
    \centering
    \includegraphics[width=0.9\columnwidth]{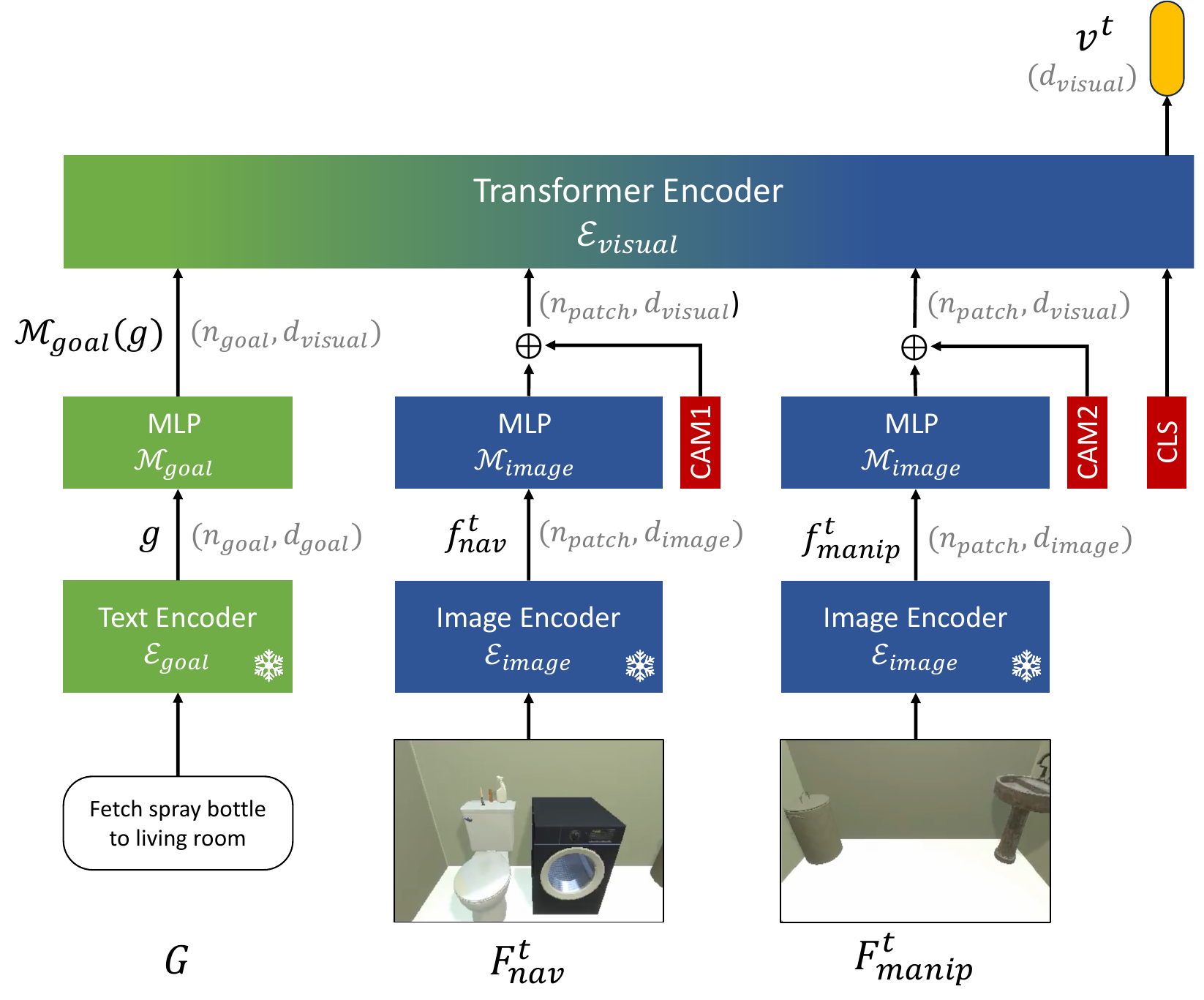}
    \caption{\textbf{Goal-conditioned Visual Encoder} for extracting goal-relevant visual information from the two cameras.}
    \vspace{-0.3cm}
    \label{fig:visual_encoder}
\end{figure}

\begin{figure}
    \centering
    \includegraphics[width=0.9\columnwidth]{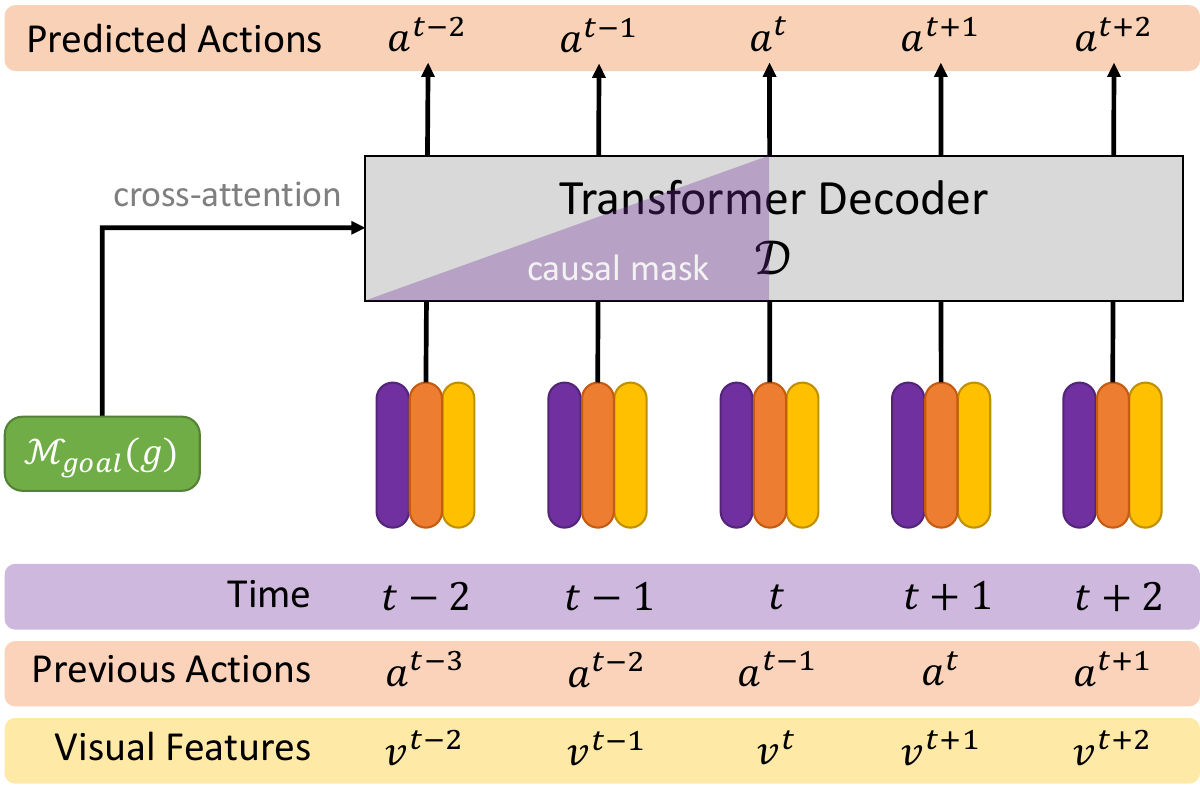}
    \caption{\textbf{Action Decoder} for predicting action at the current time step given the goal, current and past observations, and past actions.} 
    \label{fig:action_decoder}
    \vspace{-1em}
\end{figure}

\section{Imitation in Procedural Houses}
\label{sec:framework}

Embodied agents are commonly trained in simulation using on-policy RL. The agent's actions result in new observations and rewards from the environment, and the policy is updated using the rewards that guide the agent towards desirable behaviors. Practically, RL in complex visual worlds is sample inefficient, especially when using large action spaces and for long-horizon tasks. Such training is bottlenecked by the simulator's speed; the more physically and visually realistic the simulation, the lower the frame-rate. Finally, RL training, even relatively simple tasks such as navigating to an object or picking it up, requires careful reward shaping, auxiliary losses, and modular architectures.

IL is a compelling alternative to RL since learning from expert trajectories can be cast as a supervised learning problem. However, IL's big success stories have required a lot of data. Data for IL has traditionally been collected from two sources: (1) expert humans~\cite{Ramrakhya2022HabitatWeb,Brohan2023RT1} and (2) heuristic planners operating from ground-truth information not available during inference time (\eg shortest paths computed using navigation meshes in simulation). While human-collected data is the gold standard, it is extremely expensive. Planner-based approaches are cheap 
but have been found in prior work to result in suboptimal learning; for instance, \cite{Ramrakhya2022HabitatWeb} found that navigation agents trained to follow shortest paths achieved success rates of only 4.4\% on the \textsc{ObjectNav} task on the MP3D~\cite{Matterport3D} validation dataset versus ${\approx}$35\% success rates when trained to imitate human trajectories. There is also a healthy skepticism for the generalization ability of shortest-path trained IL agents in novel environments where the agent needs to balance exploration and exploitation to achieve a goal while simultaneously building an implicit map of the environment. Moreover,~\cite{Weihs2021Advisor} mathematically proves that learning such sub-optimal behavior is guaranteed in some settings due to an ``imitation gap''.

In the following sections, we show that using transformer architectures with long context windows to imitate heuristic planners at scale can help unlock the power of IL and produce effective agents in simulation and the real world. Sec.~\ref{sec:model} details our agent, \model. In Sec.~\ref{sec:procedural_data} we outline our large-scale data collection, made possible by recent breakthroughs in procedurally generating home environments~\cite{Deitke2022ProcTHORLE}, access to Objaverse 3D assets~\cite{Deitke2022ObjaverseAU}, and efficient heuristic planners that leverage rich ground truth information in the AI2-THOR simulator. We present a new benchmark, \bench in Sec.~\ref{sec:benchmark} and a comprehensive 
analysis of \model in Sec.~\ref{sec:experiments}.

\begin{figure}
    \centering
    \vspace{-2em}
    \includegraphics[width=0.85\linewidth, trim=4cm 2cm 4cm 1cm, clip]{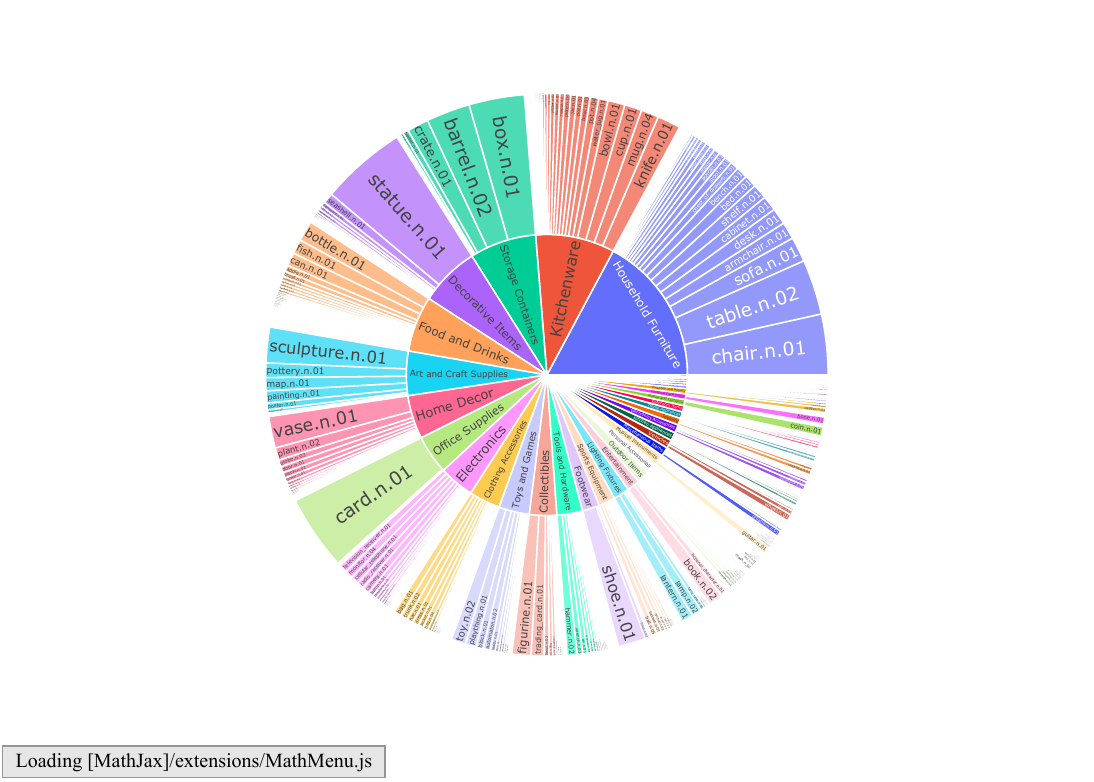}
    \caption{\textbf{Diversity of assets} in training environments.}
    \label{fig:sunburst}
    \vspace{-0.6cm}
\end{figure}

\section{The Shortest Path Oracle Clone (\model)}%
\label{sec:model}

We present \model{}\spockemoji, an agent embodied in the Stretch RE-1~\cite{Kemp2022StretchRobot} robot, trained in simulation to follow text instructions and complete long-horizon navigation and manipulation tasks. \model takes as input the text instruction and visual observation at each time step $t$ and predicts an action $a^t$. The Stretch Robot's axis of navigation is perpendicular to the axis of manipulation. This necessitates two RGB cameras, one pointing in the direction of navigation
and the other pointing at the arm. %
We discretize the action space into 20 actions: Move Base ($\pm 20$ cm), Rotate Base ($\pm 6^\circ$, $\pm 30^\circ$), Move Arm ($x,z$) ($\pm 2$ cm, $\pm 10$ cm), Rotate Grasper ($\pm 10^\circ$), pickup, dropoff, done with subtask, and terminate.

Our model consists of three main components: (1) a textual goal encoder, which processes open vocabulary language instructions; (2) an instruction-conditioned visual encoder for encoding visual inputs at each time step; and (3) a high-capacity causal transformer action decoder that predicts the action for the current time step given the goal, the current and previous visual inputs, and previous actions.

\noindent\textbf{Goal Encoder.} We use a pretrained text encoder $\mathcal{E}_\text{goal}$ that maps the goal specification $\mathcal{G}$ into a sequence of contextualized token representations $\mathbf{g}=\mathcal{E}_\text{goal}(\mathcal{G}) \in \mathbb{R}^{n_\text{goal}\times d_\text{goal}}$ where $n_\text{goal}$ and $d_\text{goal}$ are the number of sub-word tokens in the goal text and dimension of the token representation, respectively. We experiment with \tfive and \siglip text encoders.

\noindent\textbf{Goal-Conditioned Visual Encoder.} \model accepts visual inputs from two RGB cameras pointing in perpendicular directions. 
At any time step $t$, the goal-conditioned visual encoder $\mathcal{E}_\text{visual}$ extracts and integrates visual information from the two RGB frames $\mathcal{F}_{\text{nav}}^t$ and $\mathcal{F}_{\text{manip}}^t$ $\in\mathbb{R}^{h\times w\times 3}$ and represent it as a single vector $\mathbf{v}^t=\mathcal{E}_\text{visual}(\mathcal{F}_{\text{nav}}^t,\mathcal{F}_{\text{manip}}^t,\mathcal{G})$.
We use a Transformer encoder to achieve this, shown in Fig.~\ref{fig:visual_encoder}. The frames are encoded into sequences of contextualized patch embeddings $\mathbf{f}_{\text{nav}}^t$ and $\mathbf{f}_{\text{manip}}^t$ $\in\mathbb{R}^{n_\text{patch}\times d_\text{image}}$ using a pretrained image encoder $\mathcal{E}_\text{image}$ where $n_\text{patch}$ and $d_\text{image}$ are the number of image patches and output feature dimension of the image encoder.  These features are mapped to the transformer input dimension $d_\text{visual}$ using an MLP $\mathcal{M_\text{image}}$ with ReLU and LayerNorm. The goal representation $\mathbf{g}$ is also mapped to the $d_\text{visual}$ dimension using another MLP $\mathcal{M}_\text{goal}$. Next, we add learnable camera-type embeddings to differentiate features from the two cameras. Finally, we concatenate the patch features,  goal features, and a learnable [CLS] token embedding along the patch dimension and input this $(2n_\text{patch}+n_\text{goal}+1)\times d_\text{visual}$ tensor through the transformer encoder. The output at the position of the [CLS] token serves as the goal-conditioned visual representation $\mathbf{v}^t$.

\noindent\textbf{Action Decoder.} We use an autoregressive Transformer decoder $\mathcal{D}$ with causal masking to predict actions, see Fig.~\ref{fig:action_decoder}. The input to the decoder is the sequence of previous and current visual representations $\{\mathbf{v}^0,\cdots,\mathbf{v}^t\}$ additively combined with sinusoidal temporal position encodings and learned previous time step action embeddings. The decoder conditions on goal encoding $\mathcal{M}_\text{goal}(\mathbf{g})$ using cross-attention. At each time step, the output embedding from the transformer decoder is fed through linear and softmax layers to predict an action distribution for that time step
$\pi^t = \text{Softmax}(\text{Linear}(\mathcal{D}(\mathbf{v}^{0:t},a^{0:t-1};\mathcal{M}_\text{goal}(\mathbf{g}))[t]))$.
Causal masking during training ensures the decoder only attends over current and past inputs to predict the current action. The model is optimized using the cross-entropy loss in a teacher-forcing manner, \ie we minimize the cross entropy between $\pi^t$ and the one-hot encoding of the expert action for the current timestep. During inference, the agent acts in the environment at time $t$ by sampling an action $a^t$ from $\pi^t$; $a^t$ is then fed as an input to the model on the following timestep. For compute-efficient mini-batch training, we train with a limited temporal context window (\eg 100), but the model uses all past observations during inference. To enable using a larger context window during deployment, we randomly shift the time indices fed to the agent during training; in particular, if we sample temporal context window $[s, s+99]$ from an expert trajectory during training, then we input the corresponding actions and visual features to the model as-is, but pair them with the shifted time indices $[s+\ell, s+99+\ell]$ where $\ell\sim\text{Unif}\{0,\ldots, 900\}$.

\section{Procedural Data}
\label{sec:procedural_data}

We now describe our large-scale dataset of diverse household environments and the planners we use to generate expert trajectories within these environments.

\begin{table}[t]
    \centering
    \footnotesize
    \resizebox{\columnwidth}{!}{%
        \begin{tabular}{|p{0.15\columnwidth}|p{0.9\columnwidth}|}
            \hline
            \textbf{Task} & \textbf{Description \& Example} \\
            \hline
            \ObjectNav & Locate an object category: ``find a mug'' \\
            \hline
            \PickUp & Pick up a specified object in agent line of sight: ``pick up a mug'' \\
            \hline
            \Fetch & Find and pick up an object: ``locate a mug and pick up that mug'' \\
            \hline
            \SimpleExploreHouse & Traverse the house. ``Visit every room in this 5-room house. Indicate when you have seen a new room and when you are done.'' \\
            \hline
        \end{tabular}%
    }
    \vspace{-0.1cm}
    \caption{\bench tasks.}
    \label{tab:bench_tasks}
    \vspace{-1em}
\end{table}

\begin{table}[t]
    \centering
    \footnotesize
    \resizebox{\columnwidth}{!}{%
        \begin{tabular}{|p{0.3\columnwidth}|p{0.7\columnwidth}|}
            \hline
            \textbf{Task} & \textbf{Target Description \& Example} \\
            \hline
            \ObjectNav & Object's category: ``vase'' \\
            \hline
            \ObjectNavAffordance & Object's possible uses: ``a container that can best be used for holding fresh flowers decoratively'' \\
            \hline
            \ObjectNavLocalRef & Object's nearby objects: ``a vase near a tennis racket and a basketball'' \\
            \hline
            \ObjectNavRelAttr & Object category comparative attribute: ``the smallest vase in the bedroom'' \\
            \hline
            \ObjectNavRoom & Object's room type: ``vase in the living room'' \\
            \hline
            \ObjectNavOpenVocab & Open vocab instance description: ``the brown vase painted orange with a bird on the side'' \\
            \hline
            \RoomNav & Type of room: ``bedroom'' \\
            \hline
        \end{tabular}%
    }
    \caption{\benchnav tasks. The full task specification also includes a navigation verb, such as ``Search for a vase''.}
    \vspace{-0.4cm}
    \label{tab:benchnav}
\end{table}

\subsection{Environments}
To overcome the challenges of scale and diversity, we leverage recent advances in the AI2-THOR simulated environment~\cite{Kolve2017AI2THORAI} which allow for importing any of the 800k 3D assets from the Objaverse dataset~\cite{Deitke2022ObjaverseAU} into AI2-THOR scenes. Of these 800k assets, we use a subset of ${\approx}$40k objects that have received additional annotations certifying their relevance to household environments and providing additional metadata (\eg object types grounded in the WordNet 2022 hierarchy~\cite{McCrae2019WordnetOpenSource}, instance descriptions, and size in meters). When paired with object instances already existing in AI2-THOR (${\approx}$2k instances), we are left with 41{,}133 unique 3D assets corresponding to \numsynsets unique object types (henceforth synonymous with Wordnet synsets).
The composition of the resulting collection of assets, in terms of their assigned synsets, is illustrated in Fig.~\ref{fig:sunburst}.\footnote{For this visualization, synsets have been further classified by GPT-3.5~\cite{Brown2020GPT, Ouyang2022InstructGPT} as belonging to one of 36 possible semantic clusters, in turn selected based on the whole collection of synsets.}

However, importing many new 3D assets into the 120 AI2-THOR scenes is insufficient for required diversity in scene layouts and may not result in meaningful object configurations. Instead, we use ProcTHOR~\cite{Deitke2022ProcTHORLE}, a procedural house generation framework built within AI2-THOR which can, in principle, generate an unbounded number of unique houses. We use ProcTHOR to generate a total of ${\approx}$200k houses (with between 1 and 8 rooms each), all containing Objaverse assets.
Assets are partitioned into train and eval instances, resulting in some object categories evaluating zero-shot performance due to the long tail of small-instance-count categories.
For more details see the supp.

\subsection{Expert Trajectories}

In order to produce our expert trajectories for imitation learning, we need planners capable of a range of skills essential for navigating and manipulating within complex multi-room settings. These planners must recognize and interact with objects, navigate through cluttered environments, and adapt to various obstacles.
Below, we offer a high-level overview of our planners. We are able to write these planners because of the wealth of ground truth information available within the AI2-THOR environment (\eg 3D coordinates of all objects); we stress that this ground-truth information is \textit{not} available to the agent at inference time and is simply used to produce the expert trajectories used for training. For further information, please refer to the supplementary materials.

\mypara{Navigation.} Given a target object or GPS coordinate within an environment, navigate to that target by following a shortest path computed via a navigation mesh. If the target was an object, rotate so that the object is approximately centered in the agent's navigation (or manipulation) camera. As our agent takes discrete actions (\eg ``move ahead by 0.2m''), the shortest path is followed approximately.

\mypara{Manipulation.} 
Given a target object instance, the agent first uses the privileged information from the simulation to navigate to a location from which the object is reachable by the arm. Then, as the poses of the object and the agent are known, we use iterative distance minimization to bring the arm close to the target object and then grasp that object.

\mypara{Room Visitation.} For our \SimpleExploreHouse task, the agent must visit every room in the house and issue sub-task completion signal. Since the layout of the house is known during trajectory generation in simulation, we can obtain the center of the rooms. For this task, we define a shortest-path planner that navigates to each room center via depth-first-search.

\section{Benchmark}
\label{sec:benchmark}

\begin{table*}[t!]
\centering
\resizebox{1\textwidth}{!}{
\begin{tabular}{ |l|l|l|>{\columncolor{SuccessColor}}c>{\columncolor{SELColor}}c>{\columncolor{\%RoomsColor}}c|>{\columncolor{SuccessColor}}c>{\columncolor{SELColor}}c>{\columncolor{\%RoomsColor}}c|>{\columncolor{SuccessColor}}c>{\columncolor{SELColor}}c>{\columncolor{\%RoomsColor}}c|>{\columncolor{SuccessColor}}c>{\columncolor{SELColor}}c>{\columncolor{\%RoomsColor}}c|>{\columncolor{SuccessColor}}c| }
\hline
\multirow{2}{*}{\textbf{Benchmark}} & \multirow{2}{*}{\textbf{Model}} & \multirow{2}{*}{\textbf{Training}} & \multicolumn{3}{c|}{\textbf{\ObjectNav}} & \multicolumn{3}{c|}{\textbf{\PickUp}} & \multicolumn{3}{c|}{\textbf{\Fetch}} & \multicolumn{3}{c|}{\textbf{\SimpleExploreHouse}} & Avg \\  
 & & & Success & SEL & \%Rooms & Success & SEL & \%Rooms & Success & SEL & \%Rooms & Success & SEL & \%Rooms & Success \\ \hline
\multirow{4}{*}{\bench\fifteen} & EmbSigLIP$^*$~\cite{Khandelwal2022EmbCLIP} & Single-task RL & 36.5 & 24.5 & 42.2 & 71.9 & 52.9 & 30.3 & 0.0 & 0.0 & 50.5 & 16.5 & 11.9 & 44.6 & 31.2 \\
& \model-1-task & Single-task IL & 57.0 & 46.2 & 51.5 & 84.2 & 81.0 & 30.3 & 15.1 & 12.6 & 48.1 & 43.7 & 40.4 & 81.2 & 50.0 \\
& \model & Multi-task IL & 55.0 & 42.2 & 56.3 & 90.1 & 86.9 & 30.3 & 14.0 & 10.5 & 49.3 & 40.5 & 35.7 & 81.1 & 49.9 \\
& \model w/ GT Det & Multi-task IL & 85.0 & 61.4 & 58.7 & 91.2 & 87.9 & 30.3 & 47.3 & 35.6 & 61.6 & 36.7 & 33.7 & 79.3 & 65.0 \\ \hline
\multirow{2}{*}{\bench\alltype} & \model & Multi-task IL & 33.7 & 25.1 & 53.7 & 75.1 & 69.1 & 31.5 & 10.6 & 8.1 & 42.9 & 35.0 & 33.2 & 77.8 & 38.6 \\
& \model w/ GT Det & Multi-task IL & 83.9 & 58.0 & 64.0 & 78.0 & 75.7 & 31.5 & 48.6 & 38.3 & 60.0 & 42.0 & 39.1 & 83.1 & 63.1 \\ \hline
\end{tabular}
}
\vspace{-0.5em}
\caption{\textbf{Training on single tasks, IL outperforms RL even with meticulous reward shaping.} EmbSigLIP refers to using the EmbCLIP~\cite{Khandelwal2022EmbCLIP} model with an upgrade to use the \siglip backbone since that hugely outperforms the ResNet-50 CLIP backbone (See Tab~\ref{tab:encoders}). Further, IL easily extends to multitask training without any performance degradation. Equipping the agent with detection massively boosts the success rate across all tasks except \SimpleExploreHouse which does not require navigating to or manipulating objects.}
\label{tab:main_four_tasks}
\vspace{-0.5em}
\end{table*}

\begin{table*}[t]
\centering
\resizebox{1\textwidth}{!}{
\begin{tabular}{ |l|>{\columncolor{SuccessColor}}c>{\columncolor{SELColor}}c>{\columncolor{\%RoomsColor}}c|>{\columncolor{SuccessColor}}c>{\columncolor{SELColor}}c>{\columncolor{\%RoomsColor}}c|>{\columncolor{SuccessColor}}c>{\columncolor{SELColor}}c>{\columncolor{\%RoomsColor}}c|>{\columncolor{SuccessColor}}c>{\columncolor{SELColor}}c>{\columncolor{\%RoomsColor}}c|>{\columncolor{SuccessColor}}c| }
\hline
\multirow{2}{*}{\textbf{Models}} & \multicolumn{3}{c|}{\textbf{\ObjectNav}} & \multicolumn{3}{c|}{\textbf{\PickUp}} & \multicolumn{3}{c|}{\textbf{\Fetch}} & \multicolumn{3}{c|}{\textbf{\SimpleExploreHouse}} & Avg \\  
 & Success & SEL & \%Rooms & Success & SEL & \%Rooms & Success & SEL & \%Rooms & Success & SEL & \%Rooms & Success \\ \hline
TxEnc + GRU & 44.7 & 33.8 & 47.7 & 84.8 & 81.4 & 30.3 & 10.5 & 9.0 & 41.8 & 34.5 & 31.8 & 72.6 & 43.6 \\
nonTxEnc + TxDec & 42.5 & 36.8 & 49.2 & 81.9 & 77.8 & 30.3 & 14.5 & 12.9 & 46.3 & 41.5 & 36.7 & 82.4 & 45.1 \\ 
TxEnc + TxDec \textcolor{gray}{(\model)} & 55.0 & 42.2 & 56.3 & 90.1 & 86.9 & 30.3 & 14.0 & 10.5 & 49.3 & 40.5 & 35.7 & 81.1 & 49.9 \\ \hline
\end{tabular}
}
\vspace{-0.5em}
\caption{\textbf{Swapping transformer encoder and decoder with alternative architectures.} GRU performs much worse than TxDec for long horizon Fetch and \SimpleExploreHouse tasks. TxEnc also has a clear advantage over EmbodiedCLIP~\cite{Khandelwal2022EmbCLIP}-style goal-conditioned visual feature extraction. All models use \siglip image and text encoders.}
\vspace{-0.5em}
\label{tab:arch}
\end{table*}

\begin{table*}[t!]
\centering
\resizebox{\textwidth}{!}{
\begin{tabular}{ |l|>{\columncolor{SuccessColor}}c>{\columncolor{SELColor}}c>{\columncolor{\%RoomsColor}}c|>{\columncolor{SuccessColor}}c>{\columncolor{SELColor}}c>{\columncolor{\%RoomsColor}}c|>{\columncolor{SuccessColor}}c>{\columncolor{SELColor}}c>{\columncolor{\%RoomsColor}}c|>{\columncolor{SuccessColor}}c>{\columncolor{SELColor}}c>{\columncolor{\%RoomsColor}}c|>{\columncolor{SuccessColor}}c| }
\hline
\multirow{2}{*}{\textbf{Image Encoder}} & \multicolumn{3}{c|}{\textbf{\ObjectNav}} & \multicolumn{3}{c|}{\textbf{\PickUp}} & \multicolumn{3}{c|}{\textbf{\Fetch}} & \multicolumn{3}{c|}{\textbf{\SimpleExploreHouse}} & Avg \\  
 & Success & SEL & \%Rooms & Success & SEL & \%Rooms & Success & SEL & \%Rooms & Success & SEL & \%Rooms & Success \\ \hline
\clip-RN50 & 19.6 & 12.1 & 44.1 & 64.1 & 60.2 & 30.5 & 1.8 & 0.8 & 43.4 & 21.0 & 19.5 & 62.6 & 26.6 \\
\dino-ViT-S/14 & 47.5 & 32.7 & 53.1 & 87.7 & 84.2 & 30.3 & 9.9 & 7.8 & 44.7 & 34.0 & 31.3 & 77.5 & 44.8 \\
\siglip-ViT-B/16 \textcolor{gray}{(\model)}& 55.0 & 42.2 & 56.3 & 90.1 & 86.9 & 30.3 & 14.0 & 10.5 & 49.3 & 40.5 & 35.7 & 81.1 & 49.9 \\ \hline
\end{tabular}
}
\vspace{-0.5em}
\caption{\textbf{Comparing different image encoders.} \siglip~\cite{Zhai2023SigLIP} significantly outperforms \clip-RN50~\cite{Khandelwal2022EmbCLIP} and \dino~\cite{oquab2023dinov2}.}
\label{tab:encoders}
\vspace{-0.5em}
\end{table*}

We evaluate \model on a new benchmark, \bench (\textbf{C}ore \textbf{HO}usehold \textbf{R}obot \textbf{E}valuation\textbf{S}). \bench consists of 4 task types (Tab.~\ref{tab:bench_tasks}) and is designed to evaluate how well the model can handle multiple tasks at once, which require skills including navigation, object recognition, object manipulation, and environment exploration. %

We also evaluate models on \benchnav (Tab.~\ref{tab:benchnav}), an extension of our benchmark which assesses the agent's ability to interpret and follow object navigation instructions that specify target objects in different ways. In addition to evaluating navigation and object recognition capabilities, \benchnav evaluates open vocabulary instruction following, object affordance understanding, scene understanding (\eg, ``on top of", ``in the kitchen"), and relative object-attributes comparison (\eg ``largest container").  For tasks like \ObjectNavRelAttr where comparison is needed, each environment has at least one other object of the same type that doesn't meet the condition. For instance, if the task is to find the ``smallest bowl'', there will be at least two bowls of different sizes in the same room. \ObjectNavAffordance, \ObjectNavRelAttr, and \ObjectNavLocalRef tasks may also sample WordNet hypernyms of scene synsets, \eg ``container'' that would be satisfied by ``vase'' and ``mug'' or ``sports equipment'' for ``basketball''.

To analyze models 
we first present results on a subset of 15 object categories from the full \numsynsets categories, called \fifteennodash. Evaluations on the full category set are named \alltypenodash. %
Our training data contains an average of 90k episodes per task and on average each task contains 195 episodes in the evaluation benchmark. 
Please see supplementary material.

\section{Experiments}
\label{sec:experiments}

\begin{table*}[h!]
\vspace{-0.5em}
\centering
\resizebox{1\textwidth}{!}{
\begin{tabular}{ |l|>{\columncolor{SuccessColor}}c>{\columncolor{SELColor}}c>{\columncolor{\%RoomsColor}}c|>{\columncolor{SuccessColor}}c>{\columncolor{SELColor}}c>{\columncolor{\%RoomsColor}}c|>{\columncolor{SuccessColor}}c>{\columncolor{SELColor}}c>{\columncolor{\%RoomsColor}}c|>{\columncolor{SuccessColor}}c>{\columncolor{SELColor}}c>{\columncolor{\%RoomsColor}}c|>{\columncolor{SuccessColor}}c| }
\hline
\multirow{2}{*}{\textbf{Window Size}} & \multicolumn{3}{c|}{\textbf{\ObjectNav}} & \multicolumn{3}{c|}{\textbf{\PickUp}} & \multicolumn{3}{c|}{\textbf{\Fetch}} & \multicolumn{3}{c|}{\textbf{\SimpleExploreHouse}} & Avg \\  
 & Success & SEL & \%Rooms & Success & SEL & \%Rooms & Success & SEL & \%Rooms & Success & SEL & \%Rooms & Success \\ \hline
10 & 34.0 & 27.5 & 46.3 & 56.7 & 53.1 & 30.3 & 2.3 & 2.1 & 50.6 & 18.0 & 16.0 & 56.1 & 27.8 \\
50 & 40.5 & 30.6 & 48.4 & 87.1 & 83.5 & 30.3 & 4.1 & 3.8 & 37.7 & 28.0 & 25.2 & 70.3 & 39.9 \\
100 \textcolor{gray}{(\model)} & 55.0 & 42.2 & 56.3 & 90.1 & 86.9 & 30.3 & 14.0 & 10.5 & 49.3 & 40.5 & 35.7 & 81.1 & 49.9 \\ \hline
\end{tabular}
}
\vspace{-0.5em}
\caption{\textbf{Effect of context window.} Longer context is essential particular for long-horizon tasks like \Fetch and \SimpleExploreHouse.}
\label{tab:window}
\end{table*}

\begin{table*}[t!]
    \centering
    \vspace{-0.5em}
    \begin{subfigure}[t]{0.3\textwidth}
        \centering
        \resizebox{0.9\columnwidth}{!}{
        \begin{tabular}{ |l|>{\columncolor{SuccessColor}}c>{\columncolor{SELColor}}c>{\columncolor{\%RoomsColor}}c| }
        \hline
        \multirow{2}{*}{\textbf{Training Eps.}} & \multicolumn{3}{c|}{\textbf{\ObjectNav}} \\  
         & Success & SEL & \%Rooms \\ \hline
        1k & 19.0 & 14.3 & 47.6 \\
        10k & 39.0 & 31.1 & 52.9 \\
        100k \textcolor{gray}{(\model-1-task)} & 57.0 & 46.2 & 51.5 \\ \hline
        \end{tabular}
        }
        \caption{}
        \label{tab:scale}
    \end{subfigure}%
    ~ 
    \begin{subfigure}[t]{0.3\textwidth}
        \centering
        \resizebox{0.9\columnwidth}{!}{
        \begin{tabular}{ |l|>{\columncolor{SuccessColor}}c>{\columncolor{SELColor}}c>{\columncolor{\%RoomsColor}}c| }
        \hline
        \multirow{2}{*}{\textbf{Houses}} & \multicolumn{3}{c|}{\textbf{\ObjectNav}} \\  
         & Success & SEL & \%Rooms \\ \hline
        100 & 43.5 & 35.2 & 53.6 \\
        10k \textcolor{gray}{(\model-1-task)} & 57.0 & 46.2 & 51.5 \\ \hline
        \end{tabular}
        }
        \caption{}
        \label{tab:diversity}
    \end{subfigure}
    ~
    \begin{subfigure}[t]{0.3\textwidth}
        \centering
        \resizebox{0.9\columnwidth}{!}{
        \begin{tabular}{ |l|>{\columncolor{SuccessColor}}c>{\columncolor{SELColor}}c>{\columncolor{\%RoomsColor}}c| }
        \hline
        \multirow{2}{*}{\textbf{Expert}} & \multicolumn{3}{c|}{\textbf{\ObjectNav}} \\  
         & Success & SEL & \%Rooms \\ \hline
        Explore \ObjectNav & 46.5 & 27.9 & 47.7 \\ 
        \ObjectNav \textcolor{gray}{(\model-1-task)} & 57.0 & 46.2 & 51.5 \\ \hline
        \end{tabular}
        }
        \caption{}
        \label{tab:ON_vs_exploreON}
    \end{subfigure}
    \vspace{-1em}
    \caption{\textbf{Effect of scale, house diversity, and choice of experts.} (a) Performance increases with more training episodes; (b) With the same number of episodes (100k), increasing diversity of houses boosts performance; (c) Training with experts that explore until the object becomes visible provides no gains.}
    \vspace{-1em}
\end{table*}

\begin{figure*}[h]
    \centering
    \includegraphics[width=0.95\linewidth, clip]{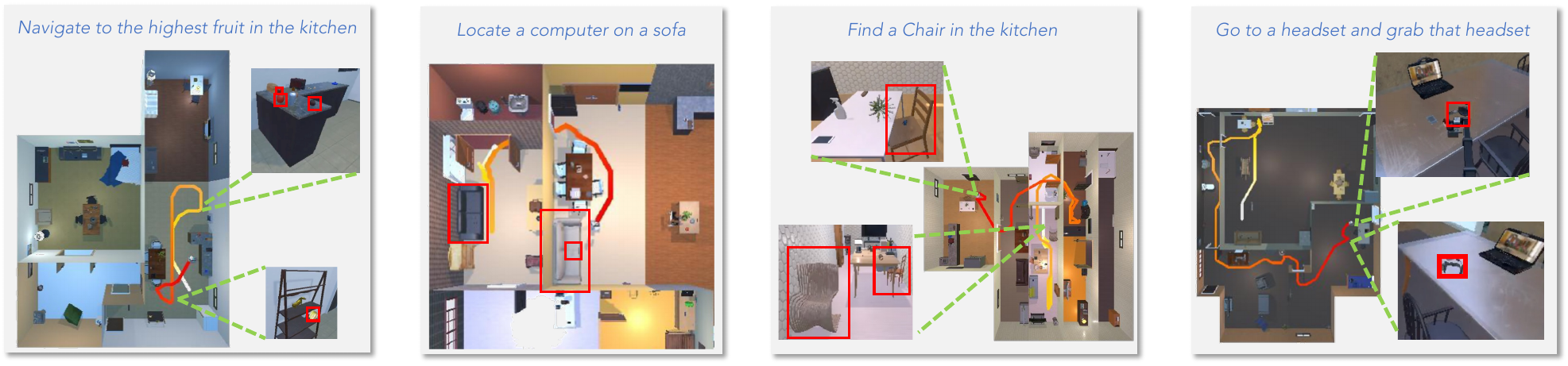}
    \vspace{-1em}
    \caption{\textbf{\modelb's Behavior.}  
    L to R: the first image depicts the agent navigating to all possible fruits in the kitchen and then going back to the highest located fruit; the second shows the agent scanning a sofa, then moving to another sofa and finally ending the episode when it sees the laptop; the third illustrates the agent skipping chairs in the living room to reach one in the kitchen; and the fourth demonstrates the agent visiting the headset, but then repositioning itself around the table to access a location where the headset is reachable.}
    \label{fig:havior}
\end{figure*}

\begin{table*}[h]
\centering
\small
\resizebox{\textwidth}{!}{
\begin{tabular}{ |l|>{\columncolor{SuccessColor}}c>{\columncolor{\%RoomsColor}}c|>{\columncolor{SuccessColor}}c>{\columncolor{\%RoomsColor}}c|>{\columncolor{SuccessColor}}c>{\columncolor{\%RoomsColor}}c|>{\columncolor{SuccessColor}}c>{\columncolor{\%RoomsColor}}c|>{\columncolor{SuccessColor}}c>{\columncolor{\%RoomsColor}}c|>{\columncolor{SuccessColor}}c>{\columncolor{\%RoomsColor}}c|>{\columncolor{SuccessColor}}c>{\columncolor{\%RoomsColor}}c|>{\columncolor{SuccessColor}}c| }
\hline
\multirow{2}{*}{\textbf{Benchmark}} & \multicolumn{2}{c|}{\textbf{\ObjectNav}} & \multicolumn{2}{c|}{\textbf{\ObjectNavRoom}} & \multicolumn{2}{c|}{\textbf{\ObjectNavRelAttr}} & \multicolumn{2}{c|}{\textbf{\ObjectNavAffordance}} & \multicolumn{2}{c|}{\textbf{\ObjectNavLocalRef}} & \multicolumn{2}{c|}{\textbf{\ObjectNavOpenVocab}} & \multicolumn{2}{c|}{\textbf{\RoomNav}} & Avg \\  
 & Success & \%Rooms & Success & \%Rooms & Success & \%Rooms & Success & \%Rooms & Success & \%Rooms & Success & \%Rooms & Success & \%Rooms & Success \\ \hline
\benchnav\fifteen & 57.5 & 55.7 & 50.3 & 54.6 & 54.6 & 62.2 & 62.4 & 53.0 & 45.1 & 51.5 & 30.6& 49.9 & 74.5 & 48.1 & 53.6 \\ \hline
\benchnav\alltype & 38.7 & 53.4 & 54.2 & 55.7 & 38.5 & 56.0 & 43.5 & 48.0 & 44.5 & 58.7 & 30.5 & 56.8 & 67.5 & 49.9 & 45.3 \\ \hline
\end{tabular}
}
\caption{\benchnav results to evaluate \model's ability to handle diverse target specifications for navigation.}
\label{tab:on_variations}
\vspace{-1em}
\end{table*}

\noindent \textbf{Implementation details.}
\model uses \siglip image and text encoders. We use $3$-layer transformer encoder and decoder and a context window of $100$. All models are trained with batch size=$224$, AdamW and LR=$0.0002$. Single-task models and multi-task models are trained for 20k and 50k iterations, respectively. Using $16$-bit mixed precision training \model trains at 
an FPS of ${\approx}$3500, compared to an FPS of ${\approx}$175 for RL implemented using AllenAct~\cite{AllenAct}. We find that data augmentation both during training and testing is critical for model performance, both in simulation and real.
In simulation, the PickUp action succeeds if the object is within 6cm of the gripper. In the real world, we leverage a heuristic object grasping model which is called when \model invokes PickUp. See supplementary for more details.

\subsection{Quantitative Analysis}
We compare single and multitask versions of \model against single-task RL baselines based on EmbCLIP~\cite{Khandelwal2022EmbCLIP} on \bench{}\fifteen and \model's ability to handle large object vocabulary on \bench{}\alltype (Tab.~\ref{tab:main_four_tasks}). 
We thoroughly investigate design decisions like architecture choices (Tab.~\ref{tab:arch}), image encoders (Tab.~\ref{tab:encoders}), context window size (Tab.~\ref{tab:window}), choice of experts (Tab.~\ref{tab:ON_vs_exploreON}), and demonstrate the importance of scale (Tab.~\ref{tab:scale}) and diversity of environments (Tab.~\ref{tab:diversity}). To assess the instruction following capabilities of \model, we evaluate on \benchnav (Tab.~\ref{tab:on_variations}). We report Success rate, Episode-length weighted Success
(SEL\footnote{We report SEL instead of the prevalent SPL metric due to the known limitations of SPL, see~\cite{eftekhar2023selective}, and because weighting by episode length is more informative than path length for tasks that include manipulation.}~\cite{eftekhar2023selective}), and percentage rooms visited (\%Rooms) for each task and the average Success across all tasks. We now discuss our findings.

\noindent \textbf{IL on shortest-path episodes at scale produces highly capable agents.} Comparing rows 1 and 2 of Tab.~\ref{tab:main_four_tasks}, we see that our IL-trained \model dramatically outperforms the popular RL-trained EmbCLIP architecture~\cite{Khandelwal2022EmbCLIP} across all \bench tasks. Note that the RL baselines were upgraded to use the \siglip visual backbone to be comparable to \model, required extensive reward shaping, were run on the same hardware as \model, but for 2x the number of hours. Indeed, despite extensive efforts, we were unable to obtain non-zero performance on the \Fetch task using RL.

\noindent \textbf{\modelb can multitask.} 
Comparing Tab{.}~\ref{tab:main_four_tasks} rows 2-3, \bench{}\fifteen multitask IL performance ($49.9\%$) matches single-task IL ($50\%$). This suggests an absence of performance degradation due to task-competition traditionally seen during multitask training.

\noindent \textbf{Detection brings huge gains.} Our RGB-only \model agent learns to navigate well, and an error analysis reveals that the majority of failures arise from perception problems.  
As seen in Tab.~\ref{tab:main_four_tasks}, compare rows 3-4 and 5-6, \model with ground truth detection (provided by the simulator) shows $15\%$ absolute average success rate gain across all tasks on \bench{}\fifteen and an even larger $24.5\%$ absolute gain on \bench{}\alltype where the detection problem is harder due to the larger object vocabulary. The gains are more prominent for \ObjectNav, which obtains an impressive 85\%, and \Fetch which both require localizing the target object. Tasks like \PickUp (where the agent begins facing the object) and \SimpleExploreHouse (no target object) show little gains as expected. These results indicate that IL trained agents can continue to improve significantly with better object perception.

\noindent \textbf{Transformers provide gains at encoding and decoding.} In Tab.~\ref{tab:arch}, we compare \model with other commonly used architectural choices for Embodied agents. First, we adapted EmbCLIP's goal-conditioned visual encoder~\cite{Khandelwal2022EmbCLIP} to input 2 RGB frames and upgraded its image encoder to SigLIP to create a non-Transformer visual encoder (nonTxEnc). Replacing \model's Transformer-based visual encoder (TxEnc) with nonTxEnc resulted in a performance drop of $4.8$ points showing the superiority of TxEnc architecture for extracting relevant visual information. Swapping the action decoder (TxDec) with GRU showed an even larger drop of $6.3$ points. We hypothesize TxDec outperforms GRU because of the ability to attend over observations and actions several 100 steps in the past while GRUs struggle with compressing history in a single history embedding.

\noindent \textbf{Strong image encoders produce strong agents.} Recent advances in image representations such as \dino and \siglip directly translate to gains in Embodied tasks. \siglip particularly nearly doubles the average success rate of \clip on \bench{}\fifteen (Tab.~\ref{tab:encoders}). Interestingly, while self-supervised \dino significantly outperforms \clip, it lags behind \siglip which is trained for image-text matching.

\noindent \textbf{Long horizon tasks require long context windows.} Transformer based embodied agents in the literature often rely on short context lengths to encode history due to compute constraints (\eg RT-1~\cite{Brohan2023RT1} uses 6 previous frames). We find that short context windows are detrimental to performance on longer horizon tasks like \Fetch and \SimpleExploreHouse (Tab.~\ref{tab:window}). Note that we train \model with limited context length but use all past observations for inference. 

\noindent \textbf{Scale and diversity of training data matters.} In Tab.~\ref{tab:scale}, we show that performance on \ObjectNav steadily increases with number of training episodes. This raises a question, is it sufficient to collect a large number of samples from a limited number of houses? To answer this, we create two training sets with different numbers of houses - 100 houses with 1000 episodes each, and 10k houses with 10 episodes each. \model trained on the latter shows an absolute gain of $13.5\%$ (Tab.~\ref{tab:diversity}). We believe that lack of house diversity in prior studies like PIRLNav~\cite{ramrakhya2023pirlnav} (which used 120 scenes) may have contributed 
to the inefficacy of IL on shortest-path trajectories for tasks like \ObjectNav.

\noindent \textbf{Exploration based planners provide no gain.} Prior work~\cite{ramrakhya2023pirlnav} found that IL with frontier exploration trajectories outperformed shortest-path trajectories. In Tab.~\ref{tab:ON_vs_exploreON} we compare 1-task \model to a variant trained with episodes generated by an exploration-based planner (see supplement for details)
and find no gains using the latter. This reinforces the finding in Tab.~\ref{tab:diversity} that the diversity of training environments is critical to the success of IL on shortest path trajectories.

\noindent \textbf{\modelb follows open vocabulary instructions.} Tab.~\ref{tab:on_variations} shows the performance of \model on several navigation tasks that require it to follow long instructions, disambiguate attributes, and understand relative distances.
High performance on \ObjectNavAffordance shows the ability of \model to understand instructions such as \emph{locate an edible fruit that can best be used as a guacamole ingredient} or \emph{find a tool that can best be used for cutting fruits and vegetables}, when there are multiple tools and edible fruits in the house. A high Success rate on \ObjectNavRoom for \bench{}\alltype compared to \ObjectNav show that locating a large vocabulary of objects is indeed easier when the agent is told which room the object lies in, and the high performance on \RoomNav confirms that the agent has learned to identify rooms. 

\noindent \textbf{\modelb transfers effectively to the real world.} 
To assess real-world generalization with no visual adaptation and no real world finetuning, we evaluate two of our best models in physical environments. These models were tested across 88 trials in two different real-world settings. Table~\ref{tab:real_world}, row 1 shows the performance of the RGB only \model. Row 2 is the performance of \model trained with GT Detection but evaluated in the real world with a DETIC object detector~\cite{Zhou2022DETIC}. Comparing Table~\ref{tab:real_world} with rows 3 and 4 from Table~\ref{tab:main_four_tasks} shows that the performance drop between simulation and real is small overall and minimal for the navigation tasks. For manipulation tasks, note that the numbers in parentheses measure Soft Success, i.e. the model is rewarded if the gripper is within 6cm of the object, regardless of whether the heuristic grasping is successful. The Soft Success numbers are very similar to the simulation results indicating that the transfer of the learned policy from sim to real is very effective.
\begin{table}[H]
\centering
\vspace{-0.6em}
\resizebox{\columnwidth}{!}{
\begin{tabular}{ |l|>{\columncolor{SuccessColor}}c|>{\columncolor{SuccessColor}}c|>{\columncolor{SuccessColor}}c|>{\columncolor{SuccessColor}}c|>{\columncolor{SuccessColor}}c| }
\hline
\textbf{Model} & \ObjectNav & \PickUp & \Fetch & \SimpleExploreHouse & Average \\  \hline
\model & 50.0 & 46.7 \textcolor{gray}{(66.7)} & 11.1 \textcolor{gray}{(33.3)} & 50.0 & 39.5\\
\model w/ \detic & 83.3 & 46.7 \textcolor{gray}{(86.7)} & 44.4 \textcolor{gray}{(44.4)} & 50.0 & 56.1 \\ \hline
\end{tabular}
}
\vspace{-0.6em}
\caption{\textbf{Real world results}. Parenthetical numbers on manip{.} tasks indicate Soft Success: \model called PickUp sufficiently near the target, ignores heuristic grasping success/failure.}
\label{tab:real_world}
\vspace{-1em}
\end{table}

\subsection{Agent Behavior}

Figure~\ref{fig:agent_behavior} presents our qualitative examples, highlighting several intriguing behaviors exhibited by \model. Figure~\ref{fig:teaser} illustrates additional qualitative trajectories in both simulated and real-world environments. These examples emphasize the model's capabilities in exploration, backtracking, scene and spatial comprehension, and instruction following. For more examples, please refer to the supplementary material.

\section{Conclusion}
\label{sec:conclusion}

In this work, we explore the potential of imitation learning for learning Embodied policies. Using shortest path expert planners in procedurally generated environments, it is now finally possible to generate training data at the scale and diversity needed to make techniques like Behavior Cloning work. More importantly, we show that agents learned by cloning experts in simulation not only generalize to novel environments but also to the real world.

{
    \small
    \bibliographystyle{ieeenat_fullname}
    \bibliography{main}

\begin{thebibliography}{89}
\providecommand{\natexlab}[1]{#1}
\providecommand{\url}[1]{\texttt{#1}}
\expandafter\ifx\csname urlstyle\endcsname\relax
  \providecommand{\doi}[1]{doi: #1}\else
  \providecommand{\doi}{doi: \begingroup \urlstyle{rm}\Url}\fi

\bibitem[Anderson et~al.(2018)Anderson, Wu, Teney, Bruce, Johnson,
  S{\"{u}}nderhauf, Reid, Gould, and van~den Hengel]{Anderson2018R2R}
Peter Anderson, Qi Wu, Damien Teney, Jake Bruce, Mark Johnson, Niko
  S{\"{u}}nderhauf, Ian~D. Reid, Stephen Gould, and Anton van~den Hengel.
\newblock Vision-and-language navigation: Interpreting visually-grounded
  navigation instructions in real environments.
\newblock In \emph{2018 {IEEE} Conference on Computer Vision and Pattern
  Recognition, {CVPR} 2018, Salt Lake City, UT, USA, June 18-22, 2018}, pages
  3674--3683. Computer Vision Foundation / {IEEE} Computer Society, 2018.

\bibitem[Batra et~al.(2020)Batra, Gokaslan, Kembhavi, Maksymets, Mottaghi,
  Savva, Toshev, and Wijmans]{Batra2020ObjectNav}
Dhruv Batra, Aaron Gokaslan, Aniruddha Kembhavi, Oleksandr Maksymets, Roozbeh
  Mottaghi, Manolis Savva, Alexander Toshev, and Erik Wijmans.
\newblock Objectnav revisited: On evaluation of embodied agents navigating to
  objects.
\newblock \emph{CoRR}, abs/2006.13171, 2020.

\bibitem[Brohan et~al.(2023{\natexlab{a}})Brohan, Brown, Carbajal, Chebotar,
  Chen, Choromanski, Ding, Driess, Dubey, Finn, Florence, Fu, Arenas,
  Gopalakrishnan, Han, Hausman, Herzog, Hsu, Ichter, Irpan, Joshi, Julian,
  Kalashnikov, Kuang, Leal, Lee, Lee, Levine, Lu, Michalewski, Mordatch,
  Pertsch, Rao, Reymann, Ryoo, Salazar, Sanketi, Sermanet, Singh, Singh,
  Soricut, Tran, Vanhoucke, Vuong, Wahid, Welker, Wohlhart, Wu, Xia, Xiao, Xu,
  Xu, Yu, and Zitkovich]{Brohan2023RT2}
Anthony Brohan, Noah Brown, Justice Carbajal, Yevgen Chebotar, Xi Chen,
  Krzysztof Choromanski, Tianli Ding, Danny Driess, Avinava Dubey, Chelsea
  Finn, Pete Florence, Chuyuan Fu, Montse~Gonzalez Arenas, Keerthana
  Gopalakrishnan, Kehang Han, Karol Hausman, Alexander Herzog, Jasmine Hsu,
  Brian Ichter, Alex Irpan, Nikhil~J. Joshi, Ryan Julian, Dmitry Kalashnikov,
  Yuheng Kuang, Isabel Leal, Lisa Lee, Tsang{-}Wei~Edward Lee, Sergey Levine,
  Yao Lu, Henryk Michalewski, Igor Mordatch, Karl Pertsch, Kanishka Rao, Krista
  Reymann, Michael~S. Ryoo, Grecia Salazar, Pannag Sanketi, Pierre Sermanet,
  Jaspiar Singh, Anikait Singh, Radu Soricut, Huong~T. Tran, Vincent Vanhoucke,
  Quan Vuong, Ayzaan Wahid, Stefan Welker, Paul Wohlhart, Jialin Wu, Fei Xia,
  Ted Xiao, Peng Xu, Sichun Xu, Tianhe Yu, and Brianna Zitkovich.
\newblock {RT-2:} vision-language-action models transfer web knowledge to
  robotic control.
\newblock \emph{CoRR}, abs/2307.15818, 2023{\natexlab{a}}.

\bibitem[Brohan et~al.(2023{\natexlab{b}})Brohan, Brown, Carbajal, Chebotar,
  Dabis, Finn, Gopalakrishnan, Hausman, Herzog, Hsu, Ibarz, Ichter, Irpan,
  Jackson, Jesmonth, Joshi, Julian, Kalashnikov, Kuang, Leal, Lee, Levine, Lu,
  Malla, Manjunath, Mordatch, Nachum, Parada, Peralta, Perez, Pertsch,
  Quiambao, Rao, Ryoo, Salazar, Sanketi, Sayed, Singh, Sontakke, Stone, Tan,
  Tran, Vanhoucke, Vega, Vuong, Xia, Xiao, Xu, Xu, Yu, and
  Zitkovich]{Brohan2023RT1}
Anthony Brohan, Noah Brown, Justice Carbajal, Yevgen Chebotar, Joseph Dabis,
  Chelsea Finn, Keerthana Gopalakrishnan, Karol Hausman, Alexander Herzog,
  Jasmine Hsu, Julian Ibarz, Brian Ichter, Alex Irpan, Tomas Jackson, Sally
  Jesmonth, Nikhil~J. Joshi, Ryan Julian, Dmitry Kalashnikov, Yuheng Kuang,
  Isabel Leal, Kuang{-}Huei Lee, Sergey Levine, Yao Lu, Utsav Malla, Deeksha
  Manjunath, Igor Mordatch, Ofir Nachum, Carolina Parada, Jodilyn Peralta,
  Emily Perez, Karl Pertsch, Jornell Quiambao, Kanishka Rao, Michael~S. Ryoo,
  Grecia Salazar, Pannag~R. Sanketi, Kevin Sayed, Jaspiar Singh, Sumedh
  Sontakke, Austin Stone, Clayton Tan, Huong~T. Tran, Vincent Vanhoucke, Steve
  Vega, Quan Vuong, Fei Xia, Ted Xiao, Peng Xu, Sichun Xu, Tianhe Yu, and
  Brianna Zitkovich.
\newblock {RT-1:} robotics transformer for real-world control at scale.
\newblock In \emph{Robotics: Science and Systems XIX, Daegu, Republic of Korea,
  July 10-14, 2023}, 2023{\natexlab{b}}.

\bibitem[Brown et~al.(2020)Brown, Mann, Ryder, Subbiah, Kaplan, Dhariwal,
  Neelakantan, Shyam, Sastry, Askell, Agarwal, Herbert{-}Voss, Krueger,
  Henighan, Child, Ramesh, Ziegler, Wu, Winter, Hesse, Chen, Sigler, Litwin,
  Gray, Chess, Clark, Berner, McCandlish, Radford, Sutskever, and
  Amodei]{Brown2020GPT}
Tom~B. Brown, Benjamin Mann, Nick Ryder, Melanie Subbiah, Jared Kaplan,
  Prafulla Dhariwal, Arvind Neelakantan, Pranav Shyam, Girish Sastry, Amanda
  Askell, Sandhini Agarwal, Ariel Herbert{-}Voss, Gretchen Krueger, Tom
  Henighan, Rewon Child, Aditya Ramesh, Daniel~M. Ziegler, Jeffrey Wu, Clemens
  Winter, Christopher Hesse, Mark Chen, Eric Sigler, Mateusz Litwin, Scott
  Gray, Benjamin Chess, Jack Clark, Christopher Berner, Sam McCandlish, Alec
  Radford, Ilya Sutskever, and Dario Amodei.
\newblock Language models are few-shot learners.
\newblock In \emph{Advances in Neural Information Processing Systems 33: Annual
  Conference on Neural Information Processing Systems 2020, NeurIPS 2020,
  December 6-12, 2020, virtual}, 2020.

\bibitem[Chang et~al.(2017)Chang, Dai, Funkhouser, Halber, Niessner, Savva,
  Song, Zeng, and Zhang]{Matterport3D}
Angel Chang, Angela Dai, Thomas Funkhouser, Maciej Halber, Matthias Niessner,
  Manolis Savva, Shuran Song, Andy Zeng, and Yinda Zhang.
\newblock Matterport3d: Learning from rgb-d data in indoor environments.
\newblock \emph{International Conference on 3D Vision (3DV)}, 2017.

\bibitem[Chang et~al.(2023)Chang, Gervet, Khanna, Yenamandra, Shah, Min, Shah,
  Paxton, Gupta, Batra, et~al.]{chang2023goat}
Matthew Chang, Theophile Gervet, Mukul Khanna, Sriram Yenamandra, Dhruv Shah,
  So~Yeon Min, Kavit Shah, Chris Paxton, Saurabh Gupta, Dhruv Batra, et~al.
\newblock Goat: Go to any thing.
\newblock \emph{arXiv preprint arXiv:2311.06430}, 2023.

\bibitem[Chaplot et~al.(2020{\natexlab{a}})Chaplot, Gandhi, Gupta, Gupta, and
  Salakhutdinov]{chaplot2020learning}
Devendra~Singh Chaplot, Dhiraj Gandhi, Saurabh Gupta, Abhinav Gupta, and Ruslan
  Salakhutdinov.
\newblock Learning to explore using active neural slam.
\newblock \emph{ICLR}, 2020{\natexlab{a}}.

\bibitem[Chaplot et~al.(2020{\natexlab{b}})Chaplot, Salakhutdinov, Gupta, and
  Gupta]{chaplot2020neural}
Devendra~Singh Chaplot, Ruslan Salakhutdinov, Abhinav Gupta, and Saurabh Gupta.
\newblock Neural topological slam for visual navigation.
\newblock In \emph{Proceedings of the IEEE/CVF Conference on Computer Vision
  and Pattern Recognition}, 2020{\natexlab{b}}.

\bibitem[Chen et~al.(2020)Chen, Jain, Schissler, Gari, Al{-}Halah, Ithapu,
  Robinson, and Grauman]{Chen2020SoundSpaces}
Changan Chen, Unnat Jain, Carl Schissler, Sebastia Vicenc~Amengual Gari, Ziad
  Al{-}Halah, Vamsi~Krishna Ithapu, Philip~W. Robinson, and Kristen Grauman.
\newblock {SoundSpaces: Audio-Visual Navigation in 3D Environments}.
\newblock In \emph{Computer Vision - {ECCV} 2020 - 16th European Conference,
  Glasgow, UK, August 23-28, 2020, Proceedings, Part {VI}}, pages 17--36.
  Springer, 2020.

\bibitem[Chen et~al.(2022{\natexlab{a}})Chen, Schissler, Garg, Kobernik, Clegg,
  Calamia, Batra, Robinson, and Grauman]{Chen2022SoundSpaces2}
Changan Chen, Carl Schissler, Sanchit Garg, Philip Kobernik, Alexander Clegg,
  Paul Calamia, Dhruv Batra, Philip~W. Robinson, and Kristen Grauman.
\newblock {SoundSpaces 2.0: {A} Simulation Platform for Visual-Acoustic
  Learning}.
\newblock In \emph{NeurIPS}, 2022{\natexlab{a}}.

\bibitem[Chen et~al.(2021)Chen, Lu, Rajeswaran, Lee, Grover, Laskin, Abbeel,
  Srinivas, and Mordatch]{Chen2021DecisionTransformers}
Lili Chen, Kevin Lu, Aravind Rajeswaran, Kimin Lee, Aditya Grover, Michael
  Laskin, Pieter Abbeel, Aravind Srinivas, and Igor Mordatch.
\newblock Decision transformer: Reinforcement learning via sequence modeling.
\newblock In \emph{Advances in Neural Information Processing Systems 34: Annual
  Conference on Neural Information Processing Systems 2021, NeurIPS 2021,
  December 6-14, 2021, virtual}, pages 15084--15097, 2021.

\bibitem[Chen et~al.(2022{\natexlab{b}})Chen, Hu, Jin, Li, and
  Wang]{Chen2022DomainRand}
Xiaoyu Chen, Jiachen Hu, Chi Jin, Lihong Li, and Liwei Wang.
\newblock {Understanding Domain Randomization for Sim-to-real Transfer}.
\newblock In \emph{The Tenth International Conference on Learning
  Representations, {ICLR} 2022, Virtual Event, April 25-29, 2022}.
  OpenReview.net, 2022{\natexlab{b}}.

\bibitem[Codevilla et~al.(2017)Codevilla, M{\"u}ller, Dosovitskiy, L{\'o}pez,
  and Koltun]{Codevilla2017EndtoEndDV}
Felipe Codevilla, Matthias M{\"u}ller, Alexey Dosovitskiy, Antonio~M.
  L{\'o}pez, and Vladlen Koltun.
\newblock End-to-end driving via conditional imitation learning.
\newblock \emph{2018 IEEE International Conference on Robotics and Automation
  (ICRA)}, pages 1--9, 2017.

\bibitem[Dasgupta et~al.(2023)Dasgupta, Kaeser-Chen, Marino, Ahuja, Babayan,
  Hill, and Fergus]{Dasgupta2023CollaboratingWL}
Ishita Dasgupta, Christine Kaeser-Chen, Kenneth Marino, Arun Ahuja, Sheila
  Babayan, Felix Hill, and Rob Fergus.
\newblock Collaborating with language models for embodied reasoning.
\newblock \emph{ArXiv}, abs/2302.00763, 2023.

\bibitem[Deitke et~al.(2020)Deitke, Han, Herrasti, Kembhavi, Kolve, Mottaghi,
  Salvador, Schwenk, VanderBilt, Wallingford, Weihs, Yatskar, and
  Farhadi]{Deitke2020RoboTHOR}
Matt Deitke, Winson Han, Alvaro Herrasti, Aniruddha Kembhavi, Eric Kolve,
  Roozbeh Mottaghi, Jordi Salvador, Dustin Schwenk, Eli VanderBilt, Matthew
  Wallingford, Luca Weihs, Mark Yatskar, and Ali Farhadi.
\newblock {RoboTHOR: An Open Simulation-to-Real Embodied {AI} Platform}.
\newblock In \emph{2020 {IEEE/CVF} Conference on Computer Vision and Pattern
  Recognition, {CVPR} 2020, Seattle, WA, USA, June 13-19, 2020}, pages
  3161--3171. Computer Vision Foundation / {IEEE}, 2020.

\bibitem[Deitke et~al.(2022{\natexlab{a}})Deitke, Schwenk, Salvador, Weihs,
  Michel, VanderBilt, Schmidt, Ehsani, Kembhavi, and
  Farhadi]{Deitke2022ObjaverseAU}
Matt Deitke, Dustin Schwenk, Jordi Salvador, Luca Weihs, Oscar Michel, Eli
  VanderBilt, Ludwig Schmidt, Kiana Ehsani, Aniruddha Kembhavi, and Ali
  Farhadi.
\newblock {Objaverse: A Universe of Annotated 3D Objects}.
\newblock \emph{2023 IEEE/CVF Conference on Computer Vision and Pattern
  Recognition (CVPR)}, pages 13142--13153, 2022{\natexlab{a}}.

\bibitem[Deitke et~al.(2022{\natexlab{b}})Deitke, VanderBilt, Herrasti, Weihs,
  Ehsani, Salvador, Han, Kolve, Kembhavi, and Mottaghi]{Deitke2022ProcTHORLE}
Matt Deitke, Eli VanderBilt, Alvaro Herrasti, Luca Weihs, Kiana Ehsani, Jordi
  Salvador, Winson Han, Eric Kolve, Aniruddha Kembhavi, and Roozbeh Mottaghi.
\newblock Procthor: Large-scale embodied {AI} using procedural generation.
\newblock In \emph{NeurIPS}, 2022{\natexlab{b}}.

\bibitem[Deitke et~al.(2023)Deitke, Hendrix, Farhadi, Ehsani, and
  Kembhavi]{Deitke2023Phone2Proc}
Matt Deitke, Rose Hendrix, Ali Farhadi, Kiana Ehsani, and Aniruddha Kembhavi.
\newblock {Phone2Proc: Bringing Robust Robots into Our Chaotic World}.
\newblock In \emph{{IEEE/CVF} Conference on Computer Vision and Pattern
  Recognition, {CVPR} 2023, Vancouver, BC, Canada, June 17-24, 2023}, pages
  9665--9675. {IEEE}, 2023.

\bibitem[Eftekhar et~al.(2023)Eftekhar, Zeng, Duan, Farhadi, Kembhavi, and
  Krishna]{eftekhar2023selective}
Ainaz Eftekhar, Kuo-Hao Zeng, Jiafei Duan, Ali Farhadi, Ani Kembhavi, and
  Ranjay Krishna.
\newblock Selective visual representations improve convergence and
  generalization for embodied ai.
\newblock \emph{arXiv preprint arXiv:2311.04193}, 2023.

\bibitem[Ehsani et~al.(2021)Ehsani, Han, Herrasti, VanderBilt, Weihs, Kolve,
  Kembhavi, and Mottaghi]{Ehsani2021ManipulaTHOR}
Kiana Ehsani, Winson Han, Alvaro Herrasti, Eli VanderBilt, Luca Weihs, Eric
  Kolve, Aniruddha Kembhavi, and Roozbeh Mottaghi.
\newblock Manipulathor: {A} framework for visual object manipulation.
\newblock In \emph{{IEEE} Conference on Computer Vision and Pattern
  Recognition, {CVPR} 2021, virtual, June 19-25, 2021}, pages 4497--4506.
  Computer Vision Foundation / {IEEE}, 2021.

\bibitem[Ehsani et~al.(2022)Ehsani, Farhadi, Kembhavi, and
  Mottaghi]{Ehsani2022ObjDis}
Kiana Ehsani, Ali Farhadi, Aniruddha Kembhavi, and Roozbeh Mottaghi.
\newblock Object manipulation via visual target localization.
\newblock In \emph{Computer Vision - {ECCV} 2022 - 17th European Conference,
  Tel Aviv, Israel, October 23-27, 2022, Proceedings, Part {XXXIX}}, pages
  321--337. Springer, 2022.

\bibitem[Fellbaum(1998)]{Fellbaum1998Wordnet}
Christiane Fellbaum.
\newblock \emph{WordNet: An Electronic Lexical Database}.
\newblock Bradford Books, 1998.

\bibitem[Gan et~al.(2021{\natexlab{a}})Gan, Schwartz, Alter, Mrowca, Schrimpf,
  Traer, Freitas, Kubilius, Bhandwaldar, Haber, Sano, Kim, Wang, Lingelbach,
  Curtis, Feigelis, Bear, Gutfreund, Cox, Torralba, DiCarlo, Tenenbaum,
  McDermott, and Yamins]{Gan2021ThreeDWorld}
Chuang Gan, Jeremy Schwartz, Seth Alter, Damian Mrowca, Martin Schrimpf, James
  Traer, Julian~De Freitas, Jonas Kubilius, Abhishek Bhandwaldar, Nick Haber,
  Megumi Sano, Kuno Kim, Elias Wang, Michael Lingelbach, Aidan Curtis, Kevin~T.
  Feigelis, Daniel Bear, Dan Gutfreund, David~D. Cox, Antonio Torralba,
  James~J. DiCarlo, Josh Tenenbaum, Josh~H. McDermott, and Dan Yamins.
\newblock {ThreeDWorld: {A} Platform for Interactive Multi-Modal Physical
  Simulation}.
\newblock In \emph{Proceedings of the Neural Information Processing Systems
  Track on Datasets and Benchmarks 1, NeurIPS Datasets and Benchmarks 2021,
  December 2021, virtual}, 2021{\natexlab{a}}.

\bibitem[Gan et~al.(2021{\natexlab{b}})Gan, Zhou, Schwartz, Alter, Bhandwaldar,
  Gutfreund, Yamins, DiCarlo, McDermott, Torralba, and
  Tenenbaum]{Gan2021TransportChallenge}
Chuang Gan, Siyuan Zhou, Jeremy Schwartz, Seth Alter, Abhishek Bhandwaldar, Dan
  Gutfreund, Daniel L.~K. Yamins, James~J. DiCarlo, Josh~H. McDermott, Antonio
  Torralba, and Joshua~B. Tenenbaum.
\newblock {The ThreeDWorld Transport Challenge: {A} Visually Guided
  Task-and-Motion Planning Benchmark for Physically Realistic Embodied {AI}}.
\newblock \emph{CoRR}, abs/2103.14025, 2021{\natexlab{b}}.

\bibitem[Gao et~al.(2022)Gao, Gao, Gong, Lin, Thattai, and
  Sukhatme]{Gao2022DialFRED}
Xiaofeng Gao, Qiaozi Gao, Ran Gong, Kaixiang Lin, Govind Thattai, and Gaurav~S.
  Sukhatme.
\newblock {DialFRED: Dialogue-Enabled Agents for Embodied Instruction
  Following}.
\newblock \emph{{IEEE} Robotics Autom. Lett.}, 7\penalty0 (4):\penalty0
  10049--10056, 2022.

\bibitem[Gervet et~al.(2023)Gervet, Chintala, Batra, Malik, and
  Chaplot]{gervet2023navigating}
Theophile Gervet, Soumith Chintala, Dhruv Batra, Jitendra Malik, and
  Devendra~Singh Chaplot.
\newblock Navigating to objects in the real world.
\newblock \emph{Science Robotics}, 2023.

\bibitem[Guo et~al.(2019)Guo, Azar, Piot, Pires, and Munos]{guo2018neural}
Zhaohan~Daniel Guo, Mohammad~Gheshlaghi Azar, Bilal Piot, Bernardo~A Pires, and
  R{\'e}mi Munos.
\newblock Neural predictive belief representations.
\newblock \emph{ICLR}, 2019.

\bibitem[Guo et~al.(2020)Guo, Pires, Piot, Grill, Altch{\'e}, Munos, and
  Azar]{guo2020bootstrap}
Zhaohan~Daniel Guo, Bernardo~Avila Pires, Bilal Piot, Jean-Bastien Grill,
  Florent Altch{\'e}, R{\'e}mi Munos, and Mohammad~Gheshlaghi Azar.
\newblock Bootstrap latent-predictive representations for multitask
  reinforcement learning.
\newblock In \emph{International Conference on Machine Learning}, 2020.

\bibitem[He et~al.(2016)He, Zhang, Ren, and Sun]{He2016ResNet}
Kaiming He, Xiangyu Zhang, Shaoqing Ren, and Jian Sun.
\newblock Deep residual learning for image recognition.
\newblock In \emph{2016 {IEEE} Conference on Computer Vision and Pattern
  Recognition, {CVPR} 2016, Las Vegas, NV, USA, June 27-30, 2016}, pages
  770--778. {IEEE} Computer Society, 2016.

\bibitem[Ho et~al.(2021)Ho, Rao, Xu, Jang, Khansari, and Bai]{Ho2021RetinaGan}
Daniel Ho, Kanishka Rao, Zhuo Xu, Eric Jang, Mohi Khansari, and Yunfei Bai.
\newblock {RetinaGAN: An Object-aware Approach to Sim-to-Real Transfer}.
\newblock In \emph{{IEEE} International Conference on Robotics and Automation,
  {ICRA} 2021, Xi'an, China, May 30 - June 5, 2021}, pages 10920--10926.
  {IEEE}, 2021.

\bibitem[Huang et~al.(2022{\natexlab{a}})Huang, Abbeel, Pathak, and
  Mordatch]{Huang2022LanguageMA}
Wenlong Huang, P. Abbeel, Deepak Pathak, and Igor Mordatch.
\newblock Language models as zero-shot planners: Extracting actionable
  knowledge for embodied agents.
\newblock \emph{ICML}, 2022{\natexlab{a}}.

\bibitem[Huang et~al.(2022{\natexlab{b}})Huang, Xia, Xiao, Chan, Liang,
  Florence, Zeng, Tompson, Mordatch, Chebotar, et~al.]{huang2022inner}
Wenlong Huang, Fei Xia, Ted Xiao, Harris Chan, Jacky Liang, Pete Florence, Andy
  Zeng, Jonathan Tompson, Igor Mordatch, Yevgen Chebotar, et~al.
\newblock Inner monologue: Embodied reasoning through planning with language
  models.
\newblock \emph{arXiv preprint arXiv:2207.05608}, 2022{\natexlab{b}}.

\bibitem[Ichter et~al.(2022)Ichter, Brohan, Chebotar, Finn, Hausman, Herzog,
  Ho, Ibarz, Irpan, Jang, Julian, Kalashnikov, Levine, Lu, Parada, Rao,
  Sermanet, Toshev, Vanhoucke, Xia, Xiao, Xu, Yan, Brown, Ahn, Cortes, Sievers,
  Tan, Xu, Reyes, Rettinghouse, Quiambao, Pastor, Luu, Lee, Kuang, Jesmonth,
  Joshi, Jeffrey, Ruano, Hsu, Gopalakrishnan, David, Zeng, and
  Fu]{Ichter2022SayCan}
Brian Ichter, Anthony Brohan, Yevgen Chebotar, Chelsea Finn, Karol Hausman,
  Alexander Herzog, Daniel Ho, Julian Ibarz, Alex Irpan, Eric Jang, Ryan
  Julian, Dmitry Kalashnikov, Sergey Levine, Yao Lu, Carolina Parada, Kanishka
  Rao, Pierre Sermanet, Alexander Toshev, Vincent Vanhoucke, Fei Xia, Ted Xiao,
  Peng Xu, Mengyuan Yan, Noah Brown, Michael Ahn, Omar Cortes, Nicolas Sievers,
  Clayton Tan, Sichun Xu, Diego Reyes, Jarek Rettinghouse, Jornell Quiambao,
  Peter Pastor, Linda Luu, Kuang{-}Huei Lee, Yuheng Kuang, Sally Jesmonth,
  Nikhil~J. Joshi, Kyle Jeffrey, Rosario~Jauregui Ruano, Jasmine Hsu, Keerthana
  Gopalakrishnan, Byron David, Andy Zeng, and Chuyuan~Kelly Fu.
\newblock Do as {I} can, not as {I} say: Grounding language in robotic
  affordances.
\newblock In \emph{Conference on Robot Learning, CoRL 2022, 14-18 December
  2022, Auckland, New Zealand}, pages 287--318. {PMLR}, 2022.

\bibitem[Janner et~al.(2021)Janner, Li, and Levine]{Janner2021TrajTransf}
Michael Janner, Qiyang Li, and Sergey Levine.
\newblock {Offline Reinforcement Learning as One Big Sequence Modeling
  Problem}.
\newblock In \emph{Advances in Neural Information Processing Systems 34: Annual
  Conference on Neural Information Processing Systems 2021, NeurIPS 2021,
  December 6-14, 2021, virtual}, pages 1273--1286, 2021.

\bibitem[Kebria et~al.(2020)Kebria, Khosravi, Salaken, and
  Nahavandi]{Kebria2020DeepIL}
Parham~Mohsenzadeh Kebria, Abbas Khosravi, Syed~Moshfeq Salaken, and Saeid
  Nahavandi.
\newblock Deep imitation learning for autonomous vehicles based on
  convolutional neural networks.
\newblock \emph{IEEE/CAA Journal of Automatica Sinica}, 7:\penalty0 82--95,
  2020.

\bibitem[Kemp et~al.(2022)Kemp, Edsinger, Clever, and
  Matulevich]{Kemp2022StretchRobot}
Charles~C. Kemp, Aaron Edsinger, Henry~M. Clever, and Blaine Matulevich.
\newblock {The Design of Stretch: {A} Compact, Lightweight Mobile Manipulator
  for Indoor Human Environments}.
\newblock In \emph{2022 International Conference on Robotics and Automation,
  {ICRA} 2022, Philadelphia, PA, USA, May 23-27, 2022}, pages 3150--3157.
  {IEEE}, 2022.

\bibitem[Khandelwal et~al.(2022)Khandelwal, Weihs, Mottaghi, and
  Kembhavi]{Khandelwal2022EmbCLIP}
Apoorv Khandelwal, Luca Weihs, Roozbeh Mottaghi, and Aniruddha Kembhavi.
\newblock Simple but effective: {CLIP} embeddings for embodied {AI}.
\newblock In \emph{{IEEE/CVF} Conference on Computer Vision and Pattern
  Recognition, {CVPR} 2022, New Orleans, LA, USA, June 18-24, 2022}, pages
  14809--14818. {IEEE}, 2022.

\bibitem[Kolve et~al.(2017)Kolve, Mottaghi, Han, VanderBilt, Weihs, Herrasti,
  Deitke, Ehsani, Gordon, Zhu, Kembhavi, Gupta, and
  Farhadi]{Kolve2017AI2THORAI}
Eric Kolve, Roozbeh Mottaghi, Winson Han, Eli VanderBilt, Luca Weihs, Alvaro
  Herrasti, Matt Deitke, Kiana Ehsani, Daniel Gordon, Yuke Zhu, Aniruddha
  Kembhavi, Abhinav~Kumar Gupta, and Ali Farhadi.
\newblock {AI2-THOR: An Interactive 3D Environment for Visual AI}.
\newblock \emph{ArXiv}, abs/1712.05474, 2017.

\bibitem[Ku et~al.(2020)Ku, Anderson, Patel, Ie, and Baldridge]{Ku2020RxR}
Alexander Ku, Peter Anderson, Roma Patel, Eugene Ie, and Jason Baldridge.
\newblock Room-across-room: Multilingual vision-and-language navigation with
  dense spatiotemporal grounding.
\newblock In \emph{Proceedings of the 2020 Conference on Empirical Methods in
  Natural Language Processing, {EMNLP} 2020, Online, November 16-20, 2020},
  pages 4392--4412. Association for Computational Linguistics, 2020.

\bibitem[Li et~al.(2021)Li, Xia, Mart{\'{\i}}n{-}Mart{\'{\i}}n, Lingelbach,
  Srivastava, Shen, Vainio, Gokmen, Dharan, Jain, Kurenkov, Liu, Gweon, Wu,
  Fei{-}Fei, and Savarese]{Li2021iGibson2}
Chengshu Li, Fei Xia, Roberto Mart{\'{\i}}n{-}Mart{\'{\i}}n, Michael
  Lingelbach, Sanjana Srivastava, Bokui Shen, Kent~Elliott Vainio, Cem Gokmen,
  Gokul Dharan, Tanish Jain, Andrey Kurenkov, C.~Karen Liu, Hyowon Gweon,
  Jiajun Wu, Li Fei{-}Fei, and Silvio Savarese.
\newblock igibson 2.0: Object-centric simulation for robot learning of everyday
  household tasks.
\newblock In \emph{Conference on Robot Learning, 8-11 November 2021, London,
  {UK}}, pages 455--465. {PMLR}, 2021.

\bibitem[Li et~al.(2022)Li, Zhang, Wong, Gokmen, Srivastava,
  Mart{\'{\i}}n{-}Mart{\'{\i}}n, Wang, Levine, Lingelbach, Sun, Anvari, Hwang,
  Sharma, Aydin, Bansal, Hunter, Kim, Lou, Matthews, Villa{-}Renteria, Tang,
  Tang, Xia, Savarese, Gweon, Liu, Wu, and Fei{-}Fei]{Li2022Behavior}
Chengshu Li, Ruohan Zhang, Josiah Wong, Cem Gokmen, Sanjana Srivastava, Roberto
  Mart{\'{\i}}n{-}Mart{\'{\i}}n, Chen Wang, Gabrael Levine, Michael Lingelbach,
  Jiankai Sun, Mona Anvari, Minjune Hwang, Manasi Sharma, Arman Aydin, Dhruva
  Bansal, Samuel Hunter, Kyu{-}Young Kim, Alan Lou, Caleb~R. Matthews, Ivan
  Villa{-}Renteria, Jerry~Huayang Tang, Claire Tang, Fei Xia, Silvio Savarese,
  Hyowon Gweon, Karen Liu, Jiajun Wu, and Li Fei{-}Fei.
\newblock {{BEHAVIOR-1K:} {A} Benchmark for Embodied {AI} with 1, 000 Everyday
  Activities and Realistic Simulation}.
\newblock In \emph{Conference on Robot Learning, CoRL 2022, 14-18 December
  2022, Auckland, New Zealand}, pages 80--93. {PMLR}, 2022.

\bibitem[Li et~al.(2018)Li, M{\"u}ller, Casser, Smith, Michels, and
  Ghanem]{Li2018OILOI}
G. Li, Matthias M{\"u}ller, Vincent Casser, Neil~G. Smith, Dominik~Ludewig
  Michels, and Bernard Ghanem.
\newblock Oil: Observational imitation learning.
\newblock \emph{Robotics: Science and Systems XV}, 2018.

\bibitem[Liu et~al.(2023)Liu, Jiang, Zhang, Liu, Zhang, Biswas, and
  Stone]{liu2023llm+}
Bo Liu, Yuqian Jiang, Xiaohan Zhang, Qiang Liu, Shiqi Zhang, Joydeep Biswas,
  and Peter Stone.
\newblock Llm+ p: Empowering large language models with optimal planning
  proficiency.
\newblock \emph{arXiv preprint arXiv:2304.11477}, 2023.

\bibitem[Majumdar et~al.(2023{\natexlab{a}})Majumdar, Yadav, Arnaud, Ma, Chen,
  Silwal, Jain, Berges, Abbeel, Malik, Batra, Lin, Maksymets, Rajeswaran, and
  Meier]{Majumdar2023VisualCortex}
Arjun Majumdar, Karmesh Yadav, Sergio Arnaud, Yecheng~Jason Ma, Claire Chen,
  Sneha Silwal, Aryan Jain, Vincent{-}Pierre Berges, Pieter Abbeel, Jitendra
  Malik, Dhruv Batra, Yixin Lin, Oleksandr Maksymets, Aravind Rajeswaran, and
  Franziska Meier.
\newblock Where are we in the search for an artificial visual cortex for
  embodied intelligence?
\newblock \emph{CoRR}, abs/2303.18240, 2023{\natexlab{a}}.

\bibitem[Majumdar et~al.(2023{\natexlab{b}})Majumdar, Yadav, Arnaud, Ma, Chen,
  Silwal, Jain, Berges, Abbeel, Malik, et~al.]{majumdar2023we}
Arjun Majumdar, Karmesh Yadav, Sergio Arnaud, Yecheng~Jason Ma, Claire Chen,
  Sneha Silwal, Aryan Jain, Vincent-Pierre Berges, Pieter Abbeel, Jitendra
  Malik, et~al.
\newblock Where are we in the search for an artificial visual cortex for
  embodied intelligence?
\newblock \emph{ICLR}, 2023{\natexlab{b}}.

\bibitem[McCrae et~al.(2019)McCrae, Rademaker, Bond, Rudnicka, and
  Fellbaum]{McCrae2019WordnetOpenSource}
John~P. McCrae, Alexandre Rademaker, Francis Bond, Ewa Rudnicka, and Christiane
  Fellbaum.
\newblock {English WordNet 2019 - An Open-Source WordNet for English}.
\newblock In \emph{Proceedings of the 10th Global Wordnet Conference, {GWC}
  2019, Wroclaw, Poland, July 23-27, 2019}, pages 245--252. Global Wordnet
  Association, 2019.

\bibitem[Mero et~al.(2022)Mero, Yi, Dianati, and Mouzakitis]{LeMero2022ASO}
Luc~Le Mero, Dewei Yi, Mehrdad Dianati, and Alexandros Mouzakitis.
\newblock A survey on imitation learning techniques for end-to-end autonomous
  vehicles.
\newblock \emph{IEEE Transactions on Intelligent Transportation Systems},
  23:\penalty0 14128--14147, 2022.

\bibitem[Mnih et~al.(2016)Mnih, Badia, Mirza, Graves, Lillicrap, Harley,
  Silver, and Kavukcuoglu]{Mnih2016A3C}
Volodymyr Mnih, Adri{\`{a}}~Puigdom{\`{e}}nech Badia, Mehdi Mirza, Alex Graves,
  Timothy~P. Lillicrap, Tim Harley, David Silver, and Koray Kavukcuoglu.
\newblock {Asynchronous Methods for Deep Reinforcement Learning}.
\newblock In \emph{Proceedings of the 33nd International Conference on Machine
  Learning, {ICML} 2016, New York City, NY, USA, June 19-24, 2016}, pages
  1928--1937. JMLR.org, 2016.

\bibitem[Oquab et~al.(2023)Oquab, Darcet, Moutakanni, Vo, Szafraniec, Khalidov,
  Fernandez, Haziza, Massa, El-Nouby, Howes, Huang, Xu, Sharma, Li, Galuba,
  Rabbat, Assran, Ballas, Synnaeve, Misra, Jegou, Mairal, Labatut, Joulin, and
  Bojanowski]{oquab2023dinov2}
Maxime Oquab, Timothée Darcet, Theo Moutakanni, Huy~V. Vo, Marc Szafraniec,
  Vasil Khalidov, Pierre Fernandez, Daniel Haziza, Francisco Massa, Alaaeldin
  El-Nouby, Russell Howes, Po-Yao Huang, Hu Xu, Vasu Sharma, Shang-Wen Li,
  Wojciech Galuba, Mike Rabbat, Mido Assran, Nicolas Ballas, Gabriel Synnaeve,
  Ishan Misra, Herve Jegou, Julien Mairal, Patrick Labatut, Armand Joulin, and
  Piotr Bojanowski.
\newblock Dinov2: Learning robust visual features without supervision, 2023.

\bibitem[Ouyang et~al.(2022)Ouyang, Wu, Jiang, Almeida, Wainwright, Mishkin,
  Zhang, Agarwal, Slama, Ray, Schulman, Hilton, Kelton, Miller, Simens, Askell,
  Welinder, Christiano, Leike, and Lowe]{Ouyang2022InstructGPT}
Long Ouyang, Jeffrey Wu, Xu Jiang, Diogo Almeida, Carroll~L. Wainwright, Pamela
  Mishkin, Chong Zhang, Sandhini Agarwal, Katarina Slama, Alex Ray, John
  Schulman, Jacob Hilton, Fraser Kelton, Luke Miller, Maddie Simens, Amanda
  Askell, Peter Welinder, Paul~F. Christiano, Jan Leike, and Ryan Lowe.
\newblock Training language models to follow instructions with human feedback.
\newblock In \emph{NeurIPS}, 2022.

\bibitem[Pan et~al.(2019)Pan, Cheng, Saigol, Lee, Yan, Theodorou, and
  Boots]{Pan2019ImitationLF}
Yunpeng Pan, Ching-An Cheng, Kamil Saigol, Keuntaek Lee, Xinyan Yan,
  Evangelos~A. Theodorou, and Byron Boots.
\newblock Imitation learning for agile autonomous driving.
\newblock \emph{The International Journal of Robotics Research}, 39:\penalty0
  286 -- 302, 2019.

\bibitem[Puig et~al.(2018)Puig, Ra, Boben, Li, Wang, Fidler, and
  Torralba]{Puig2018VirtualHome}
Xavier Puig, Kevin Ra, Marko Boben, Jiaman Li, Tingwu Wang, Sanja Fidler, and
  Antonio Torralba.
\newblock {VirtualHome: Simulating Household Activities via Programs}.
\newblock In \emph{2018 {IEEE} Conference on Computer Vision and Pattern
  Recognition, {CVPR} 2018, Salt Lake City, UT, USA, June 18-22, 2018}, pages
  8494--8502. Computer Vision Foundation / {IEEE} Computer Society, 2018.

\bibitem[Puig et~al.(2023)Puig, Undersander, Szot, Cote, Yang, Partsey, Desai,
  Clegg, Hlavac, Min, Vondrus, Gervet, Berges, Turner, Maksymets, Kira,
  Kalakrishnan, Malik, Chaplot, Jain, Batra, Rai, and
  Mottaghi]{Puig2023Habitat3}
Xavier Puig, Eric Undersander, Andrew Szot, Mikael~Dallaire Cote, Tsung{-}Yen
  Yang, Ruslan Partsey, Ruta Desai, Alexander~William Clegg, Michal Hlavac,
  So~Yeon Min, Vladimir Vondrus, Th{\'{e}}ophile Gervet, Vincent{-}Pierre
  Berges, John~M. Turner, Oleksandr Maksymets, Zsolt Kira, Mrinal Kalakrishnan,
  Jitendra Malik, Devendra~Singh Chaplot, Unnat Jain, Dhruv Batra, Akshara Rai,
  and Roozbeh Mottaghi.
\newblock {Habitat 3.0: {A} Co-Habitat for Humans, Avatars and Robots}.
\newblock \emph{CoRR}, abs/2310.13724, 2023.

\bibitem[Pulver et~al.(2021)Pulver, Eiras, Carozza, Hawasly, Albrecht, and
  Ramamoorthy]{pulver2021pilot}
Henry Pulver, Francisco Eiras, Ludovico Carozza, Majd Hawasly, Stefano~V
  Albrecht, and Subramanian Ramamoorthy.
\newblock Pilot: Efficient planning by imitation learning and optimisation for
  safe autonomous driving.
\newblock In \emph{2021 IEEE/RSJ International Conference on Intelligent Robots
  and Systems (IROS)}, pages 1442--1449. IEEE, 2021.

\bibitem[Qi et~al.(2020)Qi, Wu, Anderson, Wang, Wang, Shen, and van~den
  Hengel]{Qi2020Reverie}
Yuankai Qi, Qi Wu, Peter Anderson, Xin Wang, William~Yang Wang, Chunhua Shen,
  and Anton van~den Hengel.
\newblock {{REVERIE:} Remote Embodied Visual Referring Expression in Real
  Indoor Environments}.
\newblock In \emph{2020 {IEEE/CVF} Conference on Computer Vision and Pattern
  Recognition, {CVPR} 2020, Seattle, WA, USA, June 13-19, 2020}, pages
  9979--9988. Computer Vision Foundation / {IEEE}, 2020.

\bibitem[Radford et~al.(2021)Radford, Kim, Hallacy, Ramesh, Goh, Agarwal,
  Sastry, Askell, Mishkin, Clark, Krueger, and Sutskever]{Radford2021CLIP}
Alec Radford, Jong~Wook Kim, Chris Hallacy, Aditya Ramesh, Gabriel Goh,
  Sandhini Agarwal, Girish Sastry, Amanda Askell, Pamela Mishkin, Jack Clark,
  Gretchen Krueger, and Ilya Sutskever.
\newblock {Learning Transferable Visual Models From Natural Language
  Supervision}.
\newblock In \emph{Proceedings of the 38th International Conference on Machine
  Learning, {ICML} 2021, 18-24 July 2021, Virtual Event}, pages 8748--8763.
  {PMLR}, 2021.

\bibitem[Ramrakhya et~al.(2022)Ramrakhya, Undersander, Batra, and
  Das]{Ramrakhya2022HabitatWeb}
Ram Ramrakhya, Eric Undersander, Dhruv Batra, and Abhishek Das.
\newblock Habitat-web: Learning embodied object-search strategies from human
  demonstrations at scale.
\newblock In \emph{{IEEE/CVF} Conference on Computer Vision and Pattern
  Recognition, {CVPR} 2022, New Orleans, LA, USA, June 18-24, 2022}, pages
  5163--5173. {IEEE}, 2022.

\bibitem[Ramrakhya et~al.(2023)Ramrakhya, Batra, Wijmans, and
  Das]{ramrakhya2023pirlnav}
Ram Ramrakhya, Dhruv Batra, Erik Wijmans, and Abhishek Das.
\newblock Pirlnav: Pretraining with imitation and rl finetuning for objectnav.
\newblock In \emph{Workshop on IEEE/CVF Conference on Computer Vision and
  Pattern Recognition}, 2023.

\bibitem[Reed et~al.(2022)Reed, Zolna, Parisotto, Colmenarejo, Novikov,
  Barth{-}Maron, Gimenez, Sulsky, Kay, Springenberg, Eccles, Bruce, Razavi,
  Edwards, Heess, Chen, Hadsell, Vinyals, Bordbar, and
  de~Freitas]{Reed2022GATO}
Scott~E. Reed, Konrad Zolna, Emilio Parisotto, Sergio~G{\'{o}}mez Colmenarejo,
  Alexander Novikov, Gabriel Barth{-}Maron, Mai Gimenez, Yury Sulsky, Jackie
  Kay, Jost~Tobias Springenberg, Tom Eccles, Jake Bruce, Ali Razavi, Ashley
  Edwards, Nicolas Heess, Yutian Chen, Raia Hadsell, Oriol Vinyals, Mahyar
  Bordbar, and Nando de Freitas.
\newblock {A Generalist Agent}.
\newblock \emph{Trans. Mach. Learn. Res.}, 2022, 2022.

\bibitem[Savva et~al.(2019)Savva, Malik, Parikh, Batra, Kadian, Maksymets,
  Zhao, Wijmans, Jain, Straub, Liu, and Koltun]{Savva2019Habitat}
Manolis Savva, Jitendra Malik, Devi Parikh, Dhruv Batra, Abhishek Kadian,
  Oleksandr Maksymets, Yili Zhao, Erik Wijmans, Bhavana Jain, Julian Straub,
  Jia Liu, and Vladlen Koltun.
\newblock {Habitat: {A} Platform for Embodied {AI} Research}.
\newblock In \emph{2019 {IEEE/CVF} International Conference on Computer Vision,
  {ICCV} 2019, Seoul, Korea (South), October 27 - November 2, 2019}, pages
  9338--9346. {IEEE}, 2019.

\bibitem[Schulman et~al.(2017)Schulman, Wolski, Dhariwal, Radford, and
  Klimov]{Schulman2017A2CPPO}
John Schulman, Filip Wolski, Prafulla Dhariwal, Alec Radford, and Oleg Klimov.
\newblock {Proximal Policy Optimization Algorithms}.
\newblock \emph{CoRR}, abs/1707.06347, 2017.

\bibitem[Shen et~al.(2021)Shen, Xia, Li, Mart{\'{\i}}n{-}Mart{\'{\i}}n, Fan,
  Wang, P{\'{e}}rez{-}D'Arpino, Buch, Srivastava, Tchapmi, Tchapmi, Vainio,
  Wong, Fei{-}Fei, and Savarese]{Shen2021iGibson1}
Bokui Shen, Fei Xia, Chengshu Li, Roberto Mart{\'{\i}}n{-}Mart{\'{\i}}n, Linxi
  Fan, Guanzhi Wang, Claudia P{\'{e}}rez{-}D'Arpino, Shyamal Buch, Sanjana
  Srivastava, Lyne Tchapmi, Micael Tchapmi, Kent Vainio, Josiah Wong, Li
  Fei{-}Fei, and Silvio Savarese.
\newblock {iGibson 1.0: {A} Simulation Environment for Interactive Tasks in
  Large Realistic Scenes}.
\newblock In \emph{{IEEE/RSJ} International Conference on Intelligent Robots
  and Systems, {IROS} 2021, Prague, Czech Republic, September 27 - Oct. 1,
  2021}, pages 7520--7527. {IEEE}, 2021.

\bibitem[Shridhar et~al.(2020)Shridhar, Thomason, Gordon, Bisk, Han, Mottaghi,
  Zettlemoyer, and Fox]{Shridhar2020ALFRED}
Mohit Shridhar, Jesse Thomason, Daniel Gordon, Yonatan Bisk, Winson Han,
  Roozbeh Mottaghi, Luke Zettlemoyer, and Dieter Fox.
\newblock {ALFRED:} {A} benchmark for interpreting grounded instructions for
  everyday tasks.
\newblock In \emph{2020 {IEEE/CVF} Conference on Computer Vision and Pattern
  Recognition, {CVPR} 2020, Seattle, WA, USA, June 13-19, 2020}, pages
  10737--10746. Computer Vision Foundation / {IEEE}, 2020.

\bibitem[Singh et~al.(2023)Singh, Salvador, Weihs, and Kembhavi]{Singh2023SGC}
Kunal~Pratap Singh, Jordi Salvador, Luca Weihs, and Aniruddha Kembhavi.
\newblock {Scene Graph Contrastive Learning for Embodied Navigation}.
\newblock In \emph{Proceedings of the IEEE/CVF International Conference on
  Computer Vision (ICCV)}, pages 10884--10894, 2023.

\bibitem[Song et~al.(2023)Song, Wu, Washington, Sadler, Chao, and
  Su]{song2023llm}
Chan~Hee Song, Jiaman Wu, Clayton Washington, Brian~M Sadler, Wei-Lun Chao, and
  Yu Su.
\newblock Llm-planner: Few-shot grounded planning for embodied agents with
  large language models.
\newblock \emph{Proceedings of the IEEE/CVF International Conference on
  Computer Vision}, 2023.

\bibitem[Srivastava et~al.(2021)Srivastava, Li, Lingelbach, Mart'in-Mart'in,
  Xia, Vainio, Lian, Gokmen, Buch, Liu, Savarese, Gweon, Wu, and
  Fei-Fei]{Srivastava2021Behavior100}
Sanjana Srivastava, Chengshu Li, Michael Lingelbach, Roberto Mart'in-Mart'in,
  Fei Xia, Kent Vainio, Zheng Lian, Cem Gokmen, S. Buch, C.~Karen Liu, Silvio
  Savarese, Hyowon Gweon, Jiajun Wu, and Li Fei-Fei.
\newblock {BEHAVIOR: Benchmark for Everyday Household Activities in Virtual,
  Interactive, and Ecological Environments}.
\newblock In \emph{Conference on Robot Learning}, 2021.

\bibitem[Szot et~al.(2021)Szot, Clegg, Undersander, Wijmans, Zhao, Turner,
  Maestre, Mukadam, Chaplot, Maksymets, Gokaslan, Vondrus, Dharur, Meier,
  Galuba, Chang, Kira, Koltun, Malik, Savva, and Batra]{Szot2021Habitat2}
Andrew Szot, Alexander Clegg, Eric Undersander, Erik Wijmans, Yili Zhao,
  John~M. Turner, Noah Maestre, Mustafa Mukadam, Devendra~Singh Chaplot,
  Oleksandr Maksymets, Aaron Gokaslan, Vladimir Vondrus, Sameer Dharur,
  Franziska Meier, Wojciech Galuba, Angel~X. Chang, Zsolt Kira, Vladlen Koltun,
  Jitendra Malik, Manolis Savva, and Dhruv Batra.
\newblock {Habitat 2.0: Training Home Assistants to Rearrange their Habitat}.
\newblock In \emph{Advances in Neural Information Processing Systems 34: Annual
  Conference on Neural Information Processing Systems 2021, NeurIPS 2021,
  December 6-14, 2021, virtual}, pages 251--266, 2021.

\bibitem[Tobin et~al.(2017)Tobin, Fong, Ray, Schneider, Zaremba, and
  Abbeel]{Tobin2017DomainRand}
Josh Tobin, Rachel Fong, Alex Ray, Jonas Schneider, Wojciech Zaremba, and
  Pieter Abbeel.
\newblock {Domain randomization for transferring deep neural networks from
  simulation to the real world}.
\newblock In \emph{2017 {IEEE/RSJ} International Conference on Intelligent
  Robots and Systems, {IROS} 2017, Vancouver, BC, Canada, September 24-28,
  2017}, pages 23--30. {IEEE}, 2017.

\bibitem[Vemprala et~al.(2023)Vemprala, Bonatti, Bucker, and
  Kapoor]{vemprala2023chatgpt}
Sai Vemprala, Rogerio Bonatti, Arthur Bucker, and Ashish Kapoor.
\newblock Chatgpt for robotics: Design principles and model abilities.
\newblock Technical Report MSR-TR-2023-8, Microsoft, 2023.

\bibitem[Wang et~al.(2023)Wang, Cai, Liu, Ma, and Liang]{wang2023describe}
Zihao Wang, Shaofei Cai, Anji Liu, Xiaojian Ma, and Yitao Liang.
\newblock Describe, explain, plan and select: Interactive planning with large
  language models enables open-world multi-task agents.
\newblock \emph{arXiv preprint arXiv:2302.01560}, 2023.

\bibitem[Wei et~al.(2023)Wei, Sun, Zheng, Vemprala, Bonatti, Chen, Madaan, Ba,
  Kapoor, and Ma]{Wei2023imitation}
Yao Wei, Yanchao Sun, Ruijie Zheng, Sai Vemprala, Rogerio Bonatti, Shuhang
  Chen, Ratnesh Madaan, Zhongjie Ba, Ashish Kapoor, and Shuang Ma.
\newblock {Is Imitation All You Need? Generalized Decision-Making with
  Dual-Phase Training}.
\newblock In \emph{Proceedings of the IEEE/CVF International Conference on
  Computer Vision}, pages 16221--16231, 2023.

\bibitem[Weihs et~al.(2020)Weihs, Salvador, Kotar, Jain, Zeng, Mottaghi, and
  Kembhavi]{AllenAct}
Luca Weihs, Jordi Salvador, Klemen Kotar, Unnat Jain, Kuo-Hao Zeng, Roozbeh
  Mottaghi, and Aniruddha Kembhavi.
\newblock Allenact: A framework for embodied ai research.
\newblock \emph{arXiv preprint arXiv:2008.12760}, 2020.

\bibitem[Weihs et~al.(2021{\natexlab{a}})Weihs, Deitke, Kembhavi, and
  Mottaghi]{Weihs2021Rearrangement}
Luca Weihs, Matt Deitke, Aniruddha Kembhavi, and Roozbeh Mottaghi.
\newblock Visual room rearrangement.
\newblock In \emph{IEEE/CVF Conference on Computer Vision and Pattern
  Recognition (CVPR)}, 2021{\natexlab{a}}.

\bibitem[Weihs et~al.(2021{\natexlab{b}})Weihs, Jain, Liu, Salvador, Lazebnik,
  Kembhavi, and Schwing]{Weihs2021Advisor}
Luca Weihs, Unnat Jain, Iou{-}Jen Liu, Jordi Salvador, Svetlana Lazebnik,
  Aniruddha Kembhavi, and Alexander~G. Schwing.
\newblock Bridging the imitation gap by adaptive insubordination.
\newblock In \emph{Advances in Neural Information Processing Systems 34: Annual
  Conference on Neural Information Processing Systems 2021, NeurIPS 2021,
  December 6-14, 2021, virtual}, pages 19134--19146, 2021{\natexlab{b}}.

\bibitem[Wijmans et~al.(2020)Wijmans, Kadian, Morcos, Lee, Essa, Parikh, Savva,
  and Batra]{Wijmans2020DDPPO}
Erik Wijmans, Abhishek Kadian, Ari Morcos, Stefan Lee, Irfan Essa, Devi Parikh,
  Manolis Savva, and Dhruv Batra.
\newblock {DD-PPO:} learning near-perfect pointgoal navigators from 2.5 billion
  frames.
\newblock In \emph{8th International Conference on Learning Representations,
  {ICLR} 2020, Addis Ababa, Ethiopia, April 26-30, 2020}. OpenReview.net, 2020.

\bibitem[Xia et~al.(2018)Xia, Zamir, He, Sax, Malik, and
  Savarese]{Xia2018Gibson}
Fei Xia, Amir~R. Zamir, Zhi{-}Yang He, Alexander Sax, Jitendra Malik, and
  Silvio Savarese.
\newblock {Gibson Env: Real-World Perception for Embodied Agents}.
\newblock In \emph{2018 {IEEE} Conference on Computer Vision and Pattern
  Recognition, {CVPR} 2018, Salt Lake City, UT, USA, June 18-22, 2018}, pages
  9068--9079. Computer Vision Foundation / {IEEE} Computer Society, 2018.

\bibitem[Xiang et~al.(2020)Xiang, Qin, Mo, Xia, Zhu, Liu, Liu, Jiang, Yuan,
  Wang, Yi, Chang, Guibas, and Su]{Xiang2020SAPIEN}
Fanbo Xiang, Yuzhe Qin, Kaichun Mo, Yikuan Xia, Hao Zhu, Fangchen Liu, Minghua
  Liu, Hanxiao Jiang, Yifu Yuan, He Wang, Li Yi, Angel~X. Chang, Leonidas~J.
  Guibas, and Hao Su.
\newblock {{SAPIEN}: A SimulAted Part-based Interactive ENvironment}.
\newblock In \emph{The IEEE Conference on Computer Vision and Pattern
  Recognition (CVPR)}, 2020.

\bibitem[Yadav et~al.(2023)Yadav, Ramrakhya, Majumdar, Berges, Kuhar, Batra,
  Baevski, and Maksymets]{yadav2023offline}
Karmesh Yadav, Ram Ramrakhya, Arjun Majumdar, Vincent-Pierre Berges, Sachit
  Kuhar, Dhruv Batra, Alexei Baevski, and Oleksandr Maksymets.
\newblock Offline visual representation learning for embodied navigation.
\newblock In \emph{Workshop on Reincarnating Reinforcement Learning at ICLR
  2023}, 2023.

\bibitem[Yamauchi(1997)]{yamauchi1997frontier}
Brian Yamauchi.
\newblock A frontier-based approach for autonomous exploration.
\newblock In \emph{Proceedings 1997 IEEE International Symposium on
  Computational Intelligence in Robotics and Automation CIRA'97.'Towards New
  Computational Principles for Robotics and Automation'}, pages 146--151. IEEE,
  1997.

\bibitem[Ye et~al.(2021)Ye, Batra, Das, and Wijmans]{Ye2021AuxTasksObjectNav}
Joel Ye, Dhruv Batra, Abhishek Das, and Erik Wijmans.
\newblock Auxiliary tasks and exploration enable objectnav.
\newblock \emph{CoRR}, abs/2104.04112, 2021.

\bibitem[Yenamandra et~al.(2023)Yenamandra, Ramachandran, Yadav, Wang, Khanna,
  Gervet, Yang, Jain, Clegg, Turner, Kira, Savva, Chang, Chaplot, Batra,
  Mottaghi, Bisk, and Paxton]{Yenamandra2023HomeRobotOVMM}
Sriram Yenamandra, Arun Ramachandran, Karmesh Yadav, Austin Wang, Mukul Khanna,
  Th{\'{e}}ophile Gervet, Tsung{-}Yen Yang, Vidhi Jain, Alexander~William
  Clegg, John~M. Turner, Zsolt Kira, Manolis Savva, Angel~X. Chang,
  Devendra~Singh Chaplot, Dhruv Batra, Roozbeh Mottaghi, Yonatan Bisk, and
  Chris Paxton.
\newblock Homerobot: Open-vocabulary mobile manipulation.
\newblock \emph{CoRR}, abs/2306.11565, 2023.

\bibitem[Yokoyama et~al.(2023)Yokoyama, Clegg, Undersander, Ha, Batra, and
  Rai]{Yokoyama2023AdaptiveSC}
Naoki Yokoyama, Alexander Clegg, Eric Undersander, Sehoon Ha, Dhruv Batra, and
  Akshara Rai.
\newblock Adaptive skill coordination for robotic mobile manipulation.
\newblock \emph{ArXiv}, abs/2304.00410, 2023.

\bibitem[Zeng et~al.(2021)Zeng, Weihs, Farhadi, and Mottaghi]{zeng2021pushing}
Kuo-Hao Zeng, Luca Weihs, Ali Farhadi, and Roozbeh Mottaghi.
\newblock Pushing it out of the way: Interactive visual navigation.
\newblock In \emph{Proceedings of the IEEE/CVF Conference on Computer Vision
  and Pattern Recognition}, 2021.

\bibitem[Zhai et~al.(2023)Zhai, Mustafa, Kolesnikov, and Beyer]{Zhai2023SigLIP}
Xiaohua Zhai, Basil Mustafa, Alexander Kolesnikov, and Lucas Beyer.
\newblock Sigmoid loss for language image pre-training.
\newblock \emph{ICCV}, abs/2303.15343, 2023.

\bibitem[Zhao et~al.(2023)Zhao, Ding, An, Du, Yu, Li, Tang, and
  Wang]{zhao2023fast}
Xu Zhao, Wenchao Ding, Yongqi An, Yinglong Du, Tao Yu, Min Li, Ming Tang, and
  Jinqiao Wang.
\newblock Fast segment anything.
\newblock \emph{arXiv preprint arXiv:2306.12156}, 2023.

\bibitem[Zhou et~al.(2022)Zhou, Girdhar, Joulin, Kr{\"{a}}henb{\"{u}}hl, and
  Misra]{Zhou2022DETIC}
Xingyi Zhou, Rohit Girdhar, Armand Joulin, Philipp Kr{\"{a}}henb{\"{u}}hl, and
  Ishan Misra.
\newblock Detecting twenty-thousand classes using image-level supervision.
\newblock In \emph{Computer Vision - {ECCV} 2022 - 17th European Conference,
  Tel Aviv, Israel, October 23-27, 2022, Proceedings, Part {IX}}, pages
  350--368. Springer, 2022.

\bibitem[Zhu et~al.(2017{\natexlab{a}})Zhu, Park, Isola, and
  Efros]{Zhu2017CycleGAN}
Jun{-}Yan Zhu, Taesung Park, Phillip Isola, and Alexei~A. Efros.
\newblock {Unpaired Image-to-Image Translation Using Cycle-Consistent
  Adversarial Networks}.
\newblock In \emph{{IEEE} International Conference on Computer Vision, {ICCV}
  2017, Venice, Italy, October 22-29, 2017}, pages 2242--2251. {IEEE} Computer
  Society, 2017{\natexlab{a}}.

\bibitem[Zhu et~al.(2017{\natexlab{b}})Zhu, Mottaghi, Kolve, Lim, Gupta,
  Fei{-}Fei, and Farhadi]{Zhu2017VisualNav}
Yuke Zhu, Roozbeh Mottaghi, Eric Kolve, Joseph~J. Lim, Abhinav Gupta, Li
  Fei{-}Fei, and Ali Farhadi.
\newblock {Target-driven visual navigation in indoor scenes using deep
  reinforcement learning}.
\newblock In \emph{2017 {IEEE} International Conference on Robotics and
  Automation, {ICRA} 2017, Singapore, Singapore, May 29 - June 3, 2017}, pages
  3357--3364. {IEEE}, 2017{\natexlab{b}}.

\end{thebibliography}
}

\appendix
\clearpage
\setcounter{page}{1}
\maketitlesupplementary

\section{Training details}
\label{sec:supp_training}
\model uses \siglip image and text encoders that produce $84\times768$ ($n_\text{patch}\times d_\text{image}$) features and $768$ dimension feature per text token. We use $3$-layer transformer encoder and decoder with $512$ dimensional hidden state ($d_\text{visual}$) and $8$ attention heads for the goal-conditioned visual encoder and action decoder, respectively. We use a context window of $100$. All models are trained with a batch size of $224$ on $8\times$A100 GPUs ($80$ GB memory / GPU) with the AdamW optimizer and a learning rate of $0.0002$. Single-task models are trained for 20k iterations, while multi-task models are trained for 50k iterations. Unlike RL which involves simulation in the training loop, IL has the advantage of saving the training episodes to disk offline. Using $16$-bit mixed precision training \model trains at approximately 1.2 hours per 1000 iterations, an FPS of ${\approx}$3500, compared to an FPS of ${\approx}$175 for RL implemented using AllenAct~\cite{AllenAct}. We find that data augmentation both during training and testing is critical for model performance, both in simulation and real, when training on MP4 compressed videos. In particular, we always apply color jitter, gaussian blurring, and random cropping while posterization and sharpness adjustments are applied with specified probabilities to the RGB frames from both cameras. In addition to the RGB frames, we incorporate an \texttt{object-in-hand} sensor to indicate whether the agent is holding an object. For details on how these sensors are impremented in real experiments, see Sections \ref{sec:hardware} and \ref{sec:real_sensors}. In the decoder, we additively combine the embedding of \texttt{object-in-hand} sensor with visual representations, sinusoidal temporal position encodings, and learned embeddings of previous action. This sensor is important for tasks involving manipulation, such as \Fetch and \PickUp, where it enables the agent to learn when to issue a terminate action to end an episode.

For the model with detection (\model w/ GT Det or \model w/ DETIC), we encode the coordinates of the bounding boxes using sinusoidal positional encoding followed by a linear layer with LayerNorm and ReLU. We also add coordinate type embeddings to differentiate the 10 coordinates (5 per camera - $x_1$, $y_1$, $x_2$, $y_2$, and area). The coordinate encodings are then concatenated with the two image features and text features before feeding into Transformer Encoder $\mathcal{E}_\text{visual}$. When no object is detected in the image, we use $1000$ as the dummy coordinate value and set area to zero.

\subsection{RL training details.}
We train our RL baselines using DD-PPO with $64$ samplers on a single machine with $8$ A6000 GPUs, 
closely following the ProcTHOR implementation. During training,
we use the Adam optimizer with a fixed learning rate of \texttt{3e-4}. %
As DD-PPO is an on-policy algorithm, a replay buffer is not utilized during training.
The step count for DD-PPO is set at $128$. The same data augmentations are applied during both training and eval. The max episode length is $600$ steps for all tasks except RoomVisit which has a $1000$ step max.
The RL baselines train for \emph{double the wall-clock time} compared to SPOC, totaling around $40$M environment steps. 

We trained SPOC for two days and RL baselines for four days both using a single machine with 8 GPUs, resulting in 384 GPU hours for SPOC and 768 GPU hours for RL baselines

\section{Data generation}
\label{sec:data_generation}

In this section, we describe our \bench benchmark, a challenging collection of thousands of tasks corresponding to \numvidatasks task types in virtual household scenarios. We first focus on the definition of the procedural houses and then proceed to detail the rationale and details of the included task types.

\subsection{Houses}
\label{sec:obja_houses}

Houses are procedurally generated using ProcTHOR \cite{Deitke2022ProcTHORLE}, a powerful procedural environment generator for the THOR simulator \cite{Kolve2017AI2THORAI}. ProcTHOR can generate complete household environments by sampling floorplans, adding doors and windows, placing large object types and adding decorations and lights. The differences with respect to the original ProcTHOR houses are (1) the use of specific layouts specifications; and (2) the inclusion of a much larger set of 3D objects imported from Objaverse \cite{Deitke2022ObjaverseAU} and complemented with extended annotations for scale, standard pose, and descriptions.

\subsubsection{Layouts}

\begin{figure*}
\centering
\includegraphics[width=0.8\textwidth]{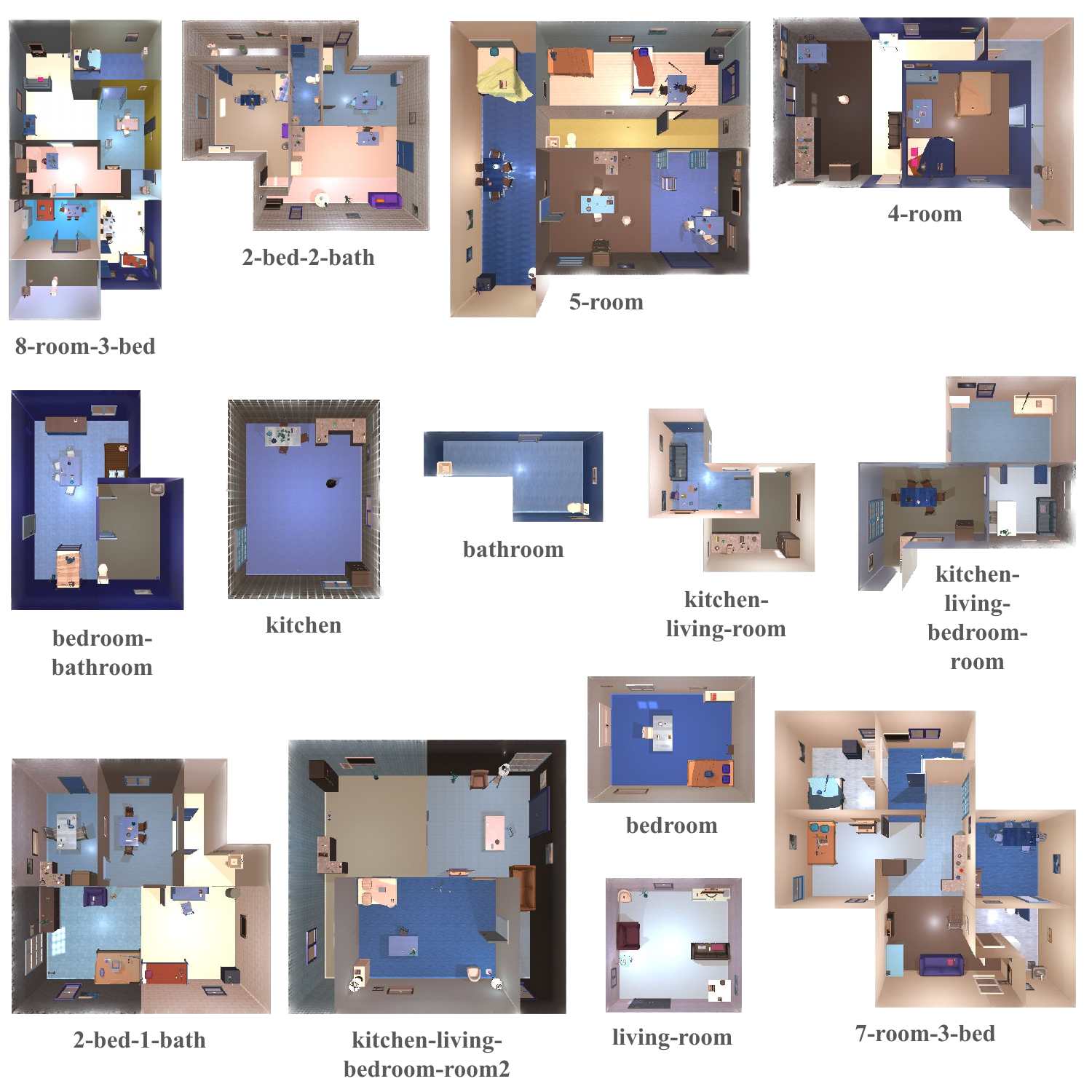}
\caption{\label{fig:layout_distribution}\textbf{Variety of layouts.} Sample houses for each of the 14 layout specifications.}
\end{figure*}

House layouts are sampled from the 14 specifications shown in Table \ref{tab:layouts}, where the numbers between parentheses after each layout identifier indicate the relative frequencies of the layout specifications, and the numbers after the dashes are area weights for the rooms or groups of rooms within the corresponding parent groups in the hierarchy. The room types included in the layout specifications are \emph{kitchen}, \emph{bathroom}, \emph{living room}, and \emph{bedroom}.
\begin{table}
\renewcommand{\DTstyle}{\small\textrm\expandafter\raisebox{-0.7ex}}
\DTsetlength{1em}{0.5em}{0.3em}{0.4pt}{0.4pt}
\setlength{\DTbaselineskip}{0.8\baselineskip}
\hspace{-0.4cm}\begin{minipage}[t]{.47\linewidth}
\dirtree{%
.1 \textbf{8-room-3-bed (1/8)}.
.2 \textcolor{gray}{Group/4}.
.3 \textcolor{gray}{Group/2}.
.4 \textcolor{violet}{Kitchen/3}.
.4 \textcolor{olive}{Living room/3}.
.3 \textcolor{gray}{Group/1}.
.4 \textcolor{olive}{Living room/2}.
.4 \textcolor{teal}{Bathroom/1}.
.2 \textcolor{orange}{Bedroom/1}.
.2 \textcolor{orange}{Bedroom/1}.
.2 \textcolor{gray}{Group/2}.
.3 \textcolor{orange}{Bedroom/1}.
.3 \textcolor{teal}{Bathroom/1}.
}
\end{minipage}%
\begin{minipage}[t]{.47\linewidth}
\dirtree{%
.1 \textbf{7-room-3-bed (1/8)}.
.2 \textcolor{gray}{Group/3}.
.3 \textcolor{gray}{Group/2}.
.4 \textcolor{violet}{Kitchen/3}.
.4 \textcolor{olive}{Living room/3}.
.3 \textcolor{gray}{Group/1}.
.4 \textcolor{olive}{Living room/2}.
.4 \textcolor{teal}{Bathroom/1}.
.2 \textcolor{gray}{Group/2}.
.3 \textcolor{orange}{Bedroom/2}.
.3 \textcolor{orange}{Bedroom/2}.
.3 \textcolor{orange}{Bedroom/2}.
}
\end{minipage}
\vspace{0.2cm}
\hspace{-0.4cm}\begin{minipage}[t]{.47\linewidth}
\dirtree{%
.1 \textbf{2-bed-2-bath (1/8)}.
.2 \textcolor{gray}{Group/2}.
.3 \textcolor{orange}{Bedroom/2}.
.3 \textcolor{teal}{Bathroom/1}.
.2 \textcolor{gray}{Group/2}.
.3 \textcolor{orange}{Bedroom/2}.
.3 \textcolor{teal}{Bathroom/1}.
.2 \textcolor{gray}{Group/2}.
.3 \textcolor{violet}{Kitchen/3}.
.3 \textcolor{olive}{Living room/2}.
}
\end{minipage}%
\begin{minipage}[t]{.47\linewidth}
\dirtree{%
.1 \textbf{4-room (1/8)}.
.2 \textcolor{gray}{Group/2}.
.3 \textcolor{orange}{Bedroom/2}.
.3 \textcolor{teal}{Bathroom/1}.
.2 \textcolor{gray}{Group/2}.
.3 \textcolor{violet}{Kitchen/3}.
.3 \textcolor{olive}{Living room/2}.
}
\end{minipage}
\vspace{0.2cm}
\hspace{-0.4cm}\begin{minipage}[t]{.47\linewidth}
\dirtree{%
.1 \textbf{5-room (1/16)}.
.2 \textcolor{gray}{Group/2}.
.3 \textcolor{orange}{Bedroom/2}.
.3 \textcolor{teal}{Bathroom/1}.
.2 \textcolor{orange}{Bedroom/2}.
.2 \textcolor{gray}{Group/2}.
.3 \textcolor{violet}{Kitchen/3}.
.3 \textcolor{olive}{Living room/2}.
}
\end{minipage}%
\begin{minipage}[t]{.47\linewidth}
\dirtree{%
.1 \textbf{2-bed-1-bath (1/16)}.
.2 \textcolor{gray}{Group/2}.
.3 \textcolor{violet}{Kitchen/3}.
.3 \textcolor{teal}{Bathroom/2}.
.3 \textcolor{olive}{Living room/3}.
.2 \textcolor{orange}{Bedroom/1}.
.2 \textcolor{orange}{Bedroom/1}.
}
\end{minipage}
\vspace{0.2cm}
\hspace{-0.4cm}\begin{minipage}[t]{.47\linewidth}
\dirtree{%
.1 \textbf{kitchen-living-bed (1/16)}.
.2 \textcolor{gray}{Group/2}.
.3 \textcolor{violet}{Kitchen/3}.
.3 \textcolor{olive}{Living room/2}.
.2 \textcolor{orange}{Bedroom/1}.
}
\end{minipage}%
\begin{minipage}[t]{.47\linewidth}
\dirtree{%
.1 \textbf{kitchen-living-bed-2 (1/16)}.
.2 \textcolor{gray}{Group/2}.
.3 \textcolor{violet}{Kitchen/1}.
.3 \textcolor{olive}{Living room/1}.
.2 \textcolor{orange}{Bedroom/1}.
}
\end{minipage}
\vspace{0.2cm}
\hspace{-0.4cm}\begin{minipage}[t]{.47\linewidth}
\dirtree{%
.1 \textbf{bed-bath (1/16)}.
.2 \textcolor{orange}{Bedroom/2}.
.2 \textcolor{teal}{Bathroom/1}.
}
\end{minipage}%
\begin{minipage}[t]{.47\linewidth}
\dirtree{%
.1 \textbf{kitchen-living (1/16)}.
.2 \textcolor{violet}{Kitchen/1}.
.2 \textcolor{olive}{Living room/1}.
}
\end{minipage}
\vspace{0.2cm}
\hspace{-0.4cm}\begin{minipage}[t]{.47\linewidth}
\dirtree{%
.1 \textbf{kitchen (1/32)}.
.2 \textcolor{violet}{Kitchen/1}.
}
\end{minipage}%
\begin{minipage}[t]{.47\linewidth}
\dirtree{%
.1 \textbf{living (1/32)}.
.2 \textcolor{olive}{Living room/1}.
}
\end{minipage}
\vspace{0.2cm}
\hspace{-0.4cm}\begin{minipage}[t]{.47\linewidth}
\dirtree{%
.1 \textbf{bed (1/32)}.
.2 \textcolor{orange}{Bedroom/1}.
}
\end{minipage}%
\begin{minipage}[t]{.47\linewidth}
\dirtree{%
.1 \textbf{bath (1/32)}.
.2 \textcolor{teal}{Bathroom/1}.
}
\end{minipage}

\caption{\label{tab:layouts}\textbf{House generation layouts}. The 14 included layouts comprise large-sized, medium-sized and small-sized houses, including one-room houses. The relative frequency of each layout is shown between parentheses after each layout identifier.}
\end{table}
\vspace{0.4cm} \\
The resulting layout distribution covers large, medium, and small houses, as well as single-room houses. The average number of rooms per house is 4.5.
A total of 191,568 houses are sampled from this distribution, with ratio of $10{:}1{:}1$ across training, validation, and test. A sample house from each of the 14 layout specifications is shown in Fig.~\ref{fig:layout_distribution}.

\subsubsection{Assets}

\begin{figure}
    \centering
    \includegraphics[width=0.48\textwidth]{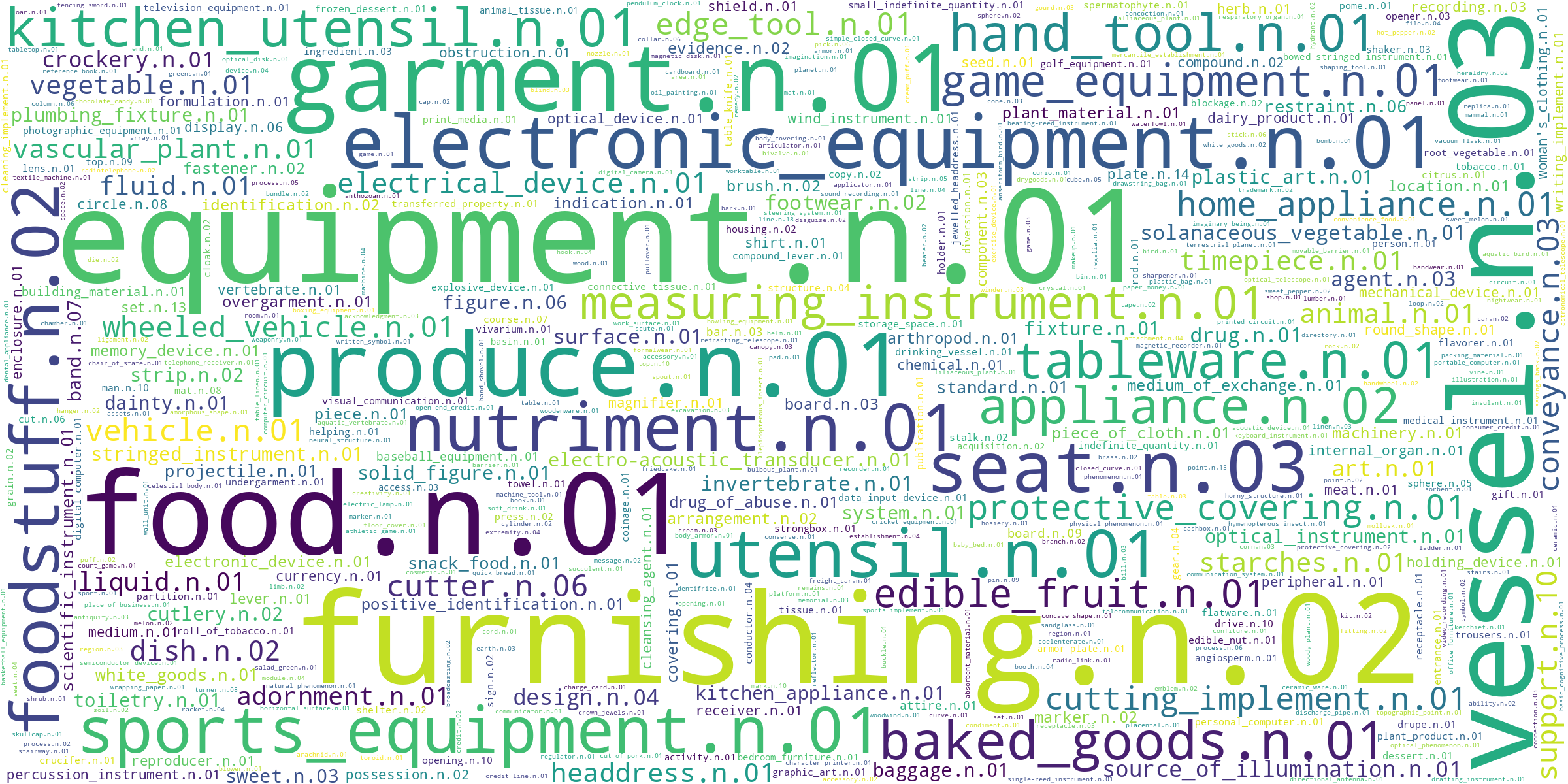}
    \caption{\textbf{Hypernyms of synsets in procedural houses}. Word map of hypernyms with font size scaled by the counts of corresponding hyponyms among the synsets in Fig.~\ref{fig:sunburst}.}
    \label{fig:all_hypernyms}
\end{figure}

Combining assets already present in THOR with a subset of high-quality assets from Objaverse \cite{Deitke2022ObjaverseAU}, an asset collection we call \objaversehome, 41,133 3D assets are available to be placed in procedural houses, including structural elements like doors and windows, large pieces of furniture and appliances, or small objects to be placed on free surfaces. For all \objaversehome assets, we extended the available annotations with scale, standard pose, descriptions, category, and nearest WordNet synset (from the 2022 version of the \emph{Open English WordNet} \cite{Fellbaum1998Wordnet,McCrae2019WordnetOpenSource}). The relative frequency of assets assigned to each synset is illustrated in Fig.~\ref{fig:sunburst}. The total number of synsets used to label the \objaversehome and THOR assets (equivalent to the number of object categories) is \numsynsets. In some tasks, we are interested in the agent being able to recognize more generic identifiers for a target object, which can be obtained in WordNet by retrieving hypernyms of the target object's labeled synset. In combination with the \numsynsets used to label the assets, and excluding too generic hypernyms like \emph{entity.n.01}, \emph{structure.n.01}, or \emph{product.n.02}, up to \numhypernyms synsets can be used to refer to the 41,133 assets, besides other open-vocabulary referring expression modalities like affordances or descriptions included in \benchnav. In addition to synsets shown in Fig.~\ref{fig:sunburst}, Fig.~\ref{fig:all_hypernyms} illustrates the relative frequency of hypernyms that may be used to refer to assets for tasks like \ObjectNavAffordance and \ObjectNavRelAttr. See Sec.~\ref{ref:goal_specifications} for details on how these synsets are used to specify targets for \benchnav. %

\paragraph{Affordances.}
 Using GPT-3.5, we extract five short descriptions of common usages for each synset, given the synset's definition, and a score indicating the confidence in the correctness of the usage. An example query is shown in Table~\ref{tab:collect_uses}. For our \numsynsets synsets, the total number of unique affordances we obtain is 4,315. Some examples are \emph{``Giving on special occasions"} for \emph{present.n.02}, \emph{``Cooking food on stovetop"} for \emph{burner.n.02}, or \emph{``Frozen summer treat on stick"} for \emph{ice\_lolly.n.01}.

\begin{table}
    \centering
    \footnotesize
    \begin{tabular}{p{0.9\linewidth}}
    \toprule
    Describe, in five words or less each, up to five common usages of an object with category \emph{``stockpot"} (definition: \emph{``a large pot for making stock, soup, or stew"}). Format each line of your response as\\
    $[\textrm{CONFIDENCE}]$ $[\textrm{USAGE DESCRIPTION}]$\\
    where $[\textrm{CONFIDENCE}]$ is a number between 0 and 10 indicating your confidence in the correctness of the usage (10 being most confident).\\
    \bottomrule
    \end{tabular}
    \caption{Example GPT query for synset uses.}
    \label{tab:collect_uses}
\end{table}

Once the collection of synset affordances is complete, we refine it by asking GPT-3.5 to determine whether each affordance can be reasonably associated to each \objaversehome asset labeled with the synset it generated from, this time conditioning on the asset's description, with a query like the example in Table \ref{tab:object_uses}. Please note that we do not perform this refinement step for THOR assets\footnote{In order to allow material randomization without compromising the validity of the descriptions and also taking into account the relative uniformity of THOR assets within their respective categories, we decided not to include descriptions for THOR assets.}.

\begin{table}
    \centering
    \footnotesize
    \begin{tabular}{p{0.9\linewidth}}
    \toprule
    For each of the following possible uses of an object with category \emph{aerosol}
    and description \emph{``A blue maya brand can of acrylic spray paint."}, indicate whether each use is a common use of the object. Format each line of your response as\\
    $[\textrm{INDEX}]$ $[\textrm{CONFIDENCE}]$\\
    where $[\textrm{CONFIDENCE}]$ is a number between 0 and 10 indicating your confidence in the correctness of the usage (10 being most confident) and $[\textrm{INDEX}]$ is the index of the usage.\\
    \emph{1. Applying hair spray.}\\
    \emph{2. Dispensing air freshener.}\\
    \emph{3. Dispensing fine particles.}\\
    \emph{4. Using spray paint.}\\
    \bottomrule
    \end{tabular}
    \caption{Example GPT query for object uses filtering.}
    \label{tab:object_uses}
\end{table}

Finally, we generate embeddings for each affordance using GPT-3's \emph{ada} \cite{Brown2020GPT} to create clusters of similar affordances, corresponding to sets of embeddings with cosine similarity $\geq$0.96. One example of such clusters would be the one composed by the affordances \emph{``Adding flavor to recipes"}, \emph{``Adding flavor to dishes"}, and \emph{``Enhancing the flavor of dishes"}. In total, we find 343 such clusters, with an average size of 2.78 affordances. This information, combined with the full set of affordances and synsets/assets they apply to (those for which the confidence score is $\geq$6) is used to sample \ObjectNavAffordance tasks as shown in Sec.~\ref{sec:sampling}. %

\subsection{Tasks}

\todo{

}

We define ten task types for domestic environments requiring the agent to learn and use different behaviors and understand different goal specifications. The behaviors included in our datasets and benchmark are navigation, pick up, fetch, and visit (all rooms in the house). Regarding goal specifications, we refer to goals by object types, affordances, instance descriptions, local references, room types, or relative attributes within a room.

The ten tasks we include in our benchmark serve as indicators of the agent's ability to execute different combinations of elemental navigation, exploration, manipulation, and vision-language skills.
Table \ref{tab:task_skills} summarizes the sets of elemental skills required to solve each of the tasks.

\begin{table}[t]
    \centering
    \footnotesize
    \newcommand\rot[1]{\multicolumn{1}{l}{\adjustbox{angle=90,lap=\width-1.0em}{\footnotesize{#1}}}}
    \begin{tabular}{rlccccccc}
         & &  \rot{Exploration} & \rot{Object identif.} & \rot{Set identif.} & \rot{Affordance identif.} & \rot{Feature identif.} & \rot{Manipulation} & \rot{Set completion}\\
         \toprule
         1 & \ObjectNavType & \checkmark & \checkmark &  &  &  &  & \\
         2 & \PickupType &  & \checkmark &  &  &  & \checkmark & \\
         3 & \FetchType & \checkmark & \checkmark &  &  &  & \checkmark & \\
         4 & \SimpleExploreHouse & \checkmark &  & \checkmark &  &  &  & \checkmark\\
         \midrule
         5 & \ObjectNavRoom & \checkmark & \checkmark & \checkmark &  &  &  & \\
         6 & \ObjectNavRelAttribute & \checkmark & \checkmark & \checkmark &  &  &  & \\
         7 & \ObjectNavAffordance & \checkmark & \checkmark &  & \checkmark &  &  & \\
         8 & \ObjectNavLocalRef & \checkmark & \checkmark & \checkmark &  &  &  & \\
         9 & \ObjectNavOpenVocab & \checkmark & \checkmark &  &  & \checkmark &  & \\
         10 & \RoomNav & \checkmark &  & \checkmark &  &  &  & \\
         \bottomrule
    \end{tabular}    \caption{\label{tab:task_skills}\textbf{Skill sets required for all task types}. Different sets of skills are required to solve each task. \emph{Set identification} encompasses the ability to recognize a room type as well as explicit local configurations or relative attributes of objects within a room.}
\end{table}

\subsubsection{Goal specifications}
\label{ref:goal_specifications}

The \emph{navigate} behavior is also combined with a diversity of goal specifications in \benchnav, eliciting the agent to identify goal objects through different referring expression types.

Please note that we select target object types by their annotated synset among the ones in the \emph{Open English WordNet} \cite{Fellbaum1998Wordnet,McCrae2019WordnetOpenSource}, which in practice falls back to choosing the lowest hypernym for the sampled synset shared by any objects present in the scene\footnote{Despite an effort to annotate all objects in the scene with the most precise possible synset, more generic hypernyms are sometimes used.}. For example, an \emph{armoire} asset coarsely labelled as \emph{furniture.n.01}, can coexist in a house with a chair labeled as \emph{chair.n.01} and a table labeled as \emph{table.n.02}, so any of three assets shall be referred to by \emph{furniture.n.01}, which is the lowest common hypernym.

\paragraph{{\sc{\textbf{ObjectType}}} goal.} The natural language instruction given to the agent refers to the most precise lemma for the chosen synset (\ie the one used less often by other synsets). Any object in the scene that can be addressed by the lemma in the instruction is a valid goal for the task.

\paragraph{{\sc{\textbf{Object-Room}}} goal.} Similar to \emph{object type}, but further constraining the valid goal objects to be present at the beginning of the task in a room of a specific type. This specification type implies that more instances of the target type also exist in at least one room of any of the other types.

\paragraph{{\sc{\textbf{RelativeAttribute}}} goal.} We consider six relative attributes regarding objects of some given type in a specific room type: \emph{highest/lowest}, \emph{largest/smallest}, and \emph{nearest/furthest} to/from some anchor type (for which a unique object exists in the room), implying the presence of more than one object of the target type in the specified room.

\paragraph{{\sc{\textbf{Affordance}}} goal.} For affordances, we provide the agent with a hypernym covering a subset of the objects in the scene such that some of the objects in the subset are plausible candidates to provide the requested affordance while some others might not, \ie, the agent must be able to determine which objects correspond to the given hypernym, and also which of these are also providing the requested affordance.

\paragraph{{\sc{\textbf{LocalReference}}} goal.} We consider two types of local reference specifications: (1) an object of given type near two other objects of other types, and (2) an object of given type on top of an object of another type. In either case, this task specification implies than more than one object of the target type exists in the house.

\paragraph{{\sc{\textbf{Description}}} goal.} In this case, the specification is an open-vocabulary description of the target object type, assuming uniqueness of the valid target, meaning that no other object in the house responds to the given description. All \ObjectNavOpenVocab tasks use \objaversehome assets as targets, since we do not include asset descriptions for THOR assets.

\paragraph{{\sc{\textbf{Room}}} goal.} In \RoomNav, the goal is just a room type, which can correspond to one or more of the rooms in the house.

\subsection{Sampling Methods}
\label{sec:sampling}

We first describe a set of criteria applied onto potential targets to decide upon their acceptance as task goals, and then proceed to outline each task sampler. %

\paragraph{Filtering.} In order for objects to be valid targets, we impose certain constraints dependent on how the agent needs to interact with the object. For example, in \textbf{navigation} tasks, we impose that the target objects are below a maximum height of 1.1 m, that a path can be computed by THOR to an interactable location near the objects, and that the bounding boxes for the objects have (1) a largest face with diagonal larger than 10 cm; and (2) a middle dimension of at least 4 cm.
In \textbf{manipulation} tasks, we additionally impose that the object can actually be picked up and the maximum dimension of its bounding box is less than 50 cm.

\paragraph{\SimpleExploreHouse sampling.} The agent is just spawned at a random location within a room sampled uniformly among the house rooms.

\paragraph{\ObjectNav sampling.} We attempt to balance the distribution of target synsets by keeping counters of the number of times we have sampled each synset and sequentially selecting candidate synsets from a random sequence of synsets available in the scene sampled with replacement with  weights given by the inverse of the respective counts. If any of the scene objects with the sampled synset passes the \emph{navigation} filters described above, we accept the task. The agent is then spawned as in \SimpleExploreHouse sampling.

\paragraph{\Fetch sampling.} Similar to \ObjectNav sampling, but using the \emph{manipulation} filter, also enforcing the object to be valid for pickup.

\paragraph{\PickUp sampling.} Similar to \Fetch sampling, but with the agent spawned at a random location where any of the target objects is at interactable distance, and oriented such that the manipulation camera is aligned with the object's center.

\paragraph{\ObjectNavRoom sampling.} Similar to \ObjectNav sampling, but enforcing that the target synset has multiple instances in the house, and that all instances of the target synset for the target room type are in a single room, besides having at least one instance of the target type in any rooms of any other type. If several rooms are feasible, we randomly pick one.

\paragraph{\ObjectNavRelAttr sampling.} In addition to the target type count in \ObjectNav sampling, we also keep counts of the number of times we sample different relative attribute types (among \emph{smallest, largest, highest, lowest, nearest to, furthest from}). Similar to \ObjectNavRoom sampling, we enforce that the target synset can be found in a specific room type, but this time without enforcing that the object is also present in a different room type. We also enforce that multiple instances of the target synset are present in a single instance of the room type. Then, we try relative attribute types prioritizing by lowest counters.
For \emph{smallest} (\emph{largest}), we accept the task if the smallest (largest) object of type in the room has a bounding box with diagonal at least twice (at most half) of the second largest (smallest) object of given synset in the room.
For \emph{highest}, we check that the bottom of the bounding box of the highest object of given synset in the room is the highest placed among the objects of the given synset, and also that the top of the second highest box is less than half way between the bottom and top of the highest one. The criterion for \emph{lowest} is the reciprocal.
For \emph{furthest from}, we first extract \emph{anchor} objects in each room, which can be \emph{beds, counter tops, dining tables, fridges, sinks, sofas, televisions,} or \emph{toilets}, with the condition that only one instance exists in the room. Then, for the given synset, we check whether the bounding box distance from the second furthest object from the anchor is less than 70\% of the distance of the furthest one. The criterion for \emph{nearest to} is built similarly.
If the target object (and potentially the anchor) pass the \emph{navigation} filters, the task is accepted.

\paragraph{\ObjectNavAffordance sampling.} We constrain target synsets to be hypernyms of the synsets in the scene and sample the available hypernyms following the same procedure as in \ObjectNav sampling. For each synset in the scene child of the target hypernym, we collect all confident affordances and merged affordances from the clusters defined above, and list the objects in the scene that provide each affordance. We then randomly sample an affordance and check that at least one of the objects providing the affordance pass the \emph{navigation} filter to accept the task.

\paragraph{\ObjectNavLocalRef sampling.} This is similar to \ObjectNavRelAttr sampling, but having two possible modes: \emph{near} two reference synsets and \emph{on} one reference synset.
For \emph{near}, we search for two additional synsets such that two instances in the house have bounding boxes at a distance $<$50 cm from the bounding box of the current target synset, such that no other such synset triplet is present in the scene and no other such triplet is present in the scene with bounding box distances $<$2 m, to avoid ambiguity. We constrain the reference instances to not be any of \emph{floors, walls, doors, windows, shelves, drawers, beds, counter tops} or \emph{sofas}.
For \emph{on}, we search for a unique combination of the target synset lying on an instance of a reference synset, which we constrain not to be any of \emph{floors, walls, doors, windows, shelves} or \emph{drawers}.
In both cases, the target object and the references must pass the \emph{navigation}  filters.

\paragraph{\ObjectNavOpenVocab sampling.} Similar to \ObjectNav sampling, but constraining the assets for the sampled synset to be part of \objaversehome (\ie, containing a natural language description), and ensuring that a single instance of the sampled asset is present in the scene. The \emph{navigation} filter must be passed by the target object.

\paragraph{\RoomNav sampling.} In this case we keep counts of the number of times we sample different room types, and sample prioritized room types.

\subsection{Planners}
\label{sec:planners}

We define planners able to solve each of the task types with a two-fold goal: (1) generate expert trajectories for supervised learning; and (2) validate the feasibility of a sampled task specification. %
We first define two subroutines for navigation and manipulation that are reused by several planners and then proceed to outline each planner.
Unless otherwise specified, if any phase in the planner fails to complete during the execution of the planner, the trajectory (and task) is discarded. 

\paragraph{Navigation subroutine.} This subroutine allows the agent to approximately follow a shortest path to a target (noting that the actual shortest path is not necessarily feasible given the discrete set of actions available for the agent). It allows to recompute the path if the agent gets stuck given to discretization errors.

\paragraph{Pick-up subroutine.} This subroutine assumes the target object is at interactable distance with the agent's manipulation camera aligned. It executes the following steps: (1) first move the agent arm up until the target object's height; (2) rotate the wrist until the grabber is oriented away from the agent's body; (3) move the arm out until reaching the object's depth; (4) rotate the wrist until the gripper is near the target object; (5) move the arm down until the grabber is in contact with the object; (6) close the gripper; (7) slightly lift the arm with the object in hand. The pickup subroutine also includes an option to retry pickup in case of either failure to grab the object during simulation or random choice (20\% probability). This enables the agent to learn local corrections for small deviations from the gold trajectory.

\paragraph{{\sc{\textbf{Navigate}}} planner.} This planner is used by all tasks in \benchnav with the exception of \RoomNav, \ie, for navigation with all possible goal specifications other than \emph{room} goal.
Given the agent spawned in a random location within the house and a list of valid target object identifiers in the scene, the planner selects the nearest target in terms of geodesic distance, and executes the navigation subroutine on the chosen path. Once the final position is reached, the planner proceeds to align the navigation camera with the target object and signal task termination.

\paragraph{{\sc{\textbf{RoomNav}}} planner.} Similar to navigate, the planner determines the nearest room center in terms of geodesic distance and approximately follows the nearest path to the center of the nearest room of the given type via the navigation subroutine. Once the center is reached, the planner signals task termination.

\paragraph{{\sc{\textbf{PickUp}}} planner.} The first phase executes navigation to an interactable location for the nearest target object. Then, the agent rotates to align the manipulation camera with the target object and proceeds to execute the pick-up subroutine (with at most one failed pickup phase). Once the object is in the gripper, the planner signals task termination.

\paragraph{{\sc{\textbf{Fetch}}} planner.} The fetch planner assumes the target object is in view, and initially allows to move the base closer to the target (as in regular navigation), and then proceeds with the pick-up subroutine described above before singnaling task termination.

\paragraph{{\sc{\textbf{RoomVisit}}} planner.} This planner builds a traversal of the rooms in the house via depth first search and approximately follows the path to each room's center using the navigation subroutine. After reaching each room's center a subtask completion action is issued, followed by a task termination signal upon visitng the last room in the house.

\subsection{Exploration behavior}
\label{sec:exploration}
Our exploration planner for investigating shortest-path usability searches rooms in a depth-first manner, as in our \SimpleExploreHouse planner, with one addition: instead of simply navigating to the center of the room, it iteratively navigates toward unseen objects in its current room until at least 75\% of objects in that room have been seen, at which point it searches the next room for its target object. Our expert exploration is object-centric - it has access to and uses substantially more privileged information. When a target object is seen during this process, it immediately navigates towards it. This is in contrast to Frontier-Based Exploration \cite{ramrakhya2023pirlnav,yamauchi1997frontier}, where empty sections of an explicit map are filled greedily. 

\subsection{Benchmark data}

\begin{table}
\centering
\scriptsize
\setlength{\tabcolsep}{4pt}
\begin{tabular}{|c|c|l|rrr|rr|}
\hline
\multirow{2}{*}{\textbf{Benchmark}} & \multirow{2}{*}{\textbf{Categories}}& \multirow{2}{*}{\textbf{Tasks}} & \multicolumn{3}{c|}{\textbf{Train}}                                                  & \multicolumn{2}{c|}{\textbf{Eval.}}                        \\
                        &         &                & \multicolumn{1}{c}{Hs.}  & \multicolumn{1}{c}{Ep.}  & \multicolumn{1}{c|}{Fr.} & \multicolumn{1}{c}{Hs.} & \multicolumn{1}{c|}{Ep.} \\ \hline
\multirow{4}{*}{\bench\fifteen}   &   \multirow{4}{*}{15} & \ObjectNav                 & \multicolumn{1}{c}{10K} & \multicolumn{1}{c}{99K} & \multicolumn{1}{c|}{5M} & \multicolumn{1}{c}{200}   & \multicolumn{1}{c|}{200}   \\
                    &    & \PickUp                 & \multicolumn{1}{c}{8K}  & \multicolumn{1}{c}{65K} & \multicolumn{1}{c|}{3M} & \multicolumn{1}{c}{171}   & \multicolumn{1}{c|}{171}   \\
                    &    & \Fetch                  & 13K                     & 114K                    & 12M                     &             172           &      172                   \\
                    &    & \SimpleExploreHouse              & 9K                      & 94K                     & 20M                     &            200            &          200               \\ \hline
\multirow{4}{*}{\bench\alltype}  & \multirow{4}{*}{\numsynsets}    & \ObjectNav                 & 10K                     & 99K                     & 6M                      &            216            &        216                 \\
                    &    & \PickUp                 & 10K                     & 92K                     & 4M                      &            173            &       173                  \\
                    &    & \Fetch                  & 9K                      & 85K                     & 9M                      &            179            &           179              \\
                    &    & \SimpleExploreHouse              & 9K                      & 94K                     & 20M                     &            200            &           200              \\ \hline
\end{tabular}
\caption{Number of unique houses, episodes, and frames (training only) in \bench.
}
\label{tab:chores_stats}
\vspace{-1em}
\end{table}

\begin{table}
\scriptsize
\setlength{\tabcolsep}{2pt}
\centering
\begin{tabular}{|c|c|l|rrr|rr|}
\hline
\multirow{2}{*}{\textbf{Benchmark}} & \multirow{2}{*}{\textbf{Categories}}& \multirow{2}{*}{\textbf{Tasks}} & \multicolumn{3}{c|}{\textbf{Train}}                                                  & \multicolumn{2}{c|}{\textbf{Eval.}}                        \\
                        &         &                & \multicolumn{1}{c}{Hs.}  & \multicolumn{1}{c}{Ep.}  & \multicolumn{1}{c|}{Fr.} & \multicolumn{1}{c}{Hs.} & \multicolumn{1}{c|}{Ep.} \\ \hline
\multirow{7}{*}{\benchnav\fifteen} & \multirow{7}{*}{15} & \ObjectNavType & 10K & 99K & 5M & 200 & 200 \\
 &  & \ObjectNavRoom & 2K & 21K & 1M & 181 & 181 \\
 &  & \ObjectNavRelAttribute & 1K & 13K & 1M & 194 & 194 \\
 &  & \ObjectNavAffordance & 2K & 23K & 1M & 197 & 197 \\
 &  & \ObjectNavLocalRef & 2K & 19K & 2M & 142 & 142 \\
 &  & \ObjectNavOpenVocab & 1K & 13K & 1M & 144 & 144 \\
 &  & \RoomNav & 2K & 22K & 1M & 200 & 200 \\
 \hline
\multirow{7}{*}{\benchnav\alltype} & \multirow{7}{*}{\numsynsets} & \ObjectNavType & 10K & 99K & 6M & 216 & 216 \\
 &  & \ObjectNavRoom & 10K & 93K & 6M & 156 & 156 \\
 &  & \ObjectNavRelAttribute & 9K & 91K & 8M & 217 & 217 \\
 &  & \ObjectNavAffordance & 10K & 99K & 5M & 228 & 228 \\
 &  & \ObjectNavLocalRef & 9K & 92K & 7M & 254 & 254 \\
 &  & \ObjectNavOpenVocab & 10K & 95K & 6M & 268 & 268 \\
 &  & \RoomNav & 9K & 87K & 5M & 200 & 200 \\
\hline
\end{tabular}

\caption{Number of unique houses, episodes, and frames (training only) in \benchnav.}
\label{tab:choresnav_stats}
\end{table}

\begin{figure*}
    \centering
    \begin{tabular}{m{0.15\textwidth}m{0.35\textwidth}m{0.35\textwidth}}
     & \multicolumn{1}{c}{\textbf{Train}} & \multicolumn{1}{c}{\textbf{Evaluation}} \\
     \bench\fifteen &
     \includegraphics[width=0.35\textwidth]{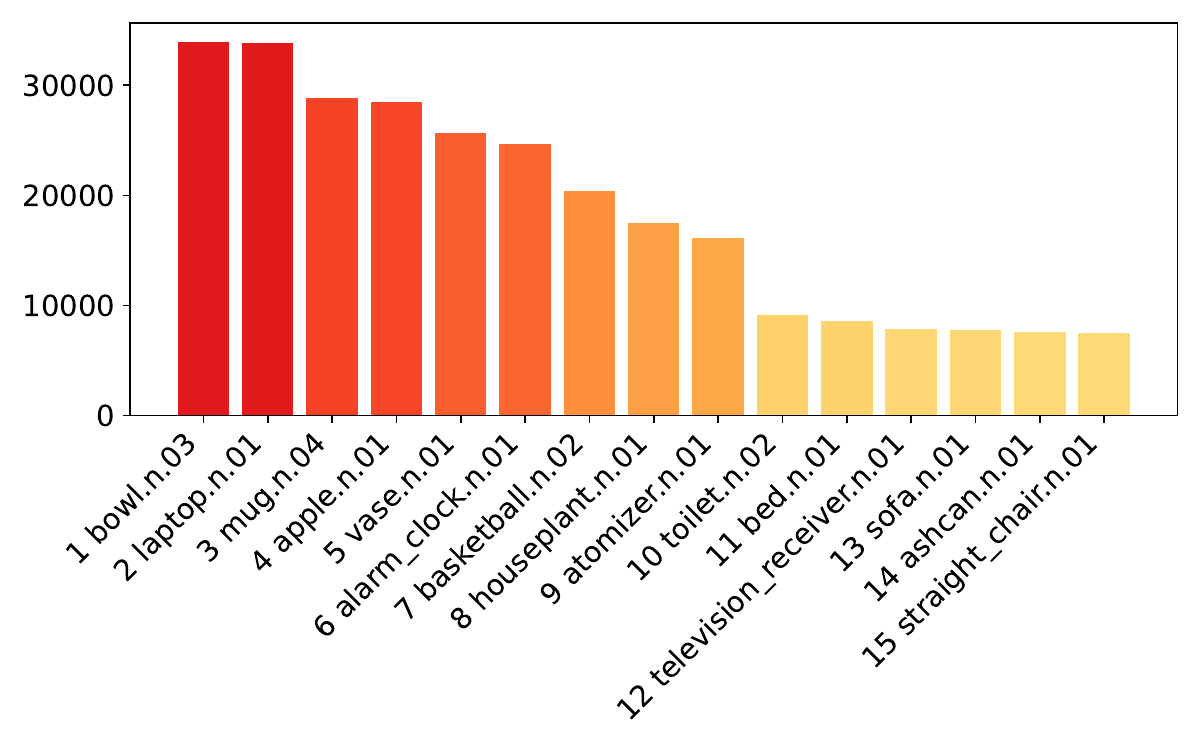} &
     \includegraphics[width=0.35\textwidth]{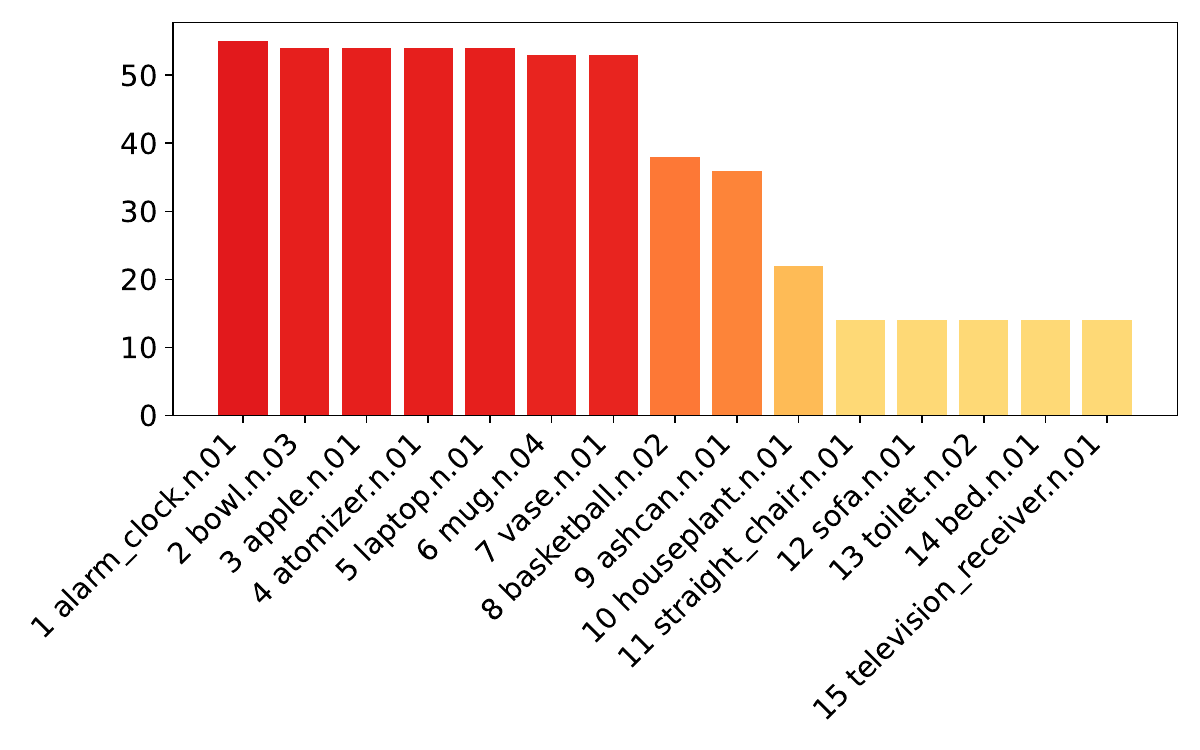} \\
     \bench\alltype &
     \includegraphics[width=0.35\textwidth]{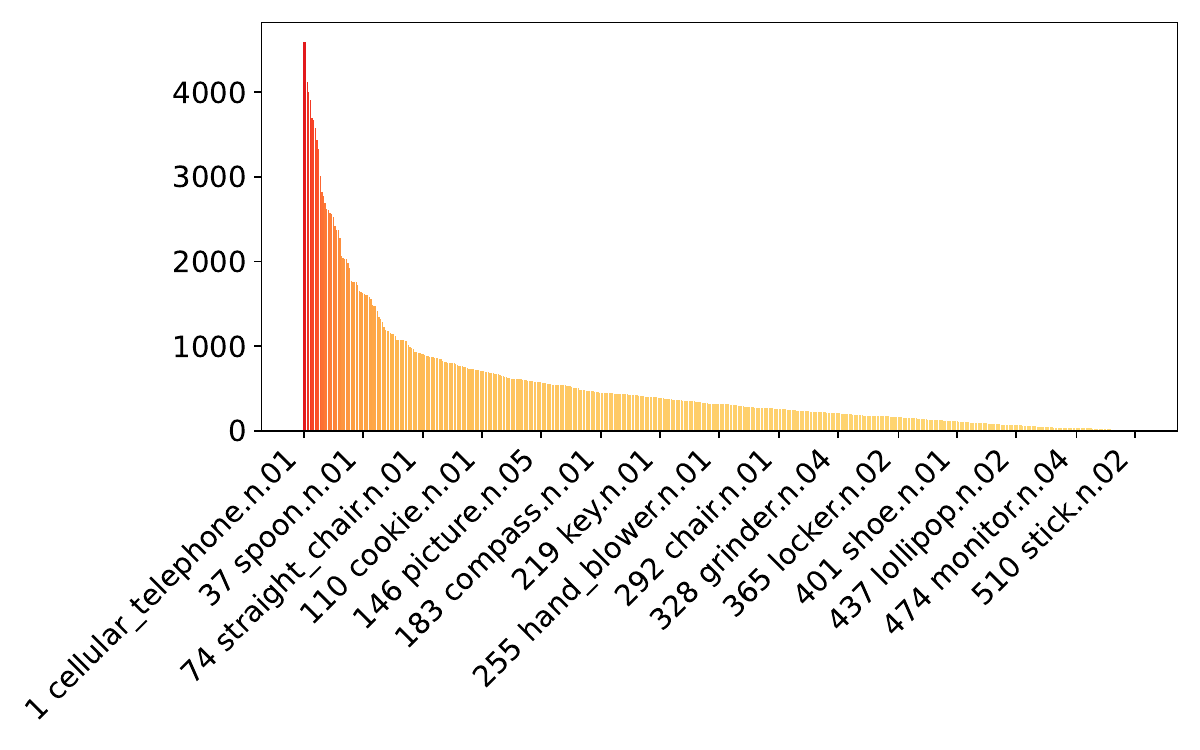} &
     \includegraphics[width=0.35\textwidth]{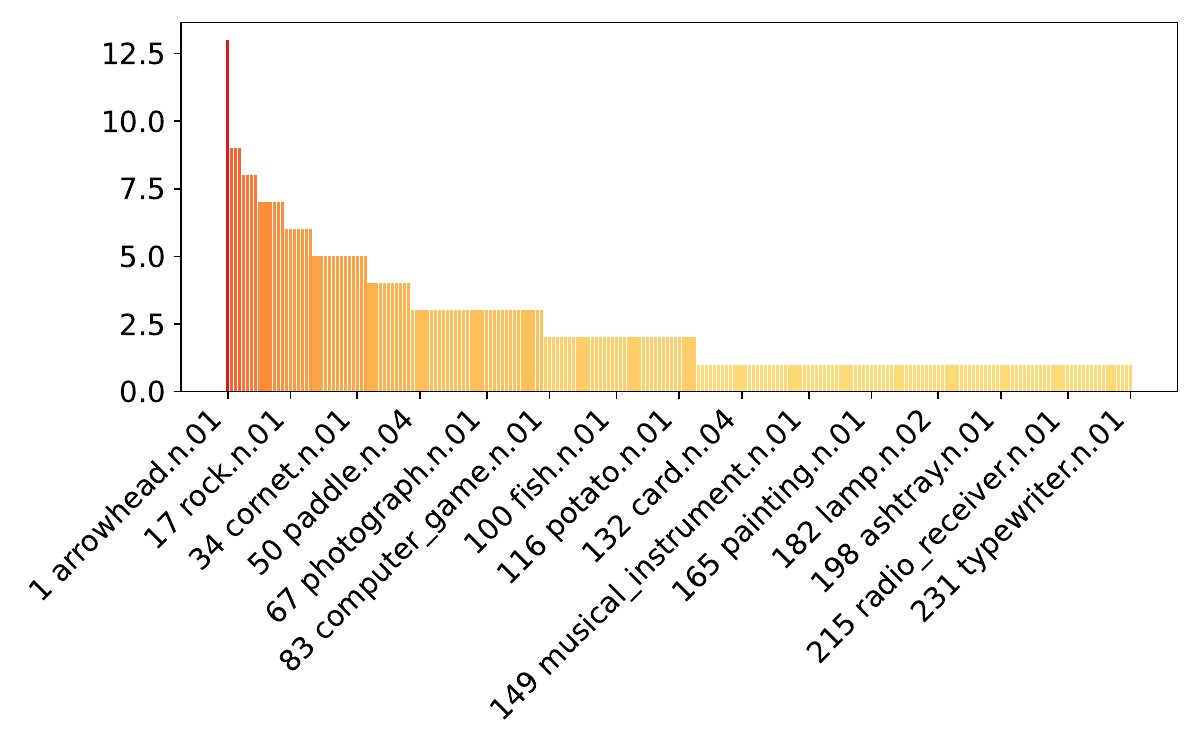} \\
     \benchnav\fifteen &
     \includegraphics[width=0.35\textwidth]{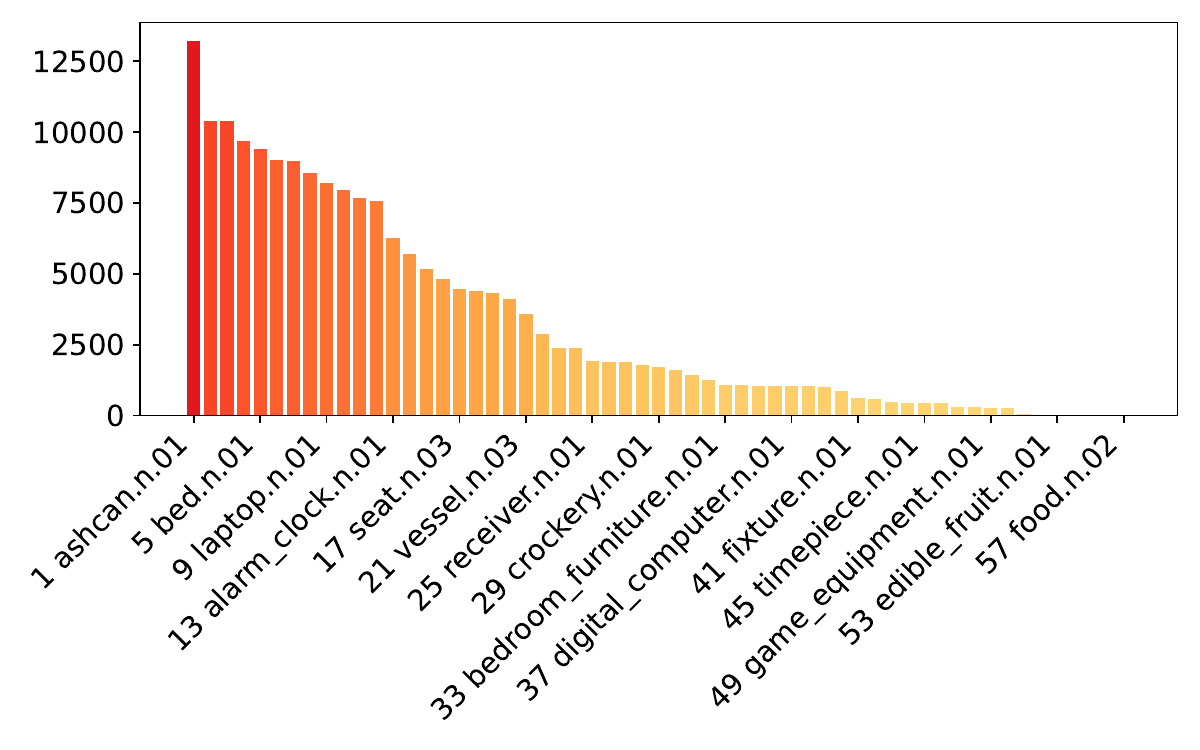} &
     \includegraphics[width=0.35\textwidth]{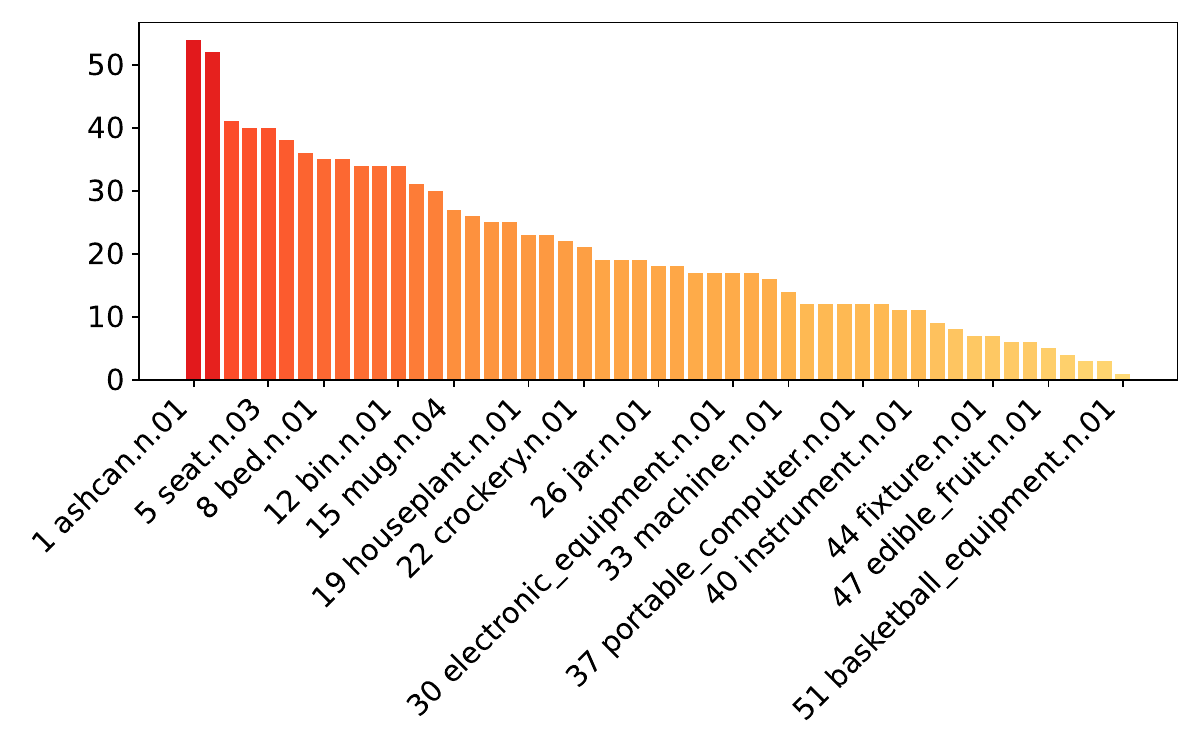} \\
     \benchnav\alltype &
     \includegraphics[width=0.35\textwidth]{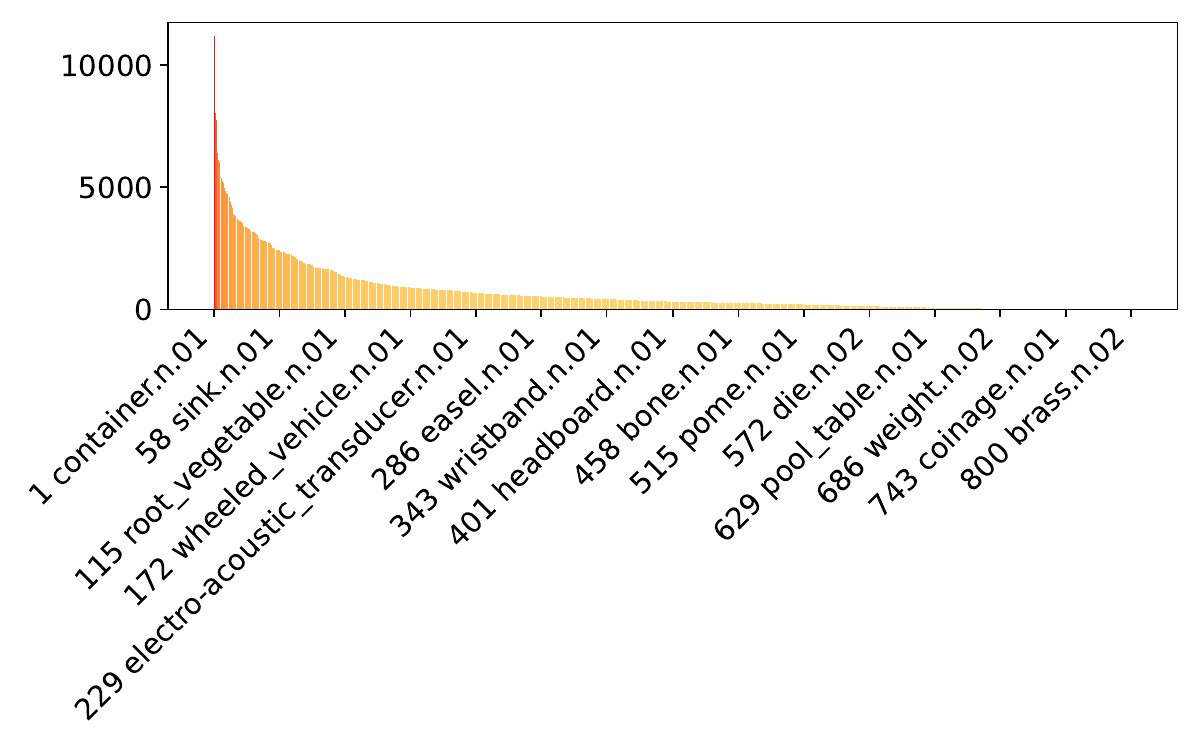} &
     \includegraphics[width=0.35\textwidth]{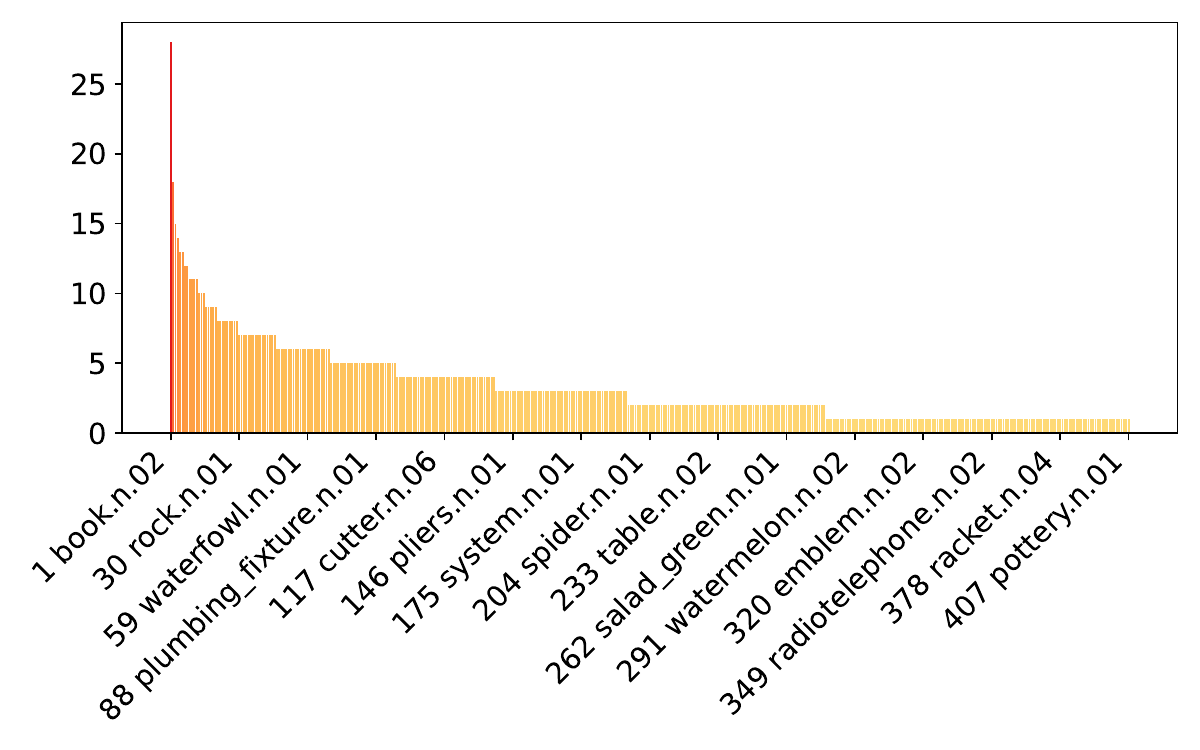} \\
    \end{tabular}
    \caption{\textbf{Frequency of target synsets} for \bench\fifteen, \bench\alltype, \benchnav\fifteen, and \benchnav\alltype }
    \label{fig:target_synsets}
\end{figure*}

\begin{figure}
    \footnotesize
    \centering
    \begin{tabular}{c}
    \bench\fifteen \\
    \includegraphics[width=0.8\linewidth]{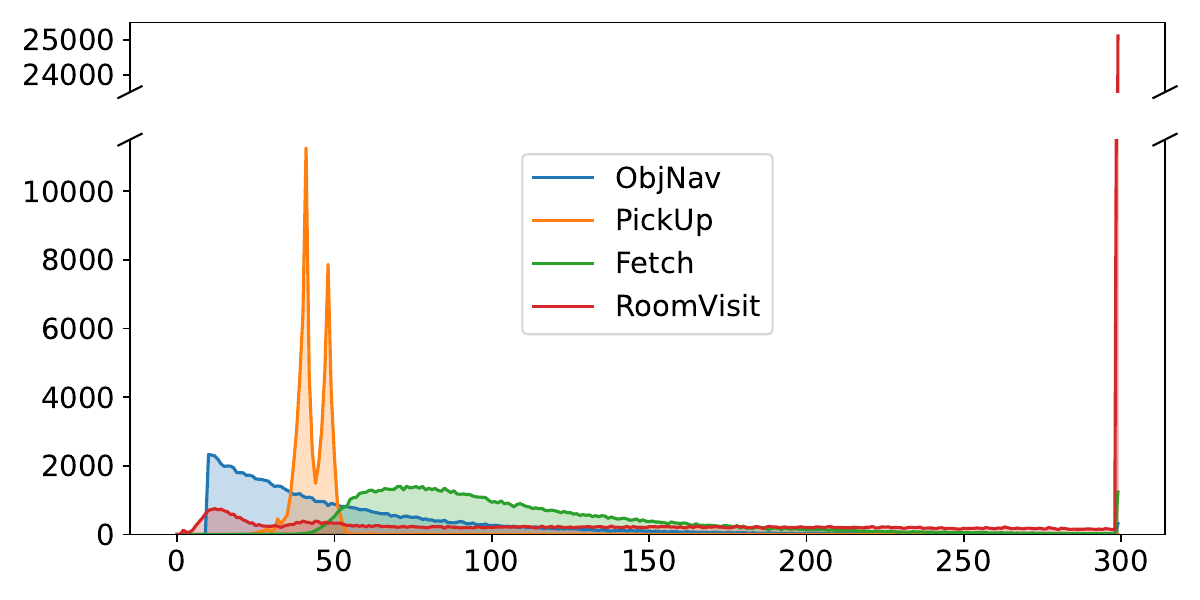} \\
    \bench\alltype \\    
    \includegraphics[width=0.8\linewidth]{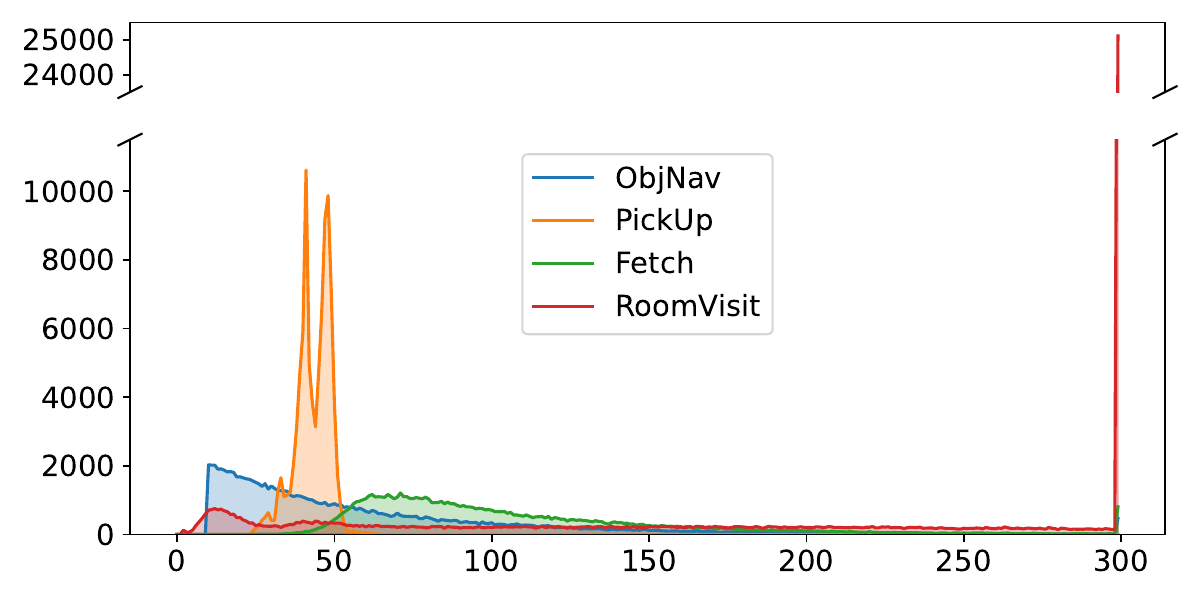} \\
    \benchnav\fifteen \\
    \includegraphics[width=0.8\linewidth]{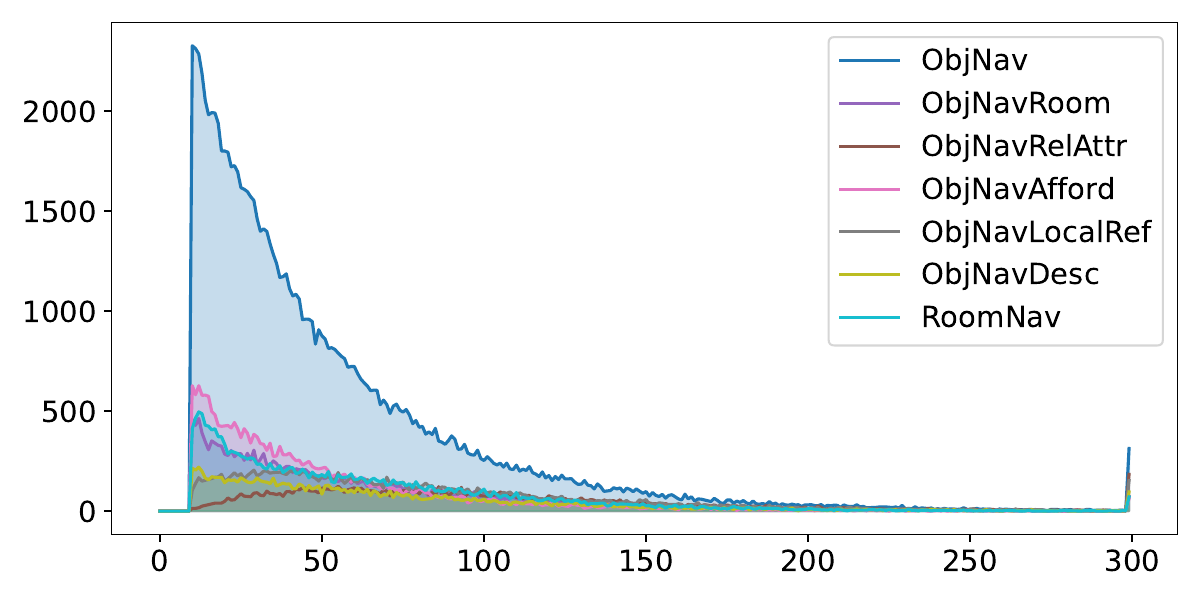} \\
    \benchnav\alltype \\    
    \includegraphics[width=0.8\linewidth]{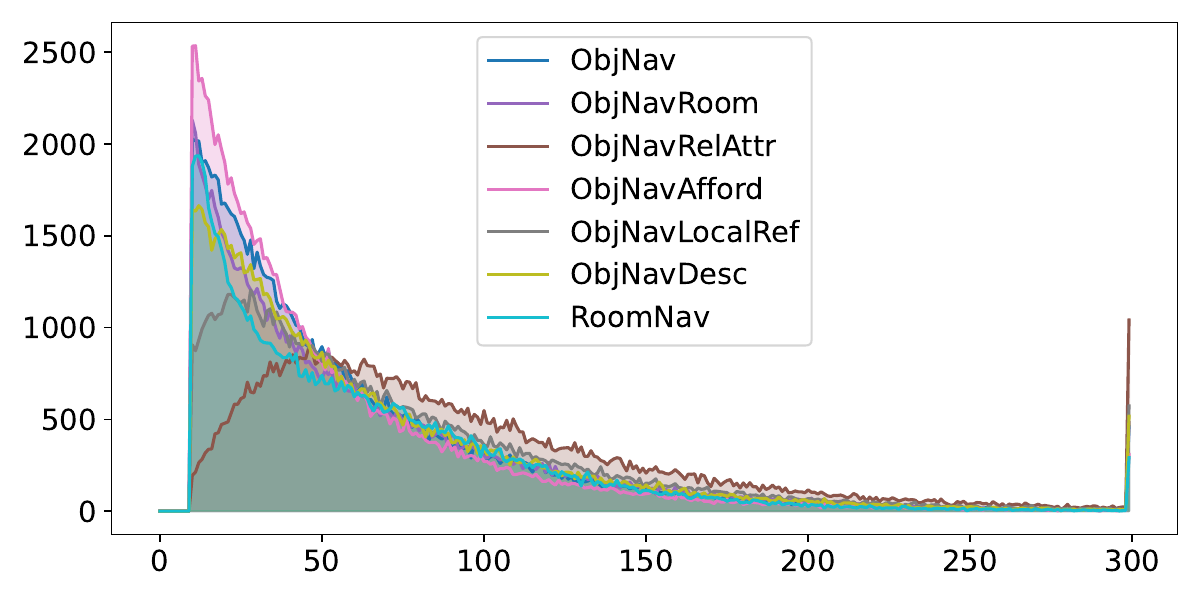} \\
    \end{tabular}
    \caption{\textbf{Distribution of planner trajectory lengths} in training episodes for all task types. \emph{x axis:} trajectory lengths; \emph{y axis:} frequency. For plot legibility, we have clamped all episode lengths to have a max
    value of 300 in these plots (in training, these episodes can be up to 1000 steps long), this explains the spike at length 300 for some tasks.}
    \label{fig:eplengths}
\end{figure}

\begin{figure*}
    \centering
    \footnotesize
    \begin{tabular}{m{0.12\textwidth}m{0.35\textwidth}m{0.35\textwidth}}
     & \multicolumn{1}{c}{\textbf{Train}} & \multicolumn{1}{c}{\textbf{Evaluation}} \\
     \bench\fifteen &
     \includegraphics[width=0.35\textwidth]{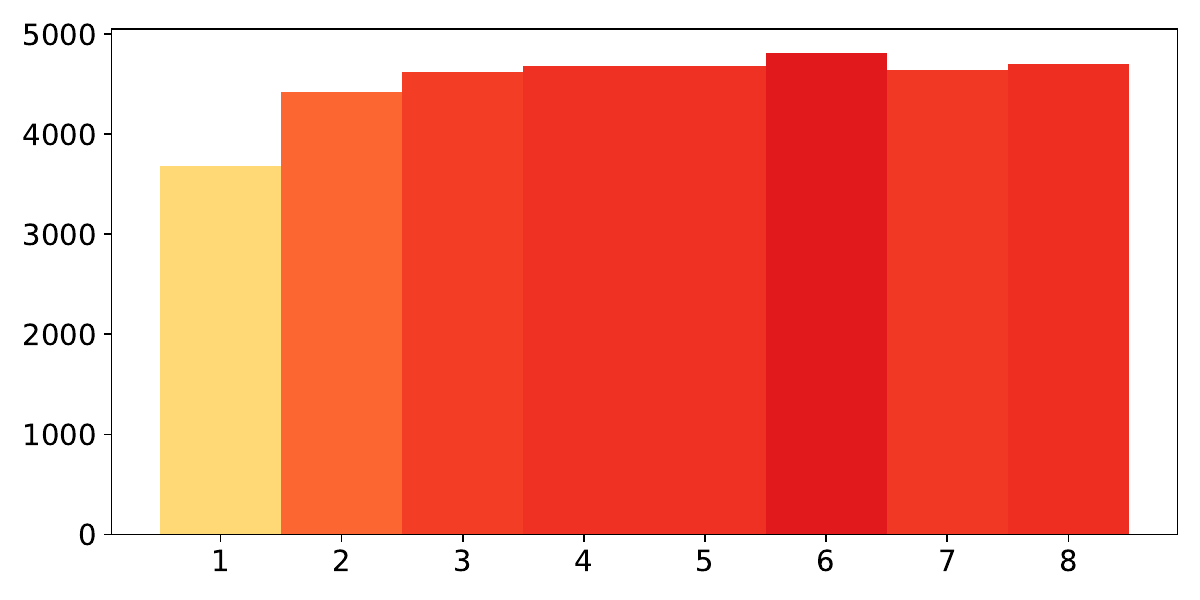} &
     \includegraphics[width=0.35\textwidth]{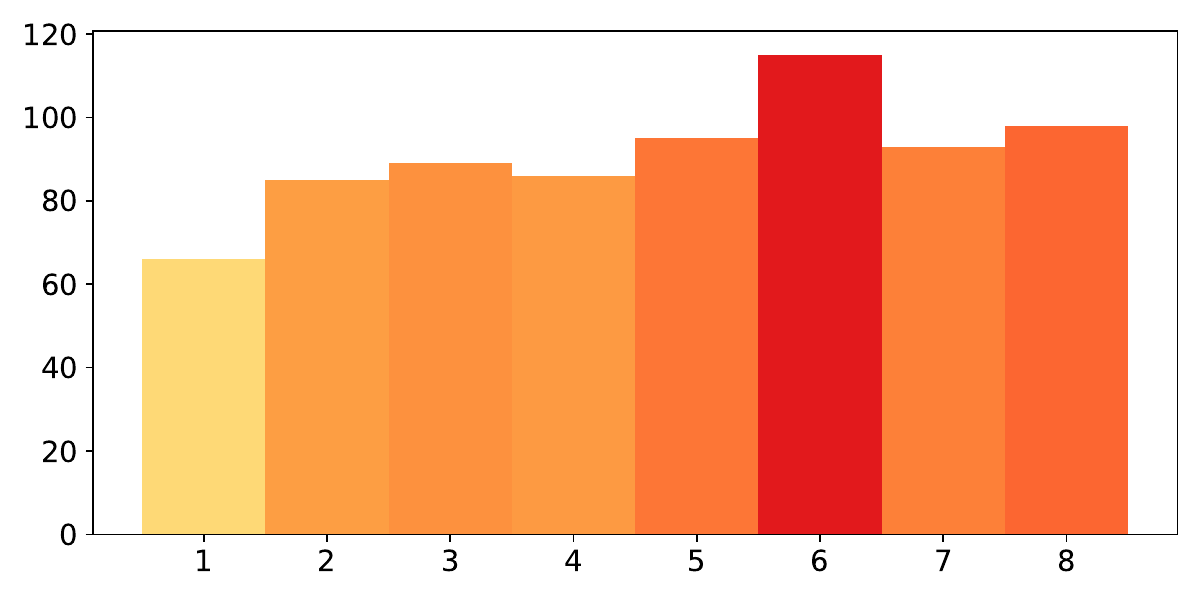} \\
     \bench\alltype &
     \includegraphics[width=0.35\textwidth]{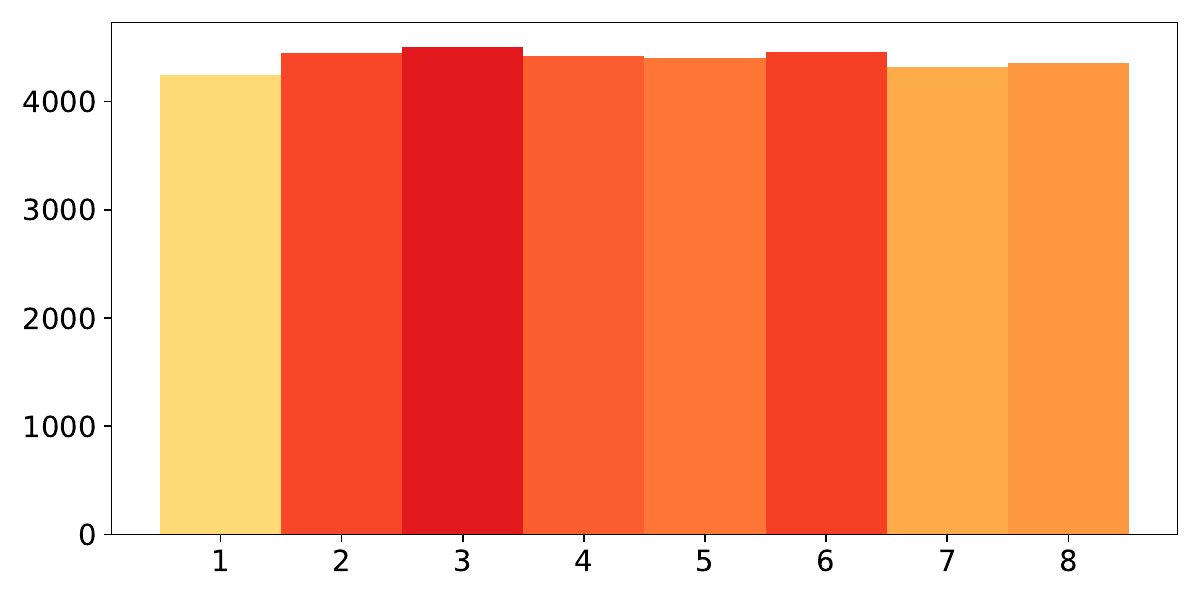} &
     \includegraphics[width=0.35\textwidth]{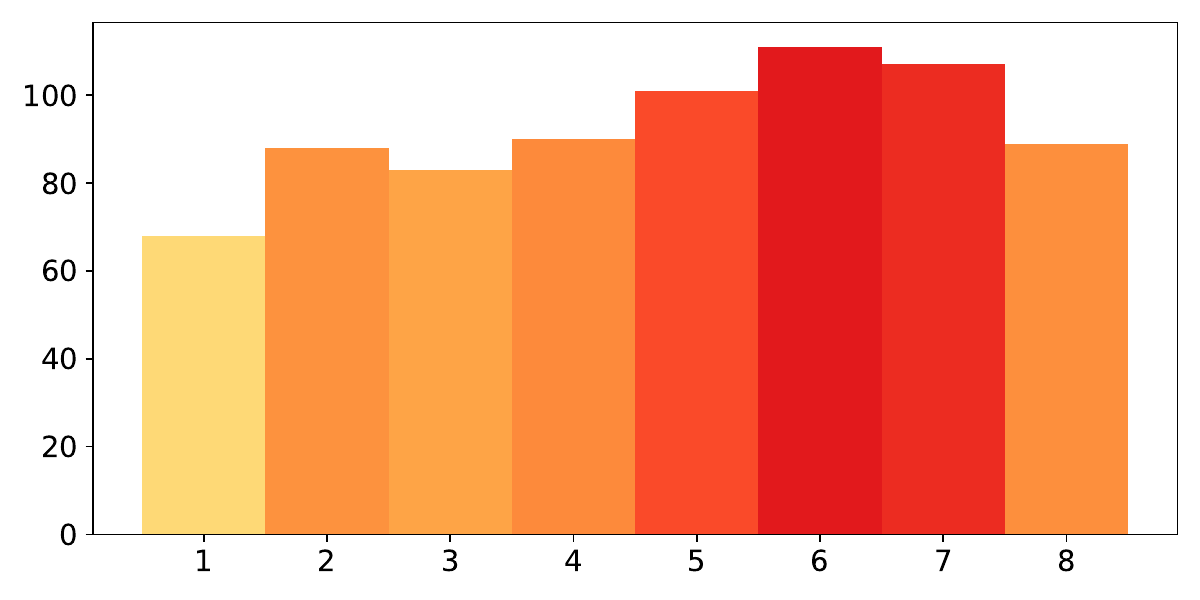} \\
     \benchnav\fifteen &
     \includegraphics[width=0.35\textwidth]{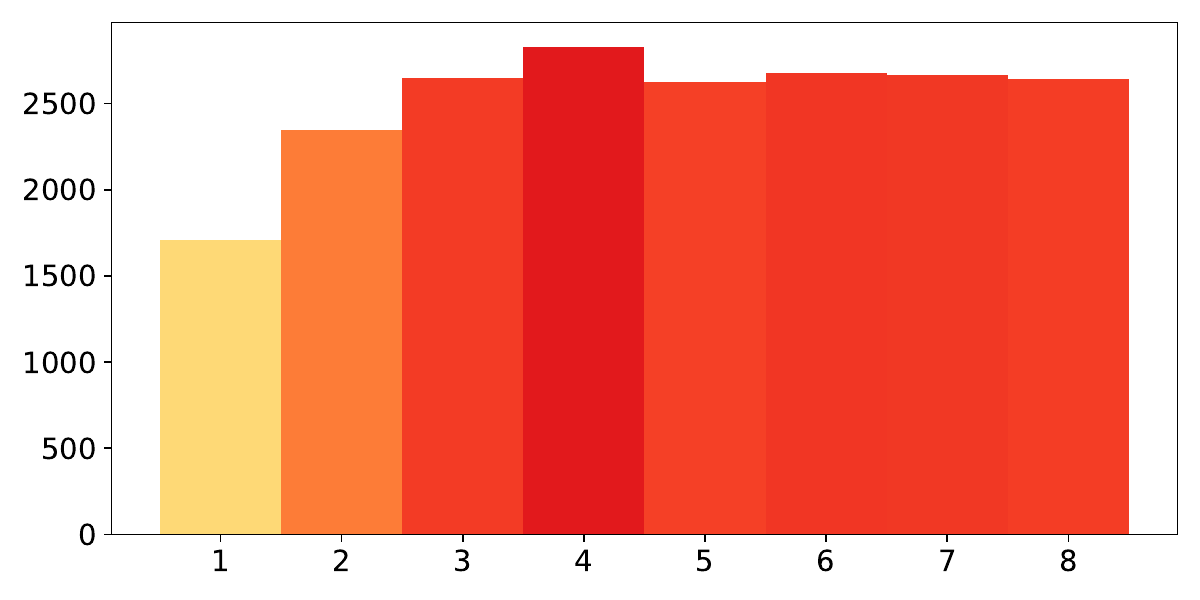} &
     \includegraphics[width=0.35\textwidth]{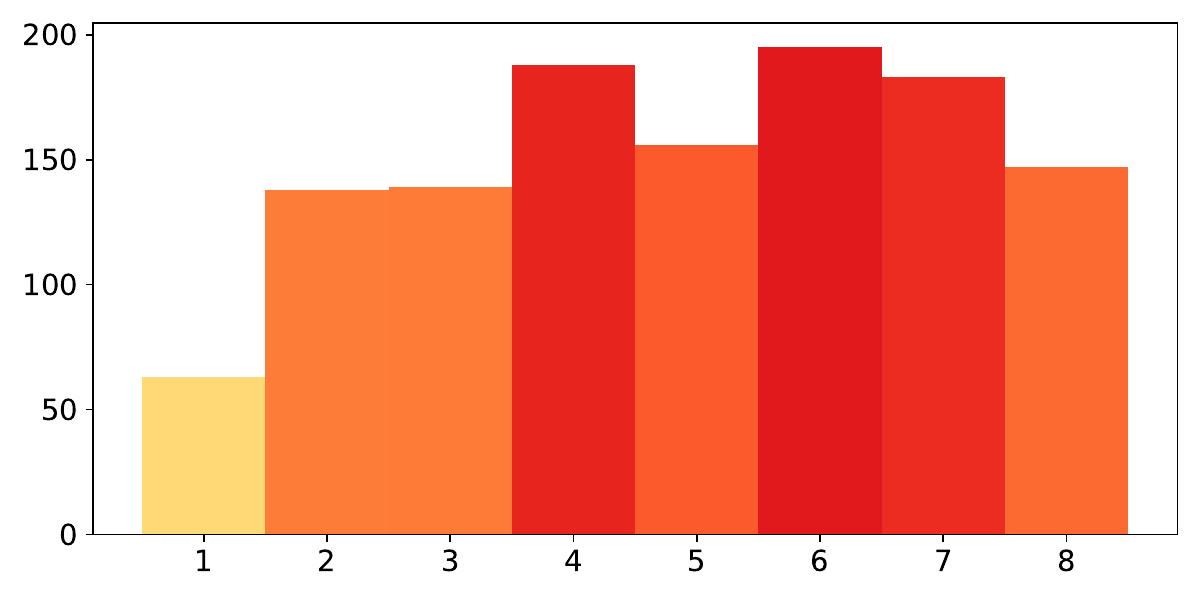} \\
     \benchnav\alltype &
     \includegraphics[width=0.35\textwidth]{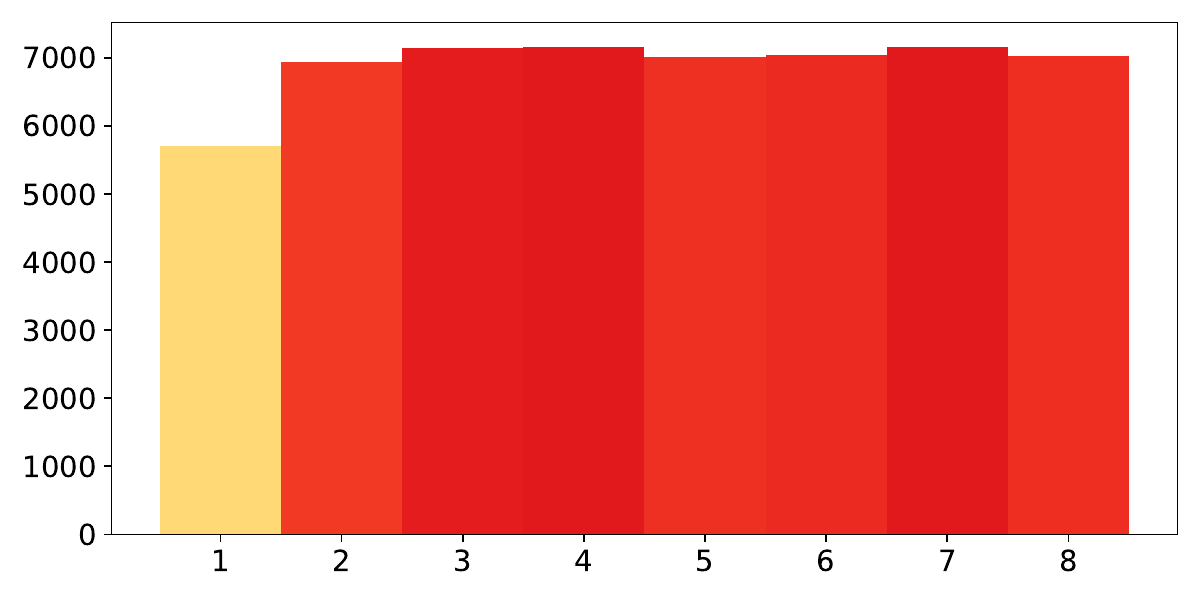} &
     \includegraphics[width=0.35\textwidth]{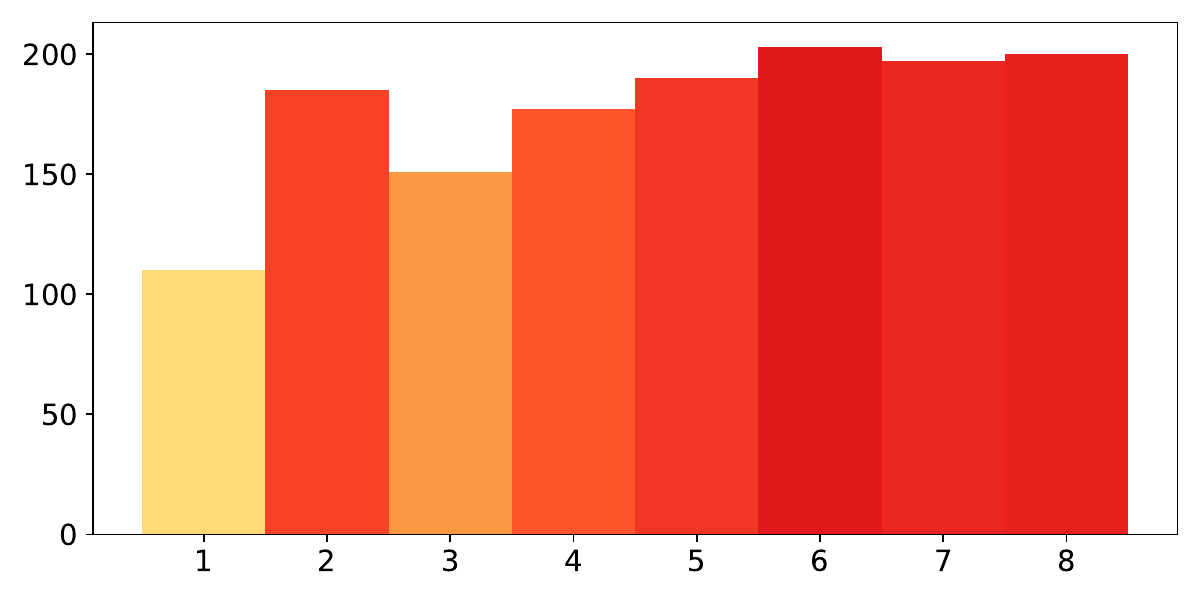} \\
    \end{tabular}
        \caption{\textbf{Distribution of number of rooms} in houses used in \bench\fifteen, \bench\alltype, \benchnav\fifteen, and \benchnav\alltype. \emph{x axis:} number of rooms in the house; \emph{y axis:} number of houses}
    \label{fig:num_rooms}
\end{figure*}

\begin{figure*}
    \centering
    \footnotesize
    \begin{tabular} {m{0.12\textwidth}m{0.35\textwidth}m{0.35\textwidth}}
     & \multicolumn{1}{c}{\textbf{Train}} & \multicolumn{1}{c}{\textbf{Evaluation}} \\
     \bench\fifteen &
     \includegraphics[width=0.35\textwidth]{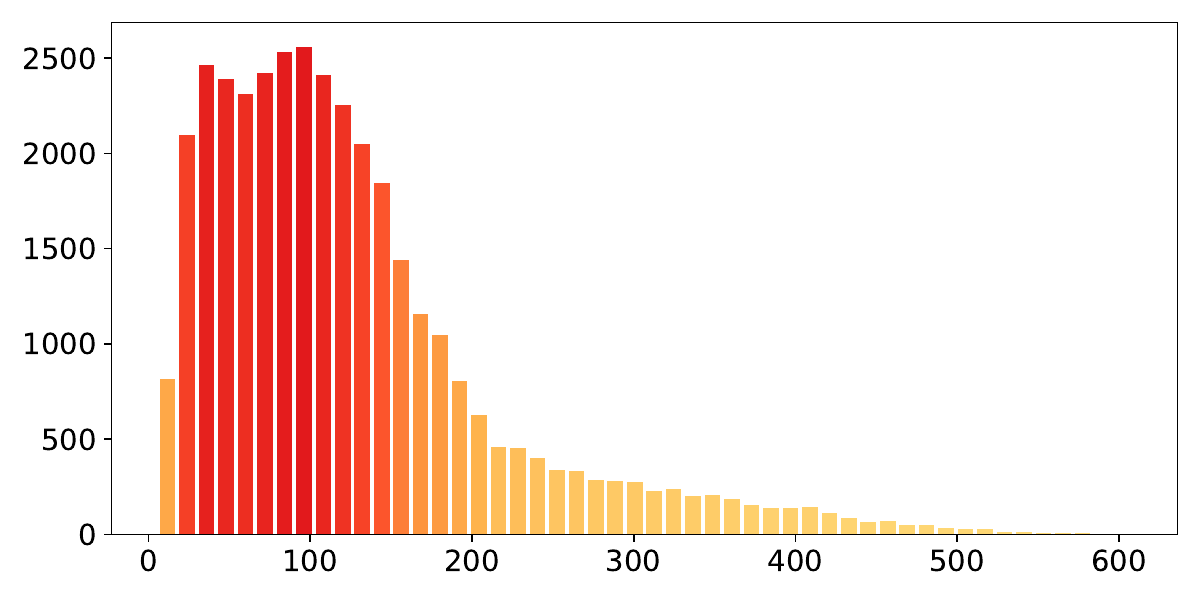} &
     \includegraphics[width=0.35\textwidth]{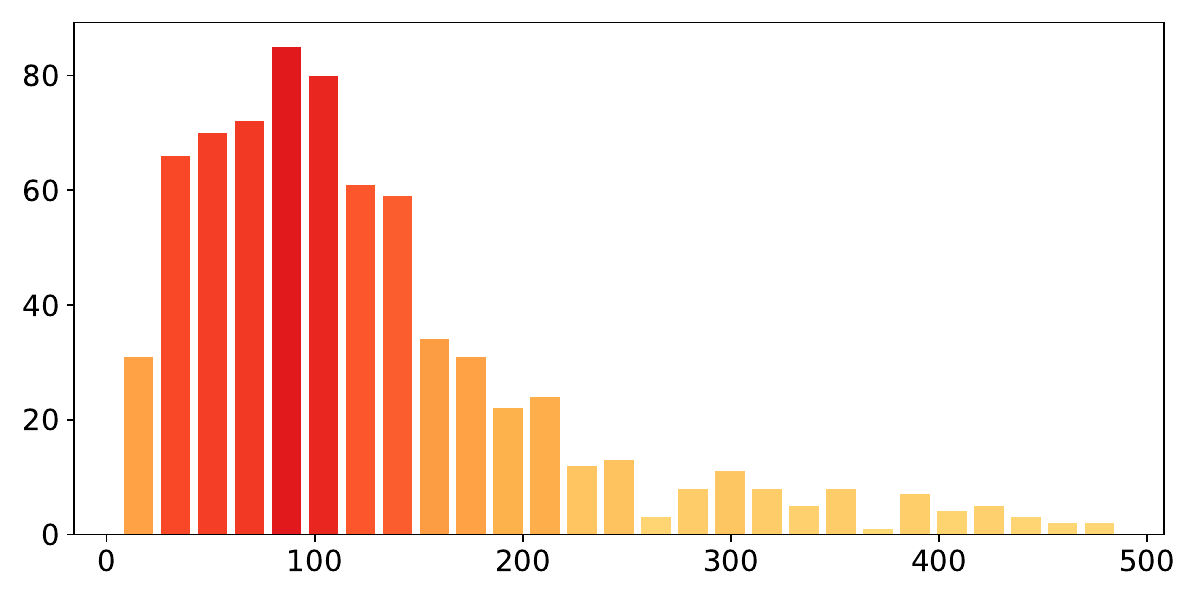} \\
     \bench\alltype &
     \includegraphics[width=0.35\textwidth]{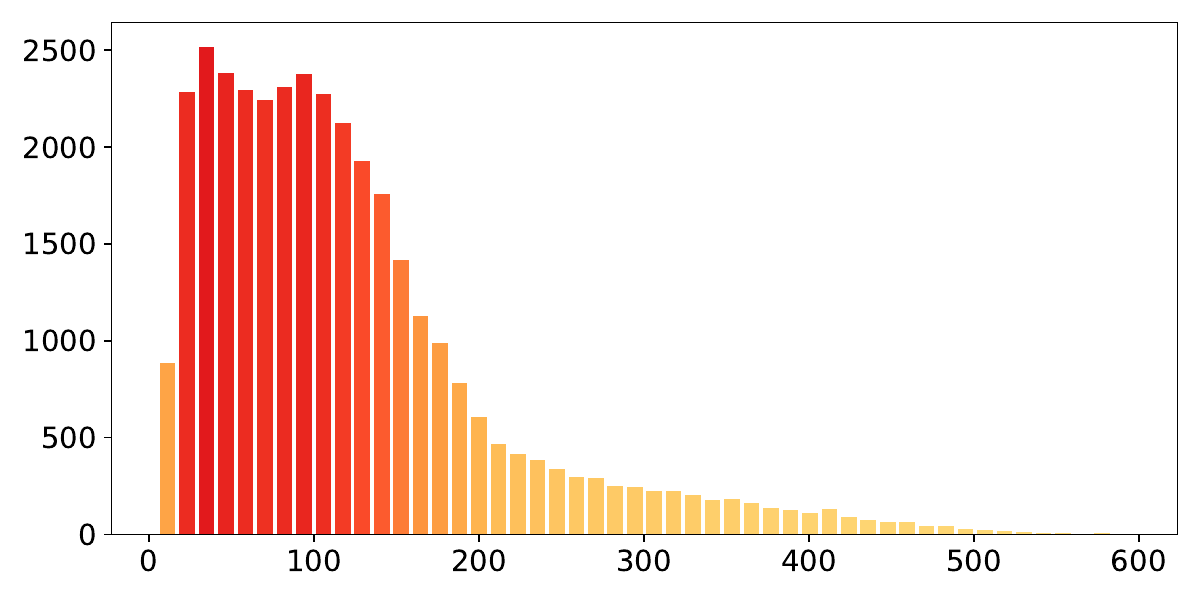} &
     \includegraphics[width=0.35\textwidth]{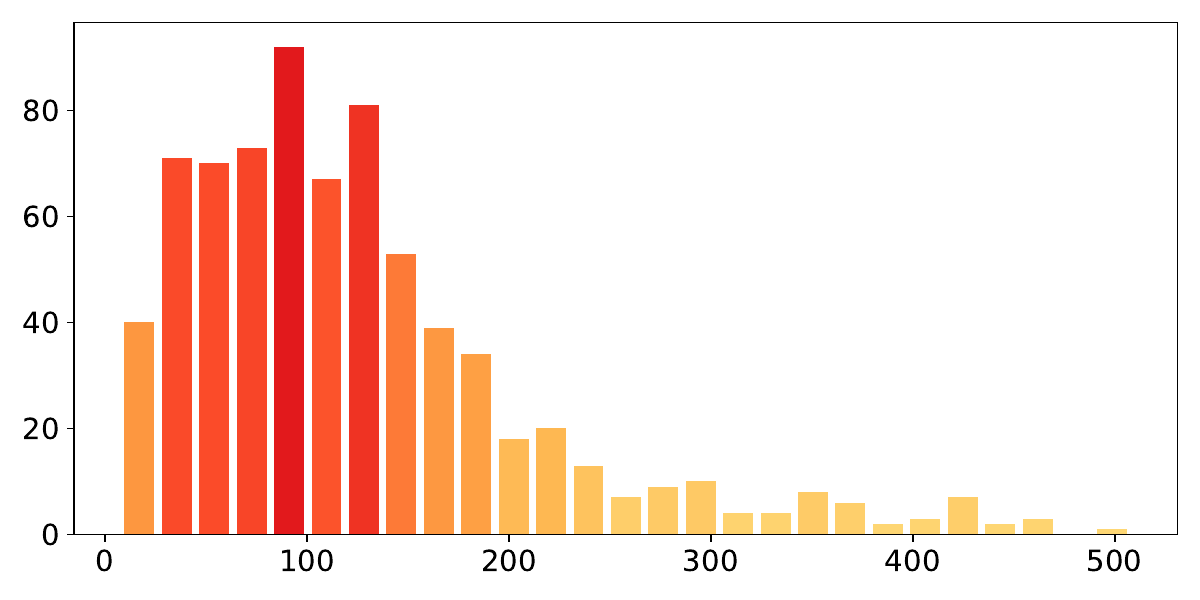} \\
     \benchnav\fifteen &
     \includegraphics[width=0.35\textwidth]{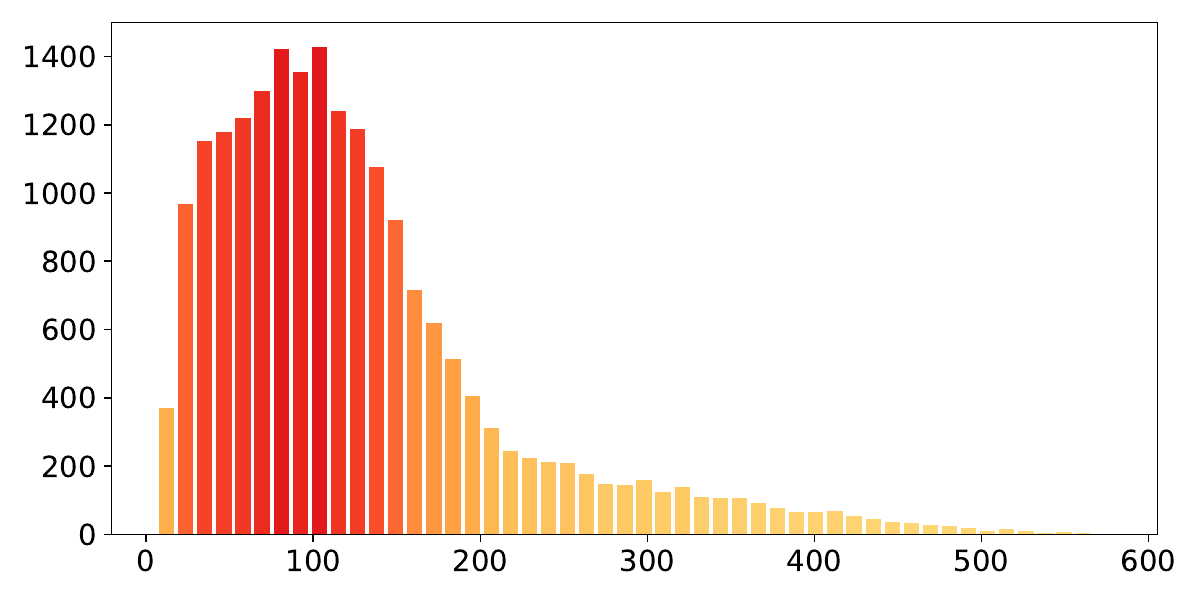} &
     \includegraphics[width=0.35\textwidth]{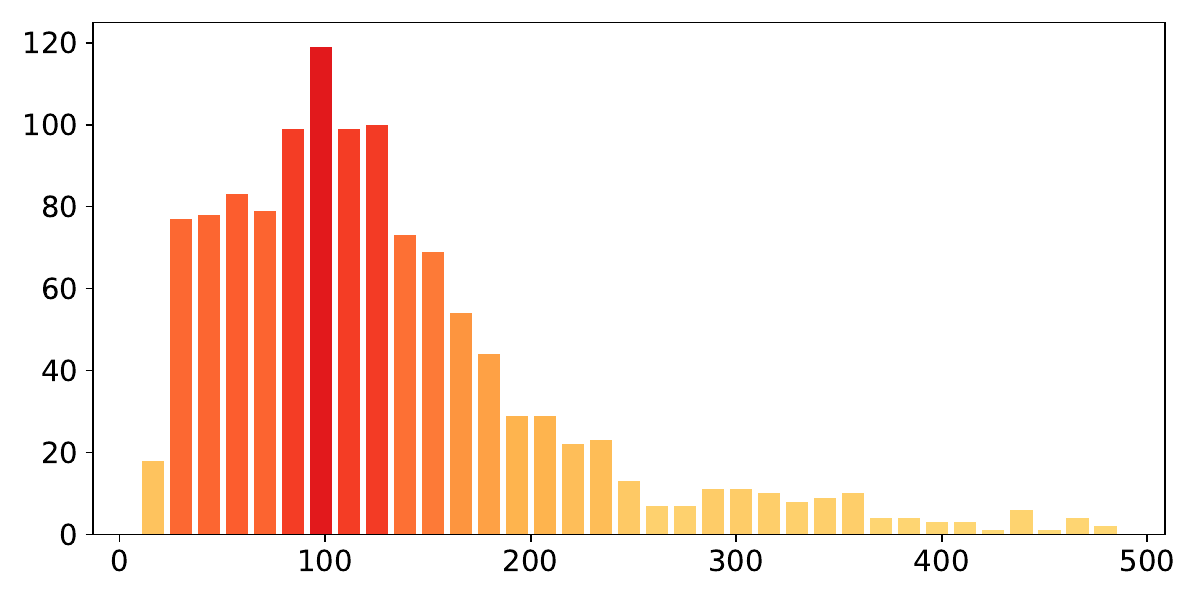} \\
     \benchnav\alltype &
     \includegraphics[width=0.35\textwidth]{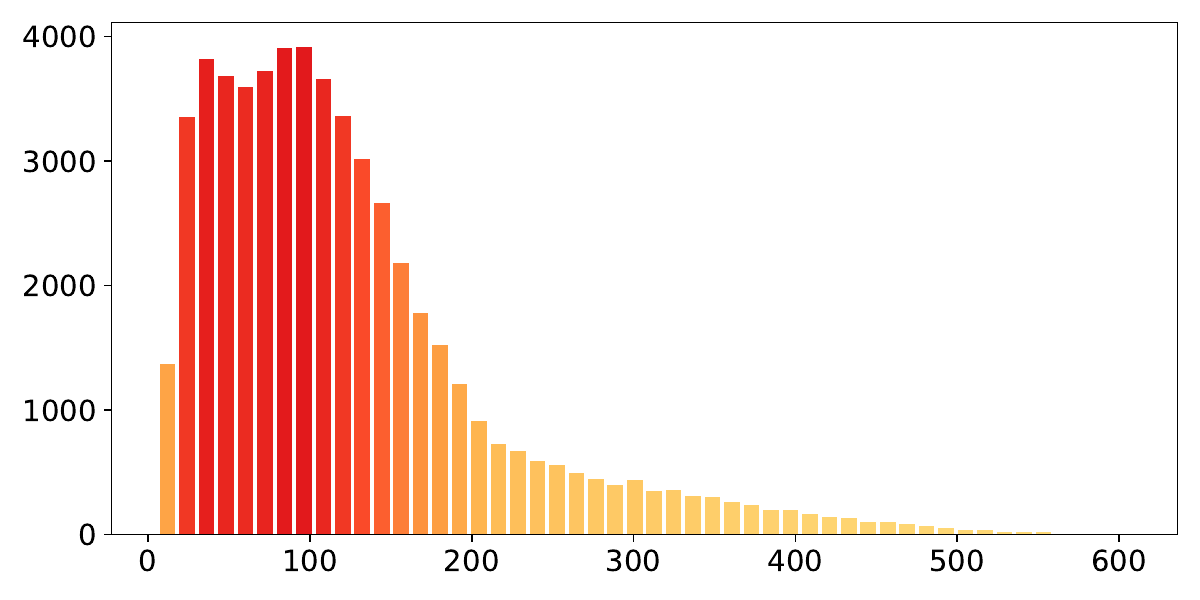} &
     \includegraphics[width=0.35\textwidth]{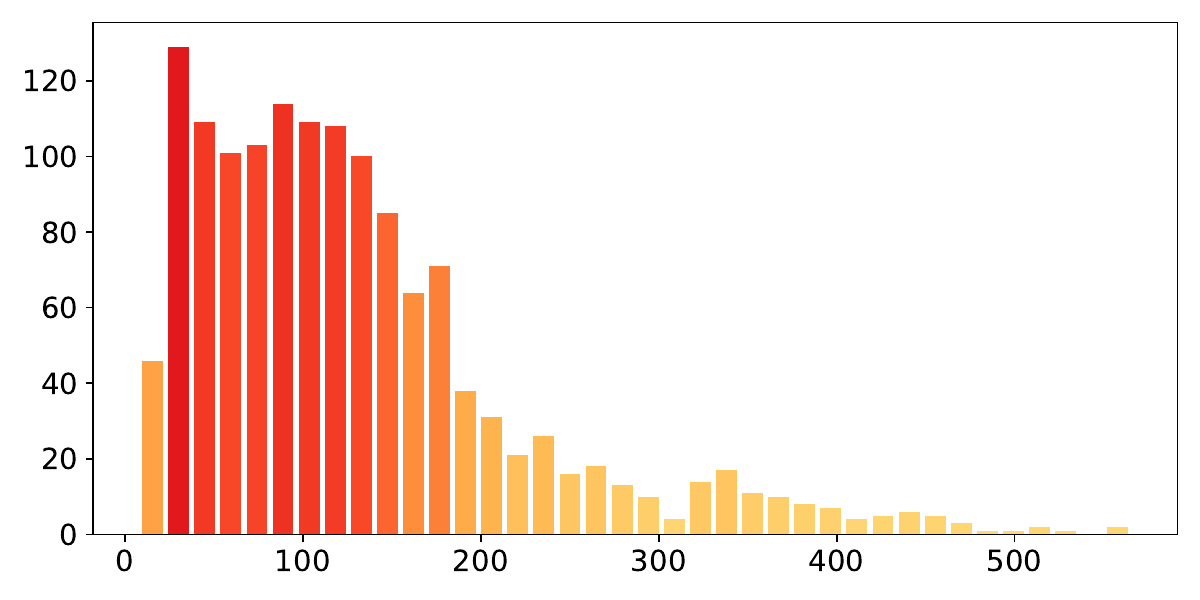} \\
    \end{tabular}
    \caption{\textbf{Distribution of house areas} (in m$^2$) in \bench\fifteen, \bench\alltype, \benchnav\fifteen, and \benchnav\alltype.}
    \label{fig:areas}
\end{figure*}

In Tables \ref{tab:chores_stats} and \ref{tab:choresnav_stats} we list the number of episodes, houses, and expert trajectory frames for the train split of \bench and \benchnav. We also list the number of episodes and houses in the evaluation split.
The number of unseen assets present in the Evaluation splits for the four combinations of \{\bench, \benchnav\} with \{\fifteen, \alltype\} is shown in Table \ref{tab:unseen_assets}.

\begin{table}
    \centering
    \footnotesize
    \begin{tabular}{|l|r|r|r|}
    \hline
    \textbf{Benchmark} & \textbf{Unseen assets} & \textbf{\% unseen assets} & \textbf{Houses} \\
    \hline
    \bench\fifteen & 1,746 & 24.0\% & 727 \\
    \hline
    \bench\alltype & 1,740 & 24.2\% & 737 \\
    \hline
    \benchnav\fifteen & 2,129 & 23.5\% & 1,209 \\
    \hline
    \benchnav\alltype & 2,079 & 22.1\% & 1,413 \\
    \hline
    \end{tabular}
    \caption{Unseen assets in evaluation houses.}
    \label{tab:unseen_assets}
\end{table}

Regarding the variety in behaviors and natural language instructions, in Fig.~\ref{fig:eplengths} we show the distribution of planner trajectory lengths for all tasks in each of the training splits, and in Fig.~\ref{fig:target_synsets} we show the distribution of target synsets across all splits. Beyond this variety, please note that some task types introduce a large number of additional concepts in the natural language instructions: for example, each \ObjectNavOpenVocab instruction uses a unique natural language description of the target asset.

Regarding the complexity of the (virtual) training and evaluation environments, in Fig.~\ref{fig:num_rooms} we show the distribution of the number of rooms for houses included in each of the splits. Note that we only count each house once, regardless of how many times it appears in the corresponding dataset. Similarly, in Fig.~\ref{fig:areas} we show the corresponding distribution of house areas.

\paragraph{Curation of benchmark}
To ensure a robust and representative evaluation dataset, we initially curate a subset of 3,000 episodes from distinct validation houses for each task. A recursive rejection process is then applied, prioritizing tasks with higher uniqueness across dimensions such as object type/synset, lemma, hypernym, and task-specific parameters. For example, in \ObjectNavRoom, balancing extends to the target object's room, while \SimpleExploreHouse is balanced on the number of rooms in the house. Manual final filtration of benchmark tasks concludes with average of 195 instances per task, enhancing the reliability of our evaluation framework.

\section{Sim-to-real transfer}
\label{sec:real_all}
\subsection{Hardware design.} 
\label{sec:hardware}

\begin{figure}[ht]
    \centering
    \begin{subfigure}{\columnwidth}
        \centering
        \includegraphics[width=0.48\textwidth]{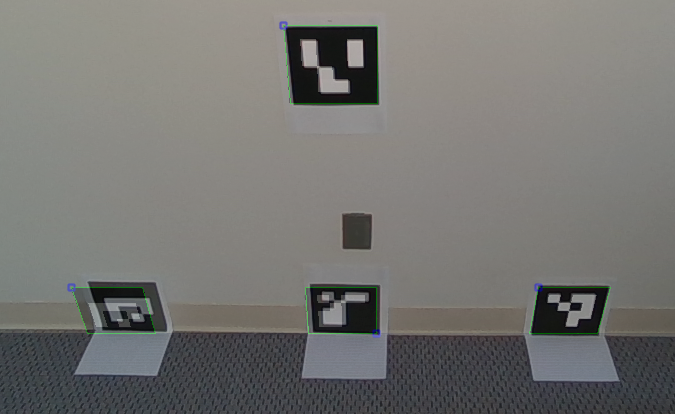}
        \caption{Manipulation camera view of calibration setup with best alignment}
    \end{subfigure}

    \begin{subfigure}{\columnwidth}
        \centering
        \includegraphics[width=0.48\textwidth]{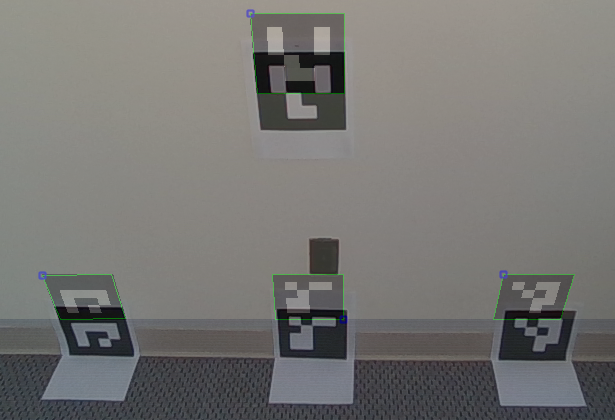}
        \caption{Manipulation camera view of calibration setup with 3$^{\circ}$ of misalignment}
    \end{subfigure}

    \caption{Simulated and real images of a fixed arrangement of Aruco markers was used to ensure correct camera horizon.}
    \label{fig:calibration}
\end{figure}

The physical and simulated embodiment of our agent is based on the Hello Robot Stretch RE-1 \cite{Kemp2022StretchRobot} mobile manipulator. We equip the Stretch with two identical Intel RealSense 455 fixed cameras, namely the \emph{navigation} and the \emph{manipulation} camera, both with a vertical field of view of 59$^{\circ}$ and capable of $1280{\times}720$ RGB-D image capture. The navigation camera is placed looking in the agent's forward direction and points slightly down, with the horizon at a nominal 30$^\circ$. The manipulation camera is placed 90$^\circ$ in clockwise direction apart from the navigation camera around the vertical axis (\ie, it looks to the right of the robot's forward direction, at the manipulator) and also points slightly down, also with a nominal 30$^{\circ}$ horizon. A 30$^\circ$ horizon for each camera was chosen to optimize the agent's perspective of its functional working space and potential navigation/manipulation targets. Quirks of fabrication and attachment mean that real horizons may have some variability up to $\pm3^{\circ}$. We calibrated all platforms with an Aruco marker setup (shown in Fig. \ref{fig:calibration}) that matched in simulation and real to validate the real horizons were near-nominal and varied the simulated camera horizons in training data generation. The Stretch is equipped with the standard lift and telescoping arm, which allows reaching at heights of up to 110 cm and distances of up to 86.7 cm from the vertical line passing through the manipulation camera's center of projections.
The STL files for 3D printing the dual-camera mount will be made available. The mount as designed may be adjusted for horizons in $[0^{\circ},60^{\circ}]$.

\subsection{Sensors.} 
\label{sec:real_sensors}The primary input is the RGB sensors corresponding to both Intel 455 cameras, which for our purposes to match simulation return resized images of $396{\times}224$ pixels. The RGB images are then cropped to $384{\times}224$. A binary ObjectInHand sensor is provided by determining if gripper effort (provided by Stretch API) corresponds to a positive grasp with force applied as opposed to an open gripper.

\subsection{Real Evaluation Environments}
\label{sec:real_environments}
We asses the performance of our models on \ObjectNav and \Fetch in a 6-room apartment also used in Phone2Proc\cite{Deitke2023Phone2Proc}, Pickup in RoboThor \cite{Deitke2020RoboTHOR}, and Explore in both environments. The 6-room apartment contains environment variations wholly unseen at train time, including: a new configuration (multiple rooms off a long corridor), two new room types (office and corridor), rooms with non-orthogonal wall alignment, and many unseen object instances. These factors, combined with traditional sim-to-real challenges like substantial texture variations and dynamic lighting changes, contribute to a comprehensive evaluation of the proposed method's robustness and generalization capabilities. The environments are showcased in agent trajectory videos highlighted in the supplementary website included along with this PDF.

\subsection{Grasping}
\label{sec:grasping}
In the real world we leverage detection and depth for heuristic last-step grasp planning using arm motions alone. If the gripper isn't close enough or grasping would require base motions, the grasp is considered failed. For potentially successful attempts, we use detection with \detic and instance segmentation from FastSAM \cite{zhao2023fast}, aligning the gripper with the projected object center and approaching from above. This approach, adapted for the Stretch RE1's limited dexterity, lacks considerations for object disturbance or protrusions and is susceptible to segmentation or detection issues, impacting overall success. We therefore report manipulation task success based on policy success (object proximity) and full success (both policy and heuristic grasping) in Table \ref{tab:real_world}.

\subsection{Real task evaluation details}
\label{sec:real_evaluations}
Object-oriented episodes each have three starting positions which are shared across objects, \eg ``navigate to an apple'' is evaluated three times: once from the living room, once from the middle of the corridor, and once from the kitchen.
\paragraph{\ObjectNav.} Target objects are Sofa, Bed, Chair, Apple, Vase, and Houseplant, each from three starting positions. 

\paragraph{\Fetch.} Target objects are Apple, Vase, and Houseplant from the same three starting positions. In one small change from \ObjectNav episodes, object instances are replaced with instances which better fit into Stretch's grasping envelope and in some cases at a better height for interaction, but availability and placement are nearly identical.

\paragraph{\PickUp} Objects are placed on three different surfaces (coffee table, desk, and nightstand) at three different heights. Objects are Apple, Houseplant, Spray Bottle, Mug, and Vase. Grasping and success are as described in \ref{sec:grasping}. 

\paragraph{\SimpleExploreHouse} The full 6-room apartment is explored, and then partitioned into two 3-room apartments to evaluate the ability of \model to explore large and small spaces. We additionally explore a section of RoboTHOR and attached workroom as a novel 3-room apartment.

\section{Discussion}
\subsection{Failure Analysis}
As highlighted in Lines 448-462 of the main paper, our experiments using ground truth object detection demonstrated a notable improvement of $32$\% over our RGB-only model. This significant enhancement shows that the majority of the failures is due to perception problems. One promising approach involves training models to utilize the output from off-the-shelf object detection models, such as DETIC~\cite{Zhou2022DETIC}, as input for object recognition. Our real-world evaluations have shown this method to be effective.

We also hypothesize that pretraining our end-to-end models on tasks that necessitate a more object-aware policy could result in further improvements. This approach could enhance the model's ability to recognize and interact with objects in its environment more effectively.

Additionally, our analysis revealed that in the fetch task, the agent fails to pick up the object $29$\% of the time when it is in proximity. 
These failures are primarily due to the robotic arm knocking over the object or missing it due to a slight distance between the grasper and the target object. Improving the precision of the arm's movements and its ability to recover from unsuccessful grasps could significantly enhance performance in such tasks.

Furthermore, within the \benchnav tasks, the task involving open vocabulary object descriptions yielded the lowest success rate. Although a success rate of $30.6$\% is remarkable, we believe that enriching the model's visual and language representations could substantially benefit its performance. These findings suggest valuable directions for future research in improving the effectiveness of robotic task execution.

\subsection{Limitations}

A primary limitation of our study is the reliance on an off-the-shelf grasping planner. Integrating this component into our full end-to-end learning pipeline could potentially enhance the model's efficiency and robustness, particularly in selecting varied grasp positions on objects. Such integration necessitates accommodating physical grasping within the simulation environment.

Another limitation is the robot's operational speed and dexterity. Our project is constrained by the capabilities of commercially available robotic hardware. The development of faster, more affordable, and lighter robots by the broader academic and industrial communities would be greatly beneficial for advancing research in this domain.

These limitations offer avenues for future work and underscore the potential for innovation in robotic systems.

\subsection{Why are the shortest path trajectories sufficient?}
 
Our intuition is:
the huge diversity of our procedural scenes, large-scale data collection, use of pretrained visual encoders, and extensive visual augmentations
enable sufficient state coverage so that generalization is possible. 
Even given the above, it is surprising that our agent learns to backtrack; we conjecture that our agent acquires a set of skills by imitating expert trajectories. During training, the agent is exposed to trajectories including various skills, such as navigating between rooms, moving towards objects, and avoiding collisions. During evaluation, even if the agent has not observed backtracking behavior, it has encountered all the necessary subskills. As the agent navigates the environment, it switches between different modes (or skills) to complete the task.

\subsection{Author Contributions}
\begin{footnotesize}
$^\Omega$ Core technical contributors; \quad $^\dagger$ Team lead.
\end{footnotesize}
~\\

\mypara{Kiana Ehsani.$^\Omega$} \\
Contributed to SPOC modeling and training. Led developing expert planners for data generation, hardware calibration and real world evaluation. Trained and evaluated SPOC policies using detection models. ~\\[-0.9em]

\mypara{Tanmay Gupta.$^\Omega$} \\
Led the development of the SPOC model and the IL training pipeline. 
~\\[-0.9em]

\mypara{Rose Hendrix.$^\Omega$} \\
Contributed to developing expert planners and samplers for data generation, evaluating SPOC in the real world, and refining the CHORES benchmark.
~\\[-0.9em]

\mypara{Jordi Salvador.$^\Omega$} \\
Contributed to developing expert planners and samplers for data generation.
~\\[-0.9em]

\mypara{Luca Weihs.$^\Omega$} \\
Designed the distributed data collection pipeline, led annotation and Objaverse$\leftrightarrow$AI2-THOR integration efforts, contributed to expert planners/samplers, ran experiments.
~\\[-0.9em]

\mypara{Kuo-Hao Zeng.$^\Omega$} \\
Implemented multi-node training experiments, participated in model development, and implemented RL baselines.
~\\[-0.9em]

\mypara{Kunal Pratap Singh.} \\
Contributed to initial stages of data collection, evaluator design, and RL training.
~\\[-0.9em]

\mypara{Yejin Kim.} \\
Contributed to real-world experiments and designed the real-world grasping heuristic.
~\\[-0.9em]

\mypara{Winson Han.} \\
Provided AI2-THOR integration support.
~\\[-0.9em]

\mypara{Alvaro Herrasti.} \\
Designed framework enabling loading optimized Objaverse assets into AI2-THOR at runtime.
~\\[-0.9em]

\mypara{Ranjay Krishna.} \\
Brainstorming and high-level discussion/direction.
~\\[-0.9em]

\mypara{Dustin Schwenk.} \\
Designed and ran human Objaverse annotation collection via Amazon Mechanical Turk.
~\\[-0.9em]

\mypara{Eli VanderBilt.} \\
Created the Objaverse asset optimization pipeline which converts arbitrary input GLB files into a highly optimized format compatible with high-speed rendering.
~\\[-0.9em]

\mypara{Aniruddha Kembhavi.$^\dagger$} \\
Led the project. Contributed to model and experimental design. Provided direction on the benchmark and real world evaluation. Wrote the paper.
~\\[-0.9em]

\end{document}